\documentclass[journal]{IEEEtran}
\ifCLASSINFOpdf
\else
\fi
\usepackage{amsmath}
\usepackage{amsfonts}
\usepackage{amssymb}
\usepackage{bm}
\usepackage{xspace}
\usepackage[pagebackref=true,breaklinks=true,letterpaper=true,colorlinks,bookmarks=false]{hyperref}
\usepackage{caption}
\usepackage{subcaption}
\usepackage{graphicx}
\usepackage{multirow}
\usepackage{adjustbox}
\usepackage{sidecap}
\usepackage{tabularx}
\usepackage{pifont}
\usepackage{enumitem}
\usepackage[framemethod=tikz]{mdframed}
\usepackage{tikz}

\DeclareMathOperator{\sign}{sign}
\makeatletter
\DeclareRobustCommand\onedot{\futurelet\@let@token\@onedot}
\def\@onedot{\ifx\@let@token.\else.\null\fi\xspace}

\def\ie{\emph{i.e}\onedot}

\def\etal{\emph{et al}\onedot}
\makeatother

\newcolumntype{Y}{>{\centering\arraybackslash}X}

\graphicspath{{figure/}}

\begin{document}
%
\title{Speeding up the Bilateral Filter:  A Joint Acceleration Way}
%
%
%

\author{Longquan~Dai,
            Mengke~Yuan,
           and~Xiaopeng~Zhang,~\IEEEmembership{Memeber,~IEEE}
\thanks{Manuscript received August 4, 2015; revised December 8, 2015 and February 8, 2016; accepted March 18, 2016. This work was supported in part by the National Natural Science Foundation of China under Grant 61331018, Grant 91338202, Grant 61572405, and Grant 61571046, in part by the China National High-tech R\&D Program (863 Program) under Grant 2015AA016402. The associate editor coordinating the review of this manuscript and approving it for publication was Dr. Debargha Mukherjee. (Corresponding author: Xiaopeng Zhang.) }
\thanks{L. Dai  is with with the National Laboratory of Pattern Recognition, Institute
of Automation, Chinese Academy of Sciences, Beijing 100190, China (e-mail:
longquan.dai@ia.ac.cn).}
\thanks{M. Yuan is with the National Laboratory of Pattern Recognition, Institute
of Automation, Chinese Academy of Sciences, Beijing 100190, China (e-mail:
mengke.yuan@nlpr.ia.ac.cn).}
\thanks{X. Zhang is with the National Laboratory of Pattern Recognition, Institute
of Automation, Chinese Academy of Sciences, Beijing 100190, China (e-mail:
xiaopeng.zhang@ia.ac.cn).}
}

%
%

\markboth{IEEE TRANSACTIONS ON IMAGE PROCESSING}%
{Shell \MakeLowercase{\textit{et al.}}: Bare Demo of IEEEtran.cls for Journals}
%



\maketitle

\begin{abstract}
Computational complexity of the brute-force implementation of the bilateral filter (BF) depends on its filter kernel size. To achieve the constant-time BF whose complexity is irrelevant to the kernel size, many techniques have been proposed, such as 2-D box filtering, dimension promotion and shiftability property. Although each of the above techniques suffers from accuracy and efficiency problems, previous algorithm designers were used to take only one of them to assemble fast implementations due to the hardness of combining them together. Hence no joint exploitation of these techniques has been proposed to construct a new cutting edge implementation that solves these problems. Jointly employing five techniques: kernel truncation, best $N$-term approximation as well as previous 2-D box filtering, dimension promotion and shiftability property, we propose a unified framework to transform BF with arbitrary spatial and range kernels into a set of 3-D box filters that can be computed in linear time. To the best of our knowledge, our algorithm is the first method that can integrate all these acceleration techniques and therefore can draw upon one another's strong point to overcome deficiencies. The strength of our method has been corroborated by several carefully designed experiments. Especially, the filtering accuracy is significantly improved without sacrificing the efficiency at running time.
\end{abstract}


%
\IEEEpeerreviewmaketitle

\section{Introduction}

The bilateral filter (BF) is probably one of the most fundamental tools in computer vision and graphics applications~\cite{Paris_BOOK_2009}. The concept of BF was first introduced by Aurich~\etal~\cite{Aurich_DAGM_1995}  under the name ``nonlinear Gaussian filter'' in 1995, and then by Smith~\etal~\cite{Smith_IJCV_1997} within the so-called ``SUSAN'' approach in 1997. Later, it was rediscovered by Tomasi~\etal~\cite{Tomasi_ICCV_1998} with the current name ``bilateral filter'' in 1998. The basic idea underlying bilateral filtering is to do in the range domain of an image what traditional spatial filters~\cite{Haddad_TSP_1991} do in its spatial domain, because BF considers that two pixels which are close or visually similar to one another have the same perceptual meaning. Unlike traditional spatial filters, the weights of BF take into account both spatial affinity and intensity similarity with respect to the central pixel. Therefore BF can be used to preserve edges while performing smoothing.


BF's output at pixel $\bm{x} = (x, y)$ is a weighted average of its neighbors $\mathcal{N}_{\bm{x}}$. The weights assigned to the pixels in $\mathcal{N}_{\bm{x}}$ are inversely proportional to both the distance in the spatial domain $\mathcal{S}$ and the dissimilarity in the range domain $\mathcal{R}$. Let $I$ be a gray-level image,
$K_r(x)$ and $K_s(x)$ be decreasing functions on the region $\mathbb{R}^+$ and symmetric functions on the entire definition domain $\mathbb{R}$,  BF is specified as follows:
%
\begin{IEEEeqnarray}{C}
\hat{I}(\bm{x}) = \frac{\sum_{\bm{y} \in \mathcal{N}_{\bm{x}}} K_s(\| \bm{x} - \bm{y} \|) K_r( I(\bm{x}) - I(\bm{y}) ) I(\bm{y}) }{ \sum_{\bm{y} \in \mathcal{N}_{\bm{x}}} K_s( \| \bm{x} -  \bm{y}\| ) K_r(I(\bm{x}) - I(\bm{y})) }
\label{eq:BF}
\end{IEEEeqnarray}
Although the Gaussian function $G_{\sigma}(x) = \exp(-{x^2}/{2\sigma^2})$ is a common choice for the spatial and range kernels, the options are not unique. More candidates can be found in~\cite{Durand_TOG_2002}.

As a simple, non-iterative and edge-preserving filtering tool, BF has been found with a wide range of applications such as  stereo matching~\cite{Yang_TPAMI_2014}, flash and no-flash images fusion~\cite{Petschnigg_TOG_2004} and contrast enhancement~\cite{Elad_SS_2005}. However, on the flip side of the power, the complexity of its brute-force implementation is $O(|\mathcal{N}_{\bm{x}}||I|)$, where $|I|$ is the number of pixels of the image $I$ and $|\mathcal{N}_{\bm{x}}|$ is the size of the neighborhood $\mathcal{N}_{\bm{x}}$. Since $O(|\mathcal{N}_{\bm{x}}||I|)$ relies on the size of $\mathcal{N}_{\bm{x}}$, the run time increases with the size of $\mathcal{N}_{\bm{x}}$. We thus have to spend several minutes for final results when $\mathcal{N}_{\bm{x}}$ is large. It is unacceptable for time-critical applications, such as stereo matching~\cite{Ansar_3DPVT_2004} and video abstraction~\cite{Winnemoller_TOG_2006}.


Considering the importance of BF in practice, we will study and retrofit the acceleration of BF to reduce its computational complexity from $O(|\mathcal{N}_{\bm{x}}||I|)$ down to $O(|I|)$. A typical acceleration approach is first to decompose $\sum_{\bm{y} \in \mathcal{N}_{\bm{x}}} K_s( \|\bm{x} - \bm{y}\|) K_r(I(\bm{x})-I(\bm{y})) I(\bm{y}) $ into a set of linear convolutions $\sum_{\bm{y} \in \mathcal{N}_{\bm{x}}} K_s(\|\bm{x} - \bm{y}\|) g(\bm{y}) $ and then to speed up the linear convolution, where $g(\bm{x})$ is a scalar function.
 %
 %
In the literature, a distinction was made between the two operations and therefore different strategies are adopted to accelerate them. Particularly, some fast implementations are limited to accelerate the specific Gaussian kernel.  Unlike these approaches, we propose a unified framework to complete the two tasks. At first, we take advantage of the best $N$-term approximation of $K_r(x)$ on truncated trigonometric functions to transform $\sum_{\bm{y} \in \mathcal{N}_{\bm{x}}} K_s(\|\bm{x} - \bm{y}\|) K_r(I(\bm{x}) - I(\bm{y})) I(\bm{y}) $ into a set of linear convolution $\sum_{\bm{y} \in \mathcal{N}_{\bm{x}}} K_s(\|\bm{x} - \bm{y}\|) g(\bm{y}) $, then we exploit the best $N$-term approximation of $K_s(x)$ on 2-D Haar functions to decompose $\sum_{\bm{y} \in \mathcal{N}_{\bm{x}}} K_s(\|\bm{x} - \bm{y}\|) g(\bm{y}) $ into a set of 3-D box filters. More importantly, we disclose that our implementation cannot only be fast computed by the summed area table~\cite{Crow_SIGGRAPH_1984}, but also be used to speed up BF with arbitrary kernels.

Contributions of this paper are fourfold: $\bm{1)}$, we propose the truncated kernels which are exploited to replace BF's original kernels; $\bm{2)}$, we use the best $N$-term approximation on Haar functions and truncated trigonometric functions to approximate the truncated spatial and range kernels respectively; $\bm{3)}$, we find that the product of the two best $N$-term approximations can be fast computed by 3-D box filters; $\bm{4)}$, compared with other methods, our filtering accuracy can be significantly improved without sacrificing efficiency. In order to clarify our contribution, the rest of this paper is structured as follows. First, existing fast bilateral filtering algorithms are reviewed in section~\ref{sec:related_work}. After that, section~\ref{sec:techniques} lists the background techniques. Sequentially, our proposed method is described in section~\ref{sec:method} and a full comparison is conducted with other acceleration techniques in section~\ref{sec:comparison}.

\section{Related work}
\label{sec:related_work}
Fast implementations of BF can be roughly divided into two categories: the first one is the high-dimensional implementations~\cite{Adams_TOG_2009,Adams_CGF_2010,Yoshizawa_CGF_2010}, the second one is the low-dimensional case. 
In this paper, we focus on the second one.  For a clear description of our method, we briefly introduce the low dimensional acceleration algorithms in two  aspects below.

\subsection{Speeding up the linear convolution of $K_s(x)$}

Acceleration techniques  of computing  $\sum_{\bm{y} \in \mathcal{N}_{\bm{x}}} K_s(\|\bm{x}  -  \bm{y}\|) g(\bm{y})$ have been well studied in the literature. Here, we roughly divided them into three classes: Fast Fourier Transform, Kernel separation and Box filtering.

\subsubsection{Fast Fourier Transform (FFT)}

Durand~\etal~\cite{Durand_TOG_2002} first employed FFT to fast compute $\sum_{\bm{y} \in \mathcal{N}_{\bm{x}}} K_s(\| \bm{x}  -  \bm{y} \|) g(\bm{y})$ as the linear convolution of $K_s(x)$ can be greatly accelerated using FFT. In mathematics, an $O(|\mathcal{N}_{\bm{x}}||I|)$ convolution with arbitrary $K_s(x)$ in the spatial domain becomes multiplication in the frequency domain with $O(|I|)$ complexity. Although FFT can be used to produce accurate filtering, FFT and its inverse have the cost $O(|I| \log (|I|))$. But, in practice, an algorithm with linear complexity $O(|I|)$ is preferred.

\subsubsection{Kernel separation}

Kernel separation based methods decompose 2-D filter kernel into two 1-D kernels. Rows of an image are filtered at first. After that, the intermediate result is filtered along the columns~\cite{Pham_ICME_2005}. Yang~\etal~\cite{Yang_CVPR_2009} advocated using Deriche's recursive method~\cite{Rachid_TR_1993} to approximate Gaussian filtering. More methods can be found in \cite{Getreuer_ipol_2013}. Compared with FFT, this kind of algorithms is much faster, but they do not perform well in texture regions.

\subsubsection{Box filtering} A 2-D box filter $\ddot{B}(\bm{x})$ is a spatial domain linear filter in which each pixel $\bm{x}$ has a value equal to the average of its neighboring pixels $\bm{y} \in \mathcal{N}_{\bm{x}}$ of the input image. Due to the property of equal weights, box filters can be implemented using the summed area table~\cite{Crow_SIGGRAPH_1984} which is significantly faster than using a sliding window algorithm. Note that box filters can be used to approximate the Gaussian filter. 
In order to decompose the Gaussian spatial kernel into several box functions, Zhang \etal~\cite{Zhang_TIP_2012} employed the de Moivre-Laplace theorem, which says that for $k$ in the neighborhood of $np$, $\lim \limits_{n \rightarrow \infty}{n \choose k} p^k (1-p)^{n-k} = \frac{1}{\sqrt{2\pi np(1-p)}} \exp(-\frac{(k-np)^2}{2np(1-p)}) $. The method however is not problem-free because it can only be applied to speed up the Gaussian filter. Gunturk~\cite{Gunturk_TIP_2011} generalized the Gaussian spatial kernel to arbitrary kernels and employed the least squares optimization to find the optimal coefficients $\beta^s_i$ that minimize the approximation error $(K_s(\|\bm{x}\|) - \sum_{i = 1}^{M_s} \beta^s_i \ddot{B}_{\mathcal{N}^i_{\bm{x}}}(\bm{x}))^2$. Unlike Gunturk, Pan~\etal~\cite{Pan_MPE_2014} formulated their objective function from the sparsity perspective and exploited the efficient Batch-OMP algorithm~\cite{Ron_TR_2008} to solve the optimal coefficients $\beta^s_i$ as well as the window radius of $\mathcal{N}^i_{\bm{x}}$.

\subsection{Speeding up the nonlinear convolution of $K_r(x)$}

Unlike the spatial kernel $K_s(x)$, the range kernel $K_r(x)$ introduces nonlinearity to BF as $K_r(I(\bm{x}) -  I(\bm{y}))$ is signal-dependent. A common idea shared by algorithm designers to speed up $\sum_{\bm{y} \in \mathcal{N}_{\bm{x}}} K_s(\|\bm{x} - \bm{y}\|) K_r(I(\bm{x}) - I(\bm{y})) I(\bm{y}) $ is to transform the nonlinear convolution of $K_r(x)$ into a set of linear convolutions of $K_s(x)$. Roughly, there are three kinds of acceleration techniques which are dimension promotion, principle bilateral filtered image component and shiftability property, respectively.

\subsubsection{Dimension promotion}

This acceleration technique is obtained through representing an image in 3-D space by adding the intensity to the spatial domain of an image as the 2-D nonlinear convolution of $K_r(x)$ becomes linear convolution in 3-D space which is easy to speed up. This idea is similar to the well-known level set method~\cite{Osher_JCP_1988} which considers that the breaking and merging operations are hard to perform in 2-D space, but they can be easily handled in higher dimensional space.

Mathematically, let $\delta_z(x)$ be an impulse function at $z$ and $F(\bm{y}, z) = I(\bm{y}) \delta_{I(\bm{y})}(z)$, BF in~\eqref{eq:BF} can be transformed to
\begin{IEEEeqnarray}{C}
\hat{I}(\bm{x}) = \frac{\sum_z  K_r(I(\bm{x}) - z) \sum_{\bm{y} \in \mathcal{N}_{\bm{x}}} K_s(\| \bm{x} - \bm{y}\| ) F(\bm{y}, z) }{  \sum_z K_r(I(\bm{x}) - z)\sum_{\bm{y} \in \mathcal{N}_{\bm{x}}} K_s(\| \bm{x} - \bm{y} \|)  }
\label{eq:BF_promotion}
\end{IEEEeqnarray}
where $z$ can be explained as the sample in the range domain. For an 8-bit image, $z \in [0, 255]$. In~\eqref{eq:BF_promotion}, BF is decomposed in the range domain, and the nonlinear relationship between $I(\bm{y})$ and $I(\bm{x})$ in the range kernel is eliminated. Hence the response of BF can be computed by first performing linear convolution on the auxiliary image $I(\bm{y}) \delta_{I(\bm{y})}(z)$ for each fixed $z$ and then calculating the sum weighted by $K_r(I(\bm{y}) - z)$ along $z$.

Porikli~\cite{Porikli_CVPR_2008} first employed this technique to speed up BF. Sequentially, Zhang~\etal~\cite{Zhang_TIP_2012} applied it to the joint bilateral filtering. Incorporating with different spatial kernel acceleration methods, Gunturk~\cite{Gunturk_TIP_2011} and Pan~\etal~\cite{Pan_MPE_2014} designed two different fast BF implementations. The biggest problem of them is that they need to perform 255 times linear filtering as well as 255 addition and multiplication operations along $z$ for each fixed $\bm{y}$. This is not efficient.

\subsubsection{Principle bilateral filtered image component (PBFIC)} This method was first proposed by Durand~\etal~\cite{Durand_TOG_2002} in 2002. Seven years later, Yang~\etal~\cite{Yang_CVPR_2009} generalized this idea for fast bilateral filtering with arbitrary range kernels. 
At first, Yang transforms BF \eqref{eq:BF} into \eqref{eq:BF_PBFIC1} by letting $I(\bm{x})  = z \in [0, 255]$ for $8$-bit images
\begin{IEEEeqnarray}{C}
\hat{I}(\bm{x}) = \frac{  \sum_{\bm{y} \in \mathcal{N}_{\bm{x}}} K_s(\| \bm{x} - \bm{y}\| ) K_r(z - I(\bm{y}) ) I(\bm{y}) }{  \sum_{\bm{y} \in \mathcal{N}_{\bm{x}}} K_s(\| \bm{x} - \bm{y} \|)  K_r(z - I(\bm{y}) ) }
\label{eq:BF_PBFIC1}
\end{IEEEeqnarray}
then he defines PBFIC $\hat{I}_z(\bm{x})$ as
\begin{IEEEeqnarray}{C}
\hat{I}_z(\bm{x}) = \frac{ \sum_{\bm{y} \in \mathcal{N}_{\bm{x}}} K_s(\| \bm{x} - \bm{y} \|) J_z(\bm{y}) }{  \sum_{\bm{y} \in \mathcal{N}_{\bm{x}}} K_s(\| \bm{x} - \bm{y} \|) W_z(\bm{y}) }
\label{eq:BF_PBFIC}
\end{IEEEeqnarray}
where $W_z(\bm{y}) =  K_r(z - I(\bm{y}))$ and $J_z(\bm{y}) = W_z(\bm{y}) I(\bm{y})$.
According to \eqref{eq:BF_PBFIC1} \eqref{eq:BF_PBFIC}, BF is decomposed into a set of linear filter responses $\hat{I}_z(\bm{x})$. So that we have
\begin{IEEEeqnarray}{C}
\hat{I}(\bm{x}) = \hat{I}_{{I}(\bm{x})}(\bm{x})
\end{IEEEeqnarray}
Further, assuming only $N$ out of 256 PBFICs are used ($z \in [L_0, \cdots, L_{N-1}]$), and the intensity $I(\bm{x}) \in [L_z, L_{z+1}]$, the value of BF can be linearly interpolated as follows:
\begin{IEEEeqnarray}{C}
\hat{I}(\bm{x}) = (L_{z+1} - I(\bm{x})) \hat{I}_{z}(\bm{x}) + (I(\bm{x}) - L_{z}) \hat{I}_{z+1}(\bm{x})
\label{eq:BF_PBFIC_interpolation}
\end{IEEEeqnarray}

PBFIC  has also been used to design the bilateral grid data structure~\cite{Paris_IJCV_2009,Chen_TOG_2007} for fast BF computation. However, the approximation accuracy is usually very low because the linear interpolation is introduced to approximate filtering results. An~\etal~\cite{An_SPL_2015} provided a quantitative error analysis for it. 

\subsubsection{Shiftability property}

A range kernel $K_r(x)$ is shiftable if there exists $N$ functions such that, for every translation $\tau$, we have
\begin{IEEEeqnarray}{C}
K_r( x - \tau ) = \sum_{i = 1}^N c_i(x) K_i(\tau)
\end{IEEEeqnarray}
Based on this shiftability property, Chaudhury~\cite{Chaudhury_SPL_2011} pointed out that BF with a shiftable range kernel $K_r(x)$ can be computed in linear complexity. This is because we have
\begin{IEEEeqnarray}{C}
\begin{split}
& \sum_{\bm{y} \in \mathcal{N}_{\bm{x}}} K_s(\| {\bm{x} - \bm{y}} \|) K_r(I(\bm{x}) -  I(\bm{y})) I(\bm{y}) \\
=& \sum_{i = 1}^N c_i(I(\bm{x})) \sum_{\bm{y} \in \mathcal{N}_{\bm{x}}} K_s(\| {\bm{x} - \bm{y}} \|) K_i(I(\bm{y})) I(\bm{y})
\end{split}
\label{eq:shiftable}
\end{IEEEeqnarray}
which transforms the nonlinear convolution of the range kernel $K_r(x)$ into a set of linear convolutions of the spatial kernel $K_s(x)$ on the auxiliary image $K_i(I(\bm{y})) I(\bm{y})$. Hence, employing different acceleration methods for $K_s(x)$, we can derive different fast implementations for BF under the shiftability based acceleration framework.

As for the non-shiftable range kernel, we can exploit a set of shiftable range kernels to approximate it. Following the idea, Chaudhury~\etal~\cite{Chaudhury_TIP_2011} took trigonometric functions to approximate the Gaussian range kernel as illustrated in \eqref{eq:approx_tri}.
\begin{IEEEeqnarray}{C}
\begin{split}
& G_{\sigma}(I(\bm{x})-I(\bm{y})) \approx \sum_{n=0}^N 2^{-N} {N \choose n} \cos ( \omega_n I(\bm{x}) ) \cos ( \omega_n I(\bm{y}) ) \\
&+   \sum_{n=0}^N 2^{-N} {N \choose n} \sin ( \omega_n I(\bm{x}) ) \sin ( \omega_n I(\bm{y}) )
\end{split}
\label{eq:approx_tri}
\end{IEEEeqnarray}
where $\omega_n = \frac{2n-N}{\sqrt{N} \sigma}$.
Let $H_n(x)$ be an Hermite polynomial with order $n$, Dai~\etal~\cite{Dai_EL_2014} employed $H_n(x)$ to approximate the Gaussian range kernel as illustrated in~\eqref{eq:approx_her}.
\begin{IEEEeqnarray}{C}
\begin{split}
& G_{\sigma}(I(\bm{x})-I(\bm{y})) \approx e^{-\frac{I^2(\bm{x})}{\sigma}} \sum_{n=0}^N  H_n(\frac{I(\bm{x})}{\sigma}) \frac{I^n(\bm{y})}{n ! \sigma^{\frac{n}{2}} }
\end{split}
\label{eq:approx_her}
\end{IEEEeqnarray}

Compared with other range kernel acceleration methods, the two methods are limited to the Gaussian kernel. Another drawback shared by the two methods is reported by Chaudhury~\cite{Chaudhury_TIP_2013} who complained that it is difficult to approximate the Gaussian range kernel using above expansions when $\sigma$ is small. In particular, a great deal of approximation terms are required to get a good approximation of a narrow Gaussian function. This will considerably increase the run time.

\section{Background Techniques}
\label{sec:techniques}

In this section, we will provide a brief introduction for the necessary background knowledge required by our method.

\subsection{Best $N$-term approximation}

Given a set of orthogonal basis functions $\varphi_k$ of a $L^2$ space $V$ with the 2 norm $\| \cdot \|_2$, a best $N$ term approximation $g_{\Lambda}$ to a function $f \in V$ minimizes
\begin{IEEEeqnarray}{C}
\begin{split}
&\min_{g_{\Lambda}}   \| f - g_{\Lambda} \|_2  \\
&\text{s.t.} \quad g_{\Lambda} = \sum_{k \in \Lambda } c_k \varphi_{k}
\end{split}
\label{eq:N_term}
\end{IEEEeqnarray}
where $\Lambda$ denotes the index set formed by $N$ functions $\varphi_k$ and the space of $g_{\Lambda}$, in which the approximation is sought, is the nonlinear manifold consisting of all linear combinations of the given bases with at most $N$ terms.

Without solving optimization~\eqref{eq:N_term}, the best $N$-term approximation of the function $f$ can be obtained by selecting the first $N$ largest coefficients, because the orthogonal projection $g_{\Lambda}$ of $f$ on the space $V_{\Lambda}$ spanned by $\varphi_k, k \in \Lambda$ is
\begin{IEEEeqnarray}{C}
g_{\Lambda} = \sum_{k \in \Lambda} \langle f, \varphi_k \rangle \varphi_k
\end{IEEEeqnarray}
Hence, the approximation error can be rewritten as $\| f - g_{\Lambda} \|^2_2 = \sum_{k \not\in \Lambda} |\langle f, \varphi_k  \rangle|^2 = \| f \|^2_2 - \sum_{k \in \Lambda} |\langle f, \varphi_k  \rangle|^2$ which indicates our conclusion.

Selecting the $N$ largest coefficients provides us a simple way to find the best $N$-term approximation. In the following sections, we use this strategy to find the best $N$-term approximations of the range kernel $K_r(x)$ and the spatial kernel $K_s(x)$ on 1-D truncated trigonometric functions and 2-D Haar functions respectively.

\subsection{Truncated trigonometric functions}

Truncated trigonometric functions form the basis function of the $L^2$ space on the interval $[-T, T]$. For arbitrary functions $f \in  L^2([-T, T])$, we have
\begin{IEEEeqnarray}{C}
f(x) = \dot{B}(x)(\sum_{j=0}^{\infty} a_j \cos(\frac{\pi j}{T} x) + \sum_{j=1}^{\infty} b_j \sin(\frac{\pi j}{T} x))
\label{eq:1D_truncated_cos}
\end{IEEEeqnarray}
where $\dot{B}(x)$ denotes a 1-D box function with the support region $[-T, T]$ and
\begin{IEEEeqnarray}{ll}
a_0 &= \frac{1}{2T} \int_{-T}^{T} f(x) dx \\
a_j &= \frac{1}{T} \int_{-T}^{T} f(x) \cos(\frac{\pi j}{T} x) dx \\
b_j & = \frac{1}{T} \int_{-T}^{T} f(x) \sin(\frac{\pi j}{T} x) dx
\end{IEEEeqnarray}

Above equations are the Fourier series which is constrained on the interval $[-T, T]$. \eqref{eq:1D_truncated_cos} holds for the reason that Fourier series can decompose a periodic function into the sum of a (possibly infinite) set of simple oscillating functions, namely sines and cosines. If we define $\hat{f}$ as the periodic extension of $f$ and compute its Fourier periodic expansion, we always have $f(x) = \hat{f}(x)= \sum_{j=0}^{\infty} a_j \cos(\frac{\pi j}{T} x) + \sum_{j=1}^{\infty} b_j \sin(\frac{\pi j}{T} x)$ on the interval $[-T, T]$.

\subsection{Haar functions}

%
%

\begin{figure}

\begin{subfigure}[b]{0.24\linewidth}
\includegraphics[width=\textwidth]{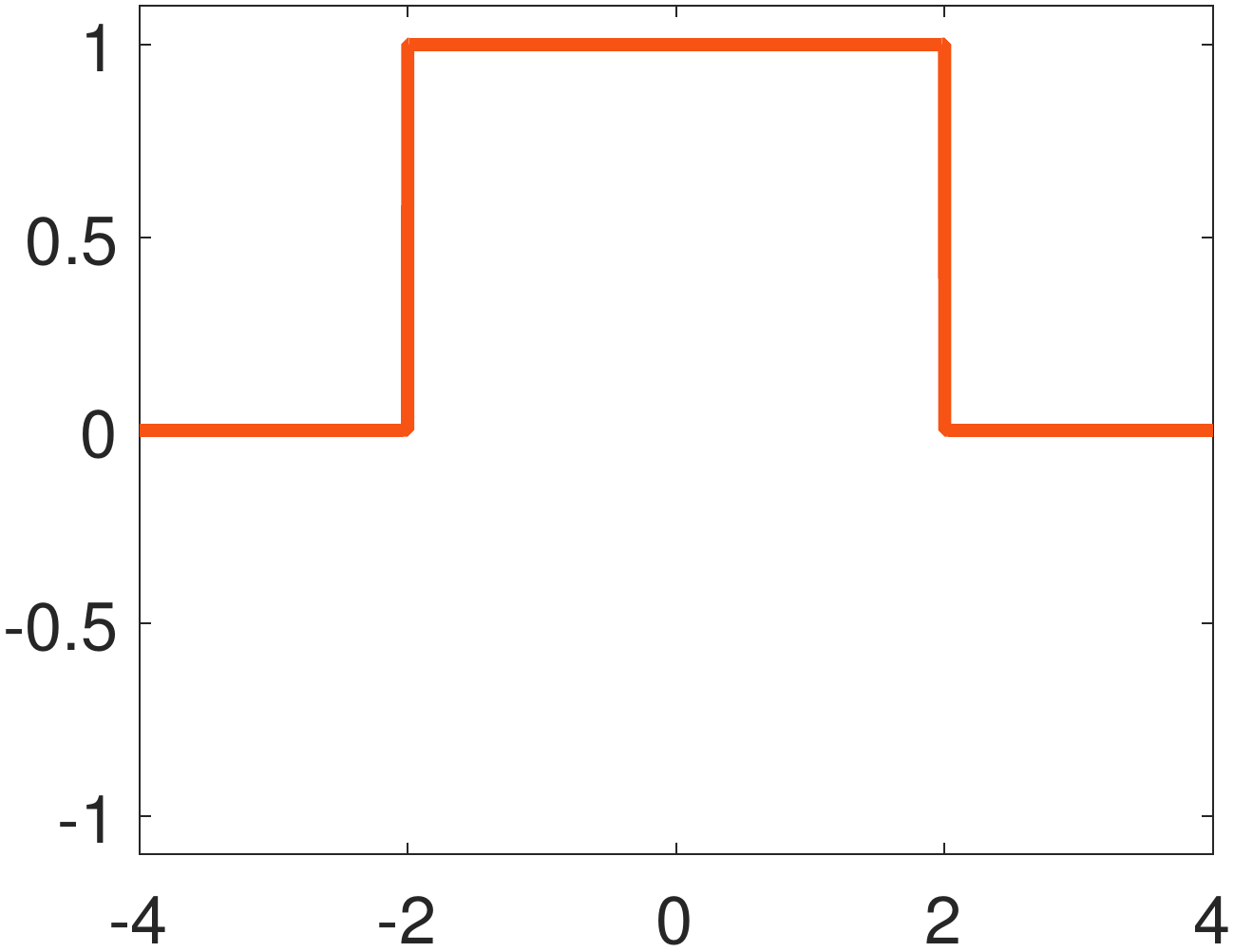}
\caption{$\phi(x)$}
\label{fig:Haar_1D:phi}
\end{subfigure}
\begin{subfigure}[b]{0.24\linewidth}
\includegraphics[width=\textwidth]{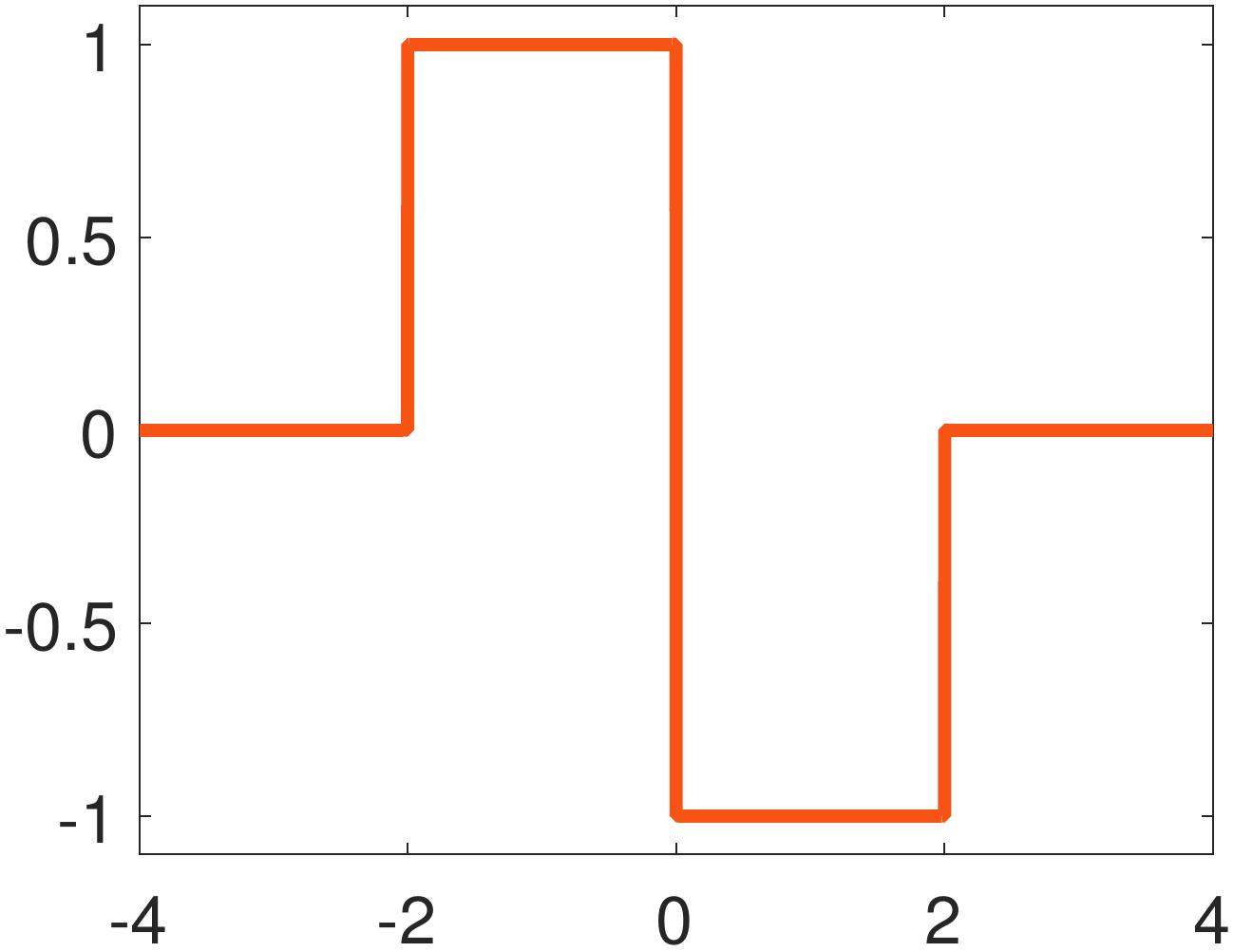}
\caption{$\psi_{0,0}(x)$}
\label{fig:Haar_1D:psi00}
\end{subfigure}
\begin{subfigure}[b]{0.24\linewidth}
\includegraphics[width=\textwidth]{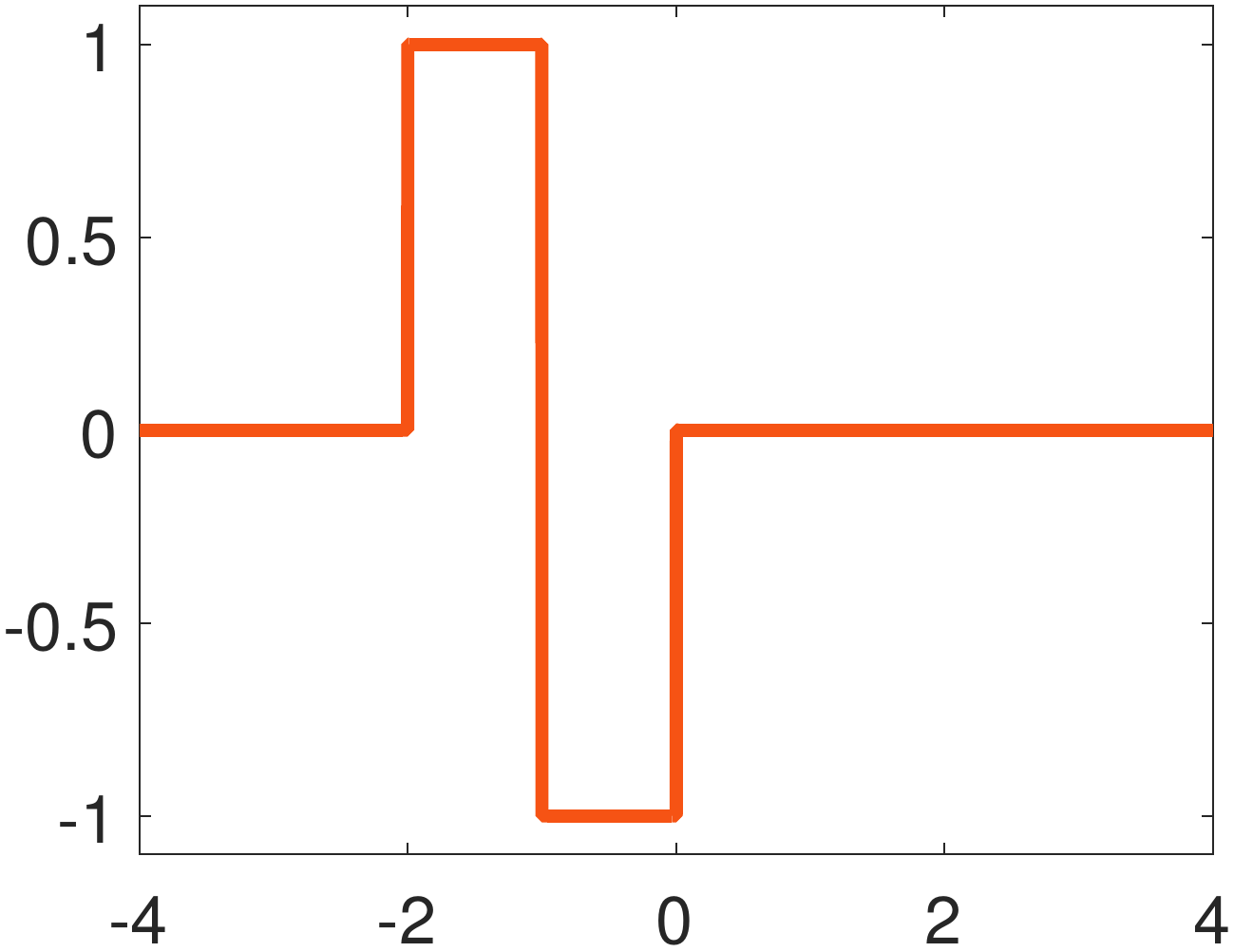}
\caption{$\psi_{1,-1}(x)$}
\label{fig:gull}
\end{subfigure}
\begin{subfigure}[b]{0.24\linewidth}
\includegraphics[width=\textwidth]{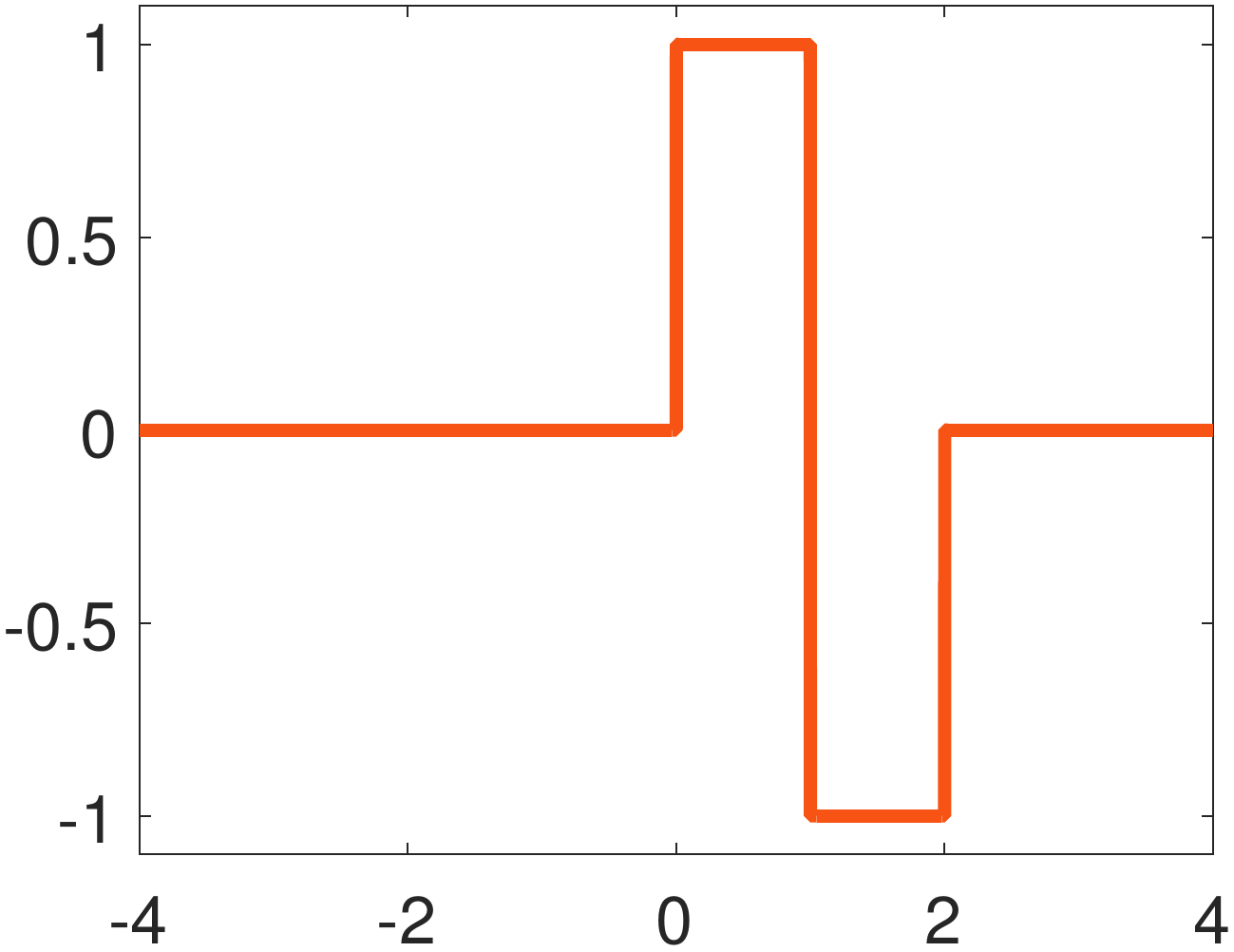}
\caption{$\psi_{1,1}(x)$}
\label{fig:gull}
\end{subfigure}

\begin{subfigure}[b]{0.24\linewidth}
\includegraphics[width=\textwidth]{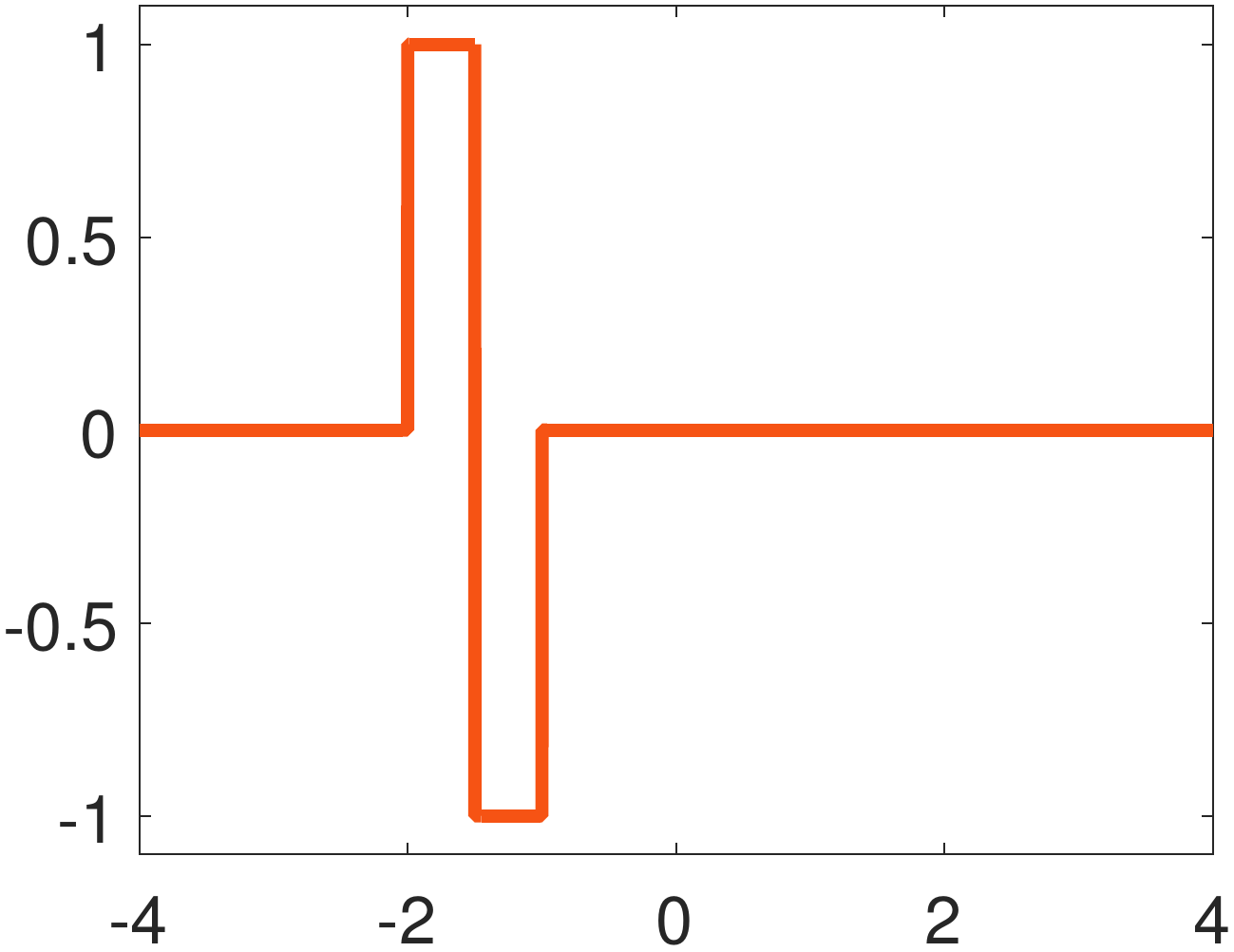}
\caption{$\psi_{2,-2}(x)$}
\label{fig:gull}
\end{subfigure}
\begin{subfigure}[b]{0.24\linewidth}
\includegraphics[width=\textwidth]{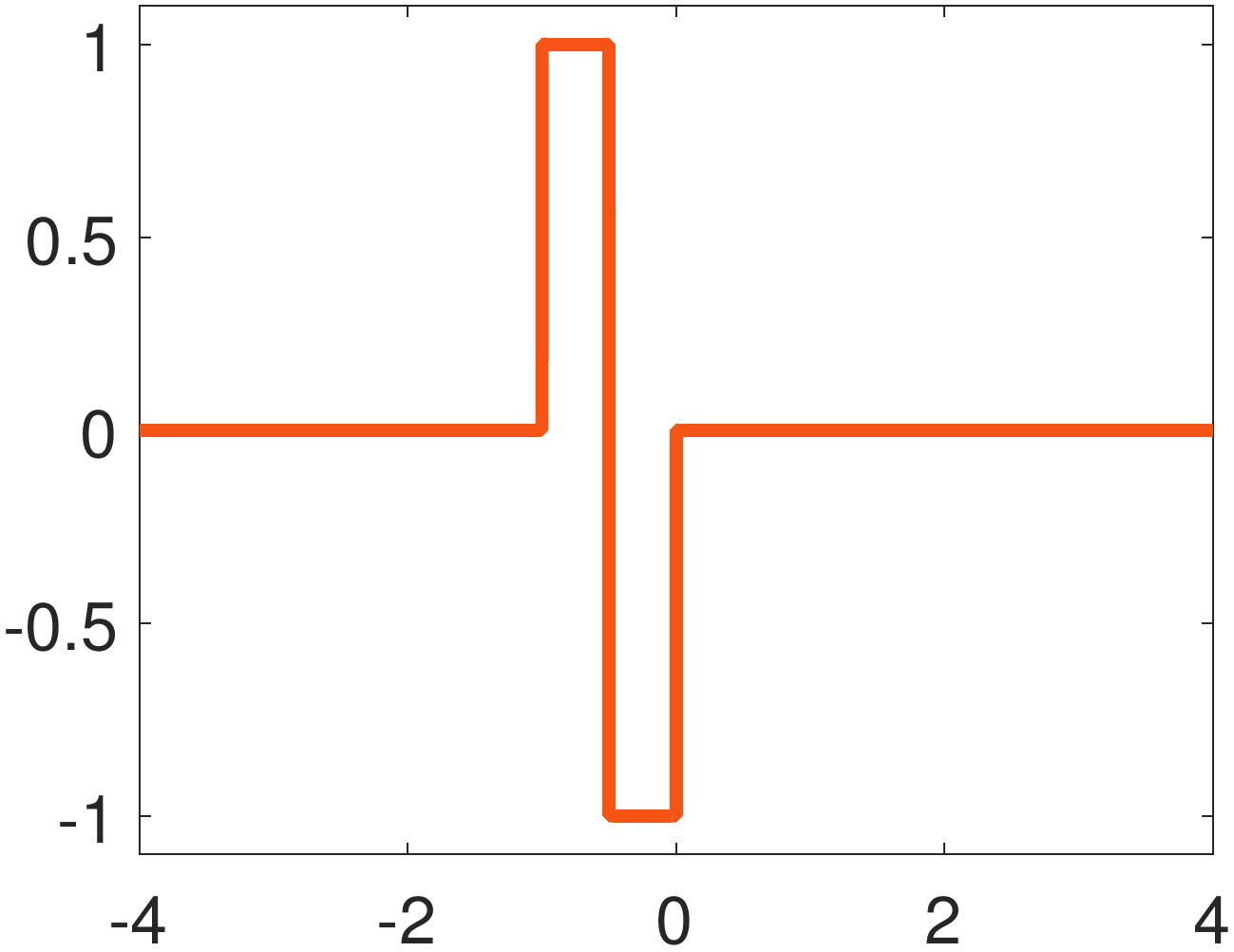}
\caption{$\psi_{2,-1}(x)$}
\label{fig:gull}
\end{subfigure}
\begin{subfigure}[b]{0.24\linewidth}
\includegraphics[width=\textwidth]{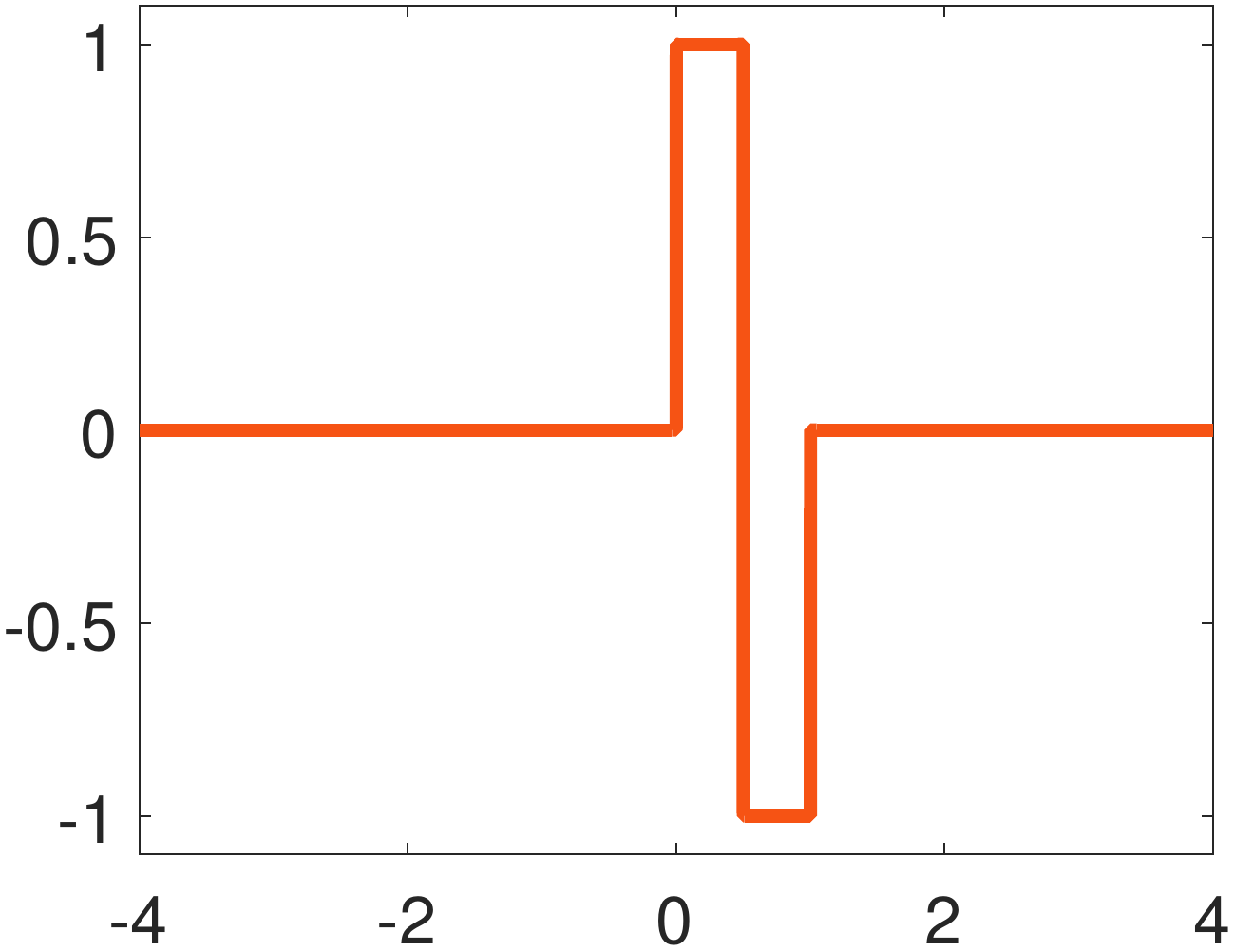}
\caption{$\psi_{2,1}(x)$}
\label{fig:gull}
\end{subfigure}
\begin{subfigure}[b]{0.24\linewidth}
\includegraphics[width=\textwidth]{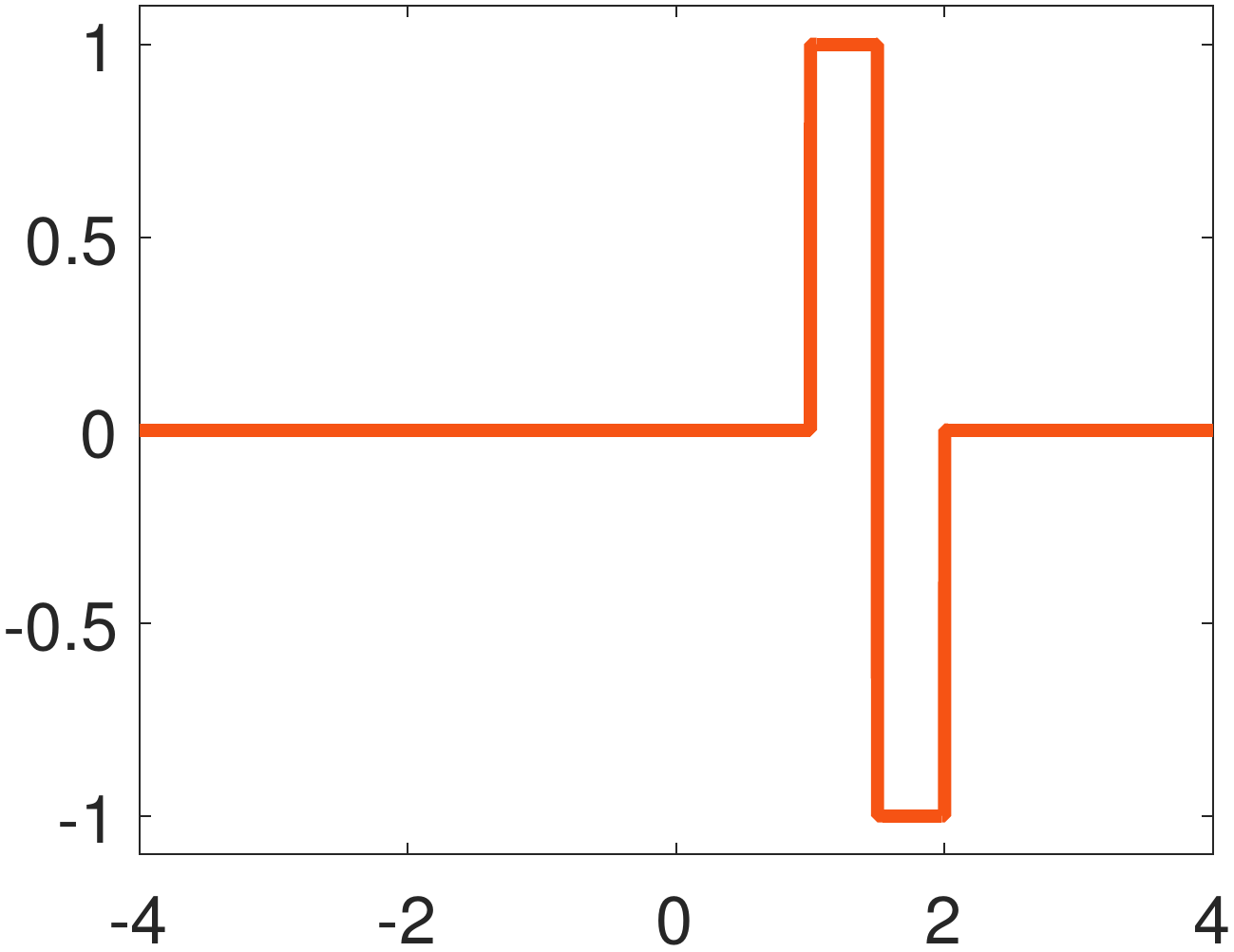}
\caption{$\psi_{2,2}(x)$}
\label{fig:Haar_1D:psi23}
\end{subfigure}
\caption{A visual illustration of the Haar functions $\phi(x)$, $\psi_{0,0}(x)$, $\psi_{1,-1}(x)$, $\psi_{1,1}(x)$, $\psi_{2,-2}(x)$, $\psi_{2,-1}(x)$, $\psi_{2,1}(x)$ and $\psi_{2,2}(x)$, where $T = 2$}
\label{fig:Haar_1D}
\end{figure}

Haar functions were proposed by Haar~\cite{Haar_MA_1910} in 1910 to give an example of the orthonormal system of square-integrable functions on the unit interval. In general, Haar functions are a sequence of rescaled ``square-shaped" functions. Specifically, the Haar scaling function $\phi(x)$ is defined as
\begin{IEEEeqnarray}{C}
\phi(x) =
\begin{cases}
1 & \quad  -T \leq x \leq T \\
0& \quad \text{otherwise}
\end{cases}
\label{eq:Haar_scaling}
\end{IEEEeqnarray}
and the Haar mother function $\psi_{0,0}(x)$ is described as
\begin{IEEEeqnarray}{C}
\psi_{0,0}(x) =
\begin{cases}
1& \quad -T \leq x < 0   \\
-1& \quad 0 \leq x \leq T \\
0& \quad \text{otherwise}
\end{cases}
\label{eq:Haar_mother}
\end{IEEEeqnarray}

Let $\mathcal{K}_j = \{-2^{j-1}, \cdots ,-1, 1, \cdots , 2^{j-1} \} $ and  $j$ be nonnegative integers. For an integer $k \in \mathcal{K}_j $,  the Haar function $\psi_{j,k}(x)$ is defined by the formula
\begin{IEEEeqnarray}{C}
\psi_{j,k}(x) = \psi_{0,0}(2^j x - \sign(k) (2|k| - 1) T )
\label{eq:Haar_set}
\end{IEEEeqnarray}
So, for example, Fig.~\ref{fig:Haar_1D} illustrates $\phi(x)$ as well as the first few values of $\psi_{j,k}(x)$, where
\begin{IEEEeqnarray}{lll}
&\psi_{1,-1}(x) = \psi_{0,0}(2x+T) \quad& \psi_{1,1}(x) = \psi_{0,0}(2x-T) \IEEEnonumber\\
&\psi_{2,-1}(x) = \psi_{0,0}(4x+T) \quad& \psi_{2,1}(x) = \psi_{0,0}(4x-T) \IEEEnonumber\\
&\psi_{2,-2}(x) = \psi_{0,0}(4x+3T) \quad& \psi_{2,2}(x) = \psi_{0,0}(4x-3T) \IEEEnonumber
\end{IEEEeqnarray}
All these functions are called Haar functions and form an orthogonal basis of $L^2([-T, T])$. Then arbitrary functions $f(x) \in L^2([-T, T])$ can be written as a series expansion by
\begin{IEEEeqnarray}{C}
f(x) = c_0 \phi(x) + \sum_{j=0}^{\infty}\sum_{k \in \mathcal{K}_j} c_{jk} \psi_{j,k}(x)
\label{eq:Haar_decomposition}
\end{IEEEeqnarray}
where $c_0 = \frac{\int_{-T}^T f(x) \phi(x) dx}{2T} $ and $c_{jk} = \frac{2^j \int_{-T}^T f(x) \psi_{j,k}(x) dx}{2T} $.

2-D Haar functions are a natural extension from the single dimension case. For any orthogonal basis $\varphi_k \in L^2([-T, T])$, one can associate a separable orthogonal basis $\varphi_{k_1}(x)\varphi_{k_2}(y)$ of $L^2([-T, T] \times [-T, T])$. Following the strategy, we define 2-D Haar functions as the set of $\{ \phi(x)\phi(y),   \phi(x)\psi_{j_2,k_2}(y),  \psi_{j_1,k_1}(x)\phi(y),  \psi_{j_1,k_1}(x)\psi_{j_2,k_2}(y)\}$. Then for arbitrary functions $f(\bm{x}) \in L^2([-T, T] \times [-T, T])$, we have
\begin{IEEEeqnarray}{C}
\begin{split}
f(\bm{x}) & = c_0 \phi(x)\phi(y) \\
& + \sum_{j_2 =0}^{\infty}\sum_{k_2 \in \mathcal{K}_{j_2}} c_{0, j_2,k_2} \phi(x)\psi_{j_2,k_2}(y) \\
& + \sum_{j_1 =0}^{\infty}\sum_{k_1 \in \mathcal{K}_{j_1}} c_{j_1,k_1,0} \psi_{j_1,k_1}(x) \phi(y) \\
& + \sum_{j_1 =0}^{\infty}\sum_{k_1 \in \mathcal{K}_{j_1}}\sum_{j_2 =0}^{\infty}\sum_{k_2 \in \mathcal{K}_{j_2}}   c_{j_1,k_1,j_2,k_2} \psi_{j_1,k_1}(x)\psi_{j_2,k_2}(y)
\end{split}
\label{eq:Haar_expansion}
\end{IEEEeqnarray}
where $\bm{x} = (x,y)$ and
\begin{IEEEeqnarray}{C}
\begin{split}
& c_0 = \frac{1}{4T^2}\int_{-T}^T \int_{-T}^T f(x,y) \phi(x)\phi(y) dxdy \\
& c_{0,j_2,k_2} = \frac{2^{j_2}}{4T^2}\int_{-T}^T \int_{-T}^T f(x,y) \phi(x)\psi_{j_2,k_2}(y) dxdy \\
& c_{j_1,k_1,0} = \frac{2^{j_1}}{4T^2}\int_{-T}^T \int_{-T}^T f(x,y) \psi_{j_1,k_1}(x)\phi(y) dxdy \\
& c_{j_1,k_1,j_2,k_2} = \frac{2^{j_1}2^{j_2}}{4T^2}\int_{-T}^T \int_{-T}^T f(x,y) \psi_{j_1,k_1}(x)\psi_{j_2,k_2}(y) dxdy
\end{split}
\label{eq:Haar_2D_coefficients}
\end{IEEEeqnarray}

\subsection{The Summed Area Table (SAT)}

As a data structure for quickly and efficiently generating the sum of values in a rectangular subset of a grid, the 2-D SAT was first introduced to computer graphics society in 1984 by Crow~\cite{Crow_SIGGRAPH_1984} for texture mapping, but was not properly introduced to the world of computer vision till 2004 by Viola and Jones~\cite{Viola_IJCV_2004} with their landmark face detection algorithm to fast compute the sum of image values $I(x', y')$ on a given rectangle $\mathfrak{R} = (x_0, x_1] \times (y_0, y_1]$.
\begin{IEEEeqnarray}{C}
S(\mathfrak{R}) = \sum_{x_0 < x' \leq x_1} \sum_{y_0 < y' \leq y_1} I(x', y')
\label{eq:sum_region_2D}
\end{IEEEeqnarray}

The complexity of \eqref{eq:sum_region_2D} is proportional to the size of the rectangle $\mathfrak{R}$. Viola and Jones employ the SAT $\mathbb{S}(x, y)$ in~\eqref{eq:sum_sat_2D} to compute arbitrary $S$ in constant time.
\begin{IEEEeqnarray}{C}
\mathbb{S}(x, y) = \sum_{0 \leq x' \leq x} \sum_{0 \leq y' \leq y} I(x', y')
\label{eq:sum_sat_2D}
\end{IEEEeqnarray}

First, SAT can be calculated in one pass over the image $I$ by putting $c(x, -1) = \mathbb{S}(-1, y) = 0$ and using the recurrence~\eqref{eq:sum_sat_recurrence_2D}
\begin{IEEEeqnarray}{lC}
\begin{split}
&c(x, y) = c(x, y-1) + I(x, y) \\
&\mathbb{S}(x, y) = \mathbb{S}(x-1, y) + c(x, y)
\end{split}
\label{eq:sum_sat_recurrence_2D}
\end{IEEEeqnarray}

After that, $S(\mathfrak{R})$ can be computed in linear complexity over the image $I$ according to
\begin{IEEEeqnarray}{lC}
S(\mathfrak{R}) = \mathbb{S}(x_1, y_1) - \mathbb{S}(x_1, y_0) - \mathbb{S}(x_0, y_1) + \mathbb{S}(x_0, y_0)
\label{eq:sum_sat_sum_region_2D}
\end{IEEEeqnarray}
which only contains four references to $\mathbb{S}(x, y)$.

A special case of the 2-D SAT is the 1-D SAT defined by $\mathbb{S}(x) = \sum_{0 \leq x' \leq x} I(x')$. We can computed it in one pass according to $\mathbb{S}(x) = \mathbb{S}(x-1) + I(x)$. The sum $S((x_0, x_1])$ on the interval $(x_0, x_1]$ thus equals to
\begin{IEEEeqnarray}{lC}
S((x_0, x_1]) = \mathbb{S}(x_1) - \mathbb{S}(x_0)
\label{eq:sat_sum_1D}
\end{IEEEeqnarray}

The 2-D SAT can also be extended to the high-dimensional case to compute the sum of values in a $N$-D cube. In the literature, Ke~\etal~\cite{Ke_ICCV_2005} considered the image sequences as three-dimensional images and defined the integral video (\ie the 3-D SAT) to compute volumetric features from the optical flow of videos for the motion and activity detection. Six years later, Tapia~\cite{Tapia_PRL_2011} provided a generalized procedure to compute the sum of values in a $N$ dimensional cube using the $N$-D SAT. Here, instead of directly computing the 3-D SAT, we jointly employ the 1-D SAT and the 2-D SAT to compute the sum $S(\mathfrak{C}) = \sum_{x_0 < x' \leq x_1} \sum_{y_0 < y' \leq y_1} \sum_{z_0 < z' \leq z_1} I(x', y', z')$ of a 3-D image $I(x', y', z')$ on a given cube $\mathfrak{C} = (x_0, x_1] \times (y_0, y_1] \times (z_0, z_1] $. This is because we have
\begin{IEEEeqnarray}{rl}
\bar{I}(x, y, z) & = \sum_{x_0 < x' \leq x_1} \sum_{y_0 < y' \leq y_1} I(x', y', z) \label{eq:sat_sum_3D_1}\\
S(\mathfrak{C}) & = \sum_{z_0 < z' \leq z_1} \bar{I}(x, y, z') \label{eq:sat_sum_3D_2}
\end{IEEEeqnarray}
and for each given $z$, we can employ the 2-D SAT in \eqref{eq:sum_sat_sum_region_2D} to fast compute $\bar{I}(x, y, z)$. Similarly, \eqref{eq:sat_sum_3D_2} can be fast calculated by the 1-D SAT in \eqref{eq:sat_sum_1D}.


\section{Proposed Method}
\label{sec:method}

In order to decompose BF into a set of 3-D box filters while keeping high accuracy, our method employs Haar functions and truncated trigonometric functions to compute the best N-term approximation of the truncated spatial and range kernels.
Further analysis discloses that our approximation can be computed by the 3-D SAT with complexity $O(|I|)$.

\subsection{Truncated spatial kernel $K^T_s(x)$ and range kernel $K^T_r(x)$ }

Since $K_s(x)$ and $K_r(x)$ are symmetric on the region $\mathbb{R}$ and decrease their values on the region $\mathbb{R}^+$, a point $x$ with large values $K_s(x)$ or $K_r(x)$ usually locates at a small region around the original point. For instance, the Gaussian kernel $G_{\sigma}(x)$ falls off very fast, and almost vanishes outside the interval $[-3\sigma, 3\sigma]$. Hence, we can simply discard the points outside $[-3\sigma, 3\sigma]$ without introducing significant errors. It is thus reasonable to substitute  $K_s(x)$ and $K_r(x)$  with the truncated kernels $K^T_s(x)$ and $K^T_r(x)$ (refer to sections~\ref{sec:comparison_Ks} and \ref{sec:comparison_Kr}). Here $K^T_s(x)$ and $K^T_r(x)$ equal to $K_s(x)$ and $K_r(x)$ on the intervals $[-T_s, T_s]$ and $[-T_r, T_r]$ respectively, otherwise $K^T_s(x) = K^T_r(x) = 0$, where $T_s = K^{-1}_s(\epsilon)$, $T_r = K^{-1}_r(\epsilon)$, and $\epsilon$ is a predefined value. In practice, $0.01$ is a reasonable value.


Note that the truncation operation for $K^T_s(x), K^T_r(x)$ does not increase the run time of BF because we can precompute the truncated regions $[-T_s, T_s]$ or $[-T_r, T_r]$. In the sequent sections, we will describe the methods to approximate the truncated kernels.

\subsection{Approximation for the linear convolution of $K_s( x )$}
\label{sec:Ks}

\begin{figure}
\centering
\begin{subfigure}[b]{0.45\linewidth}
\includegraphics[width=\textwidth]{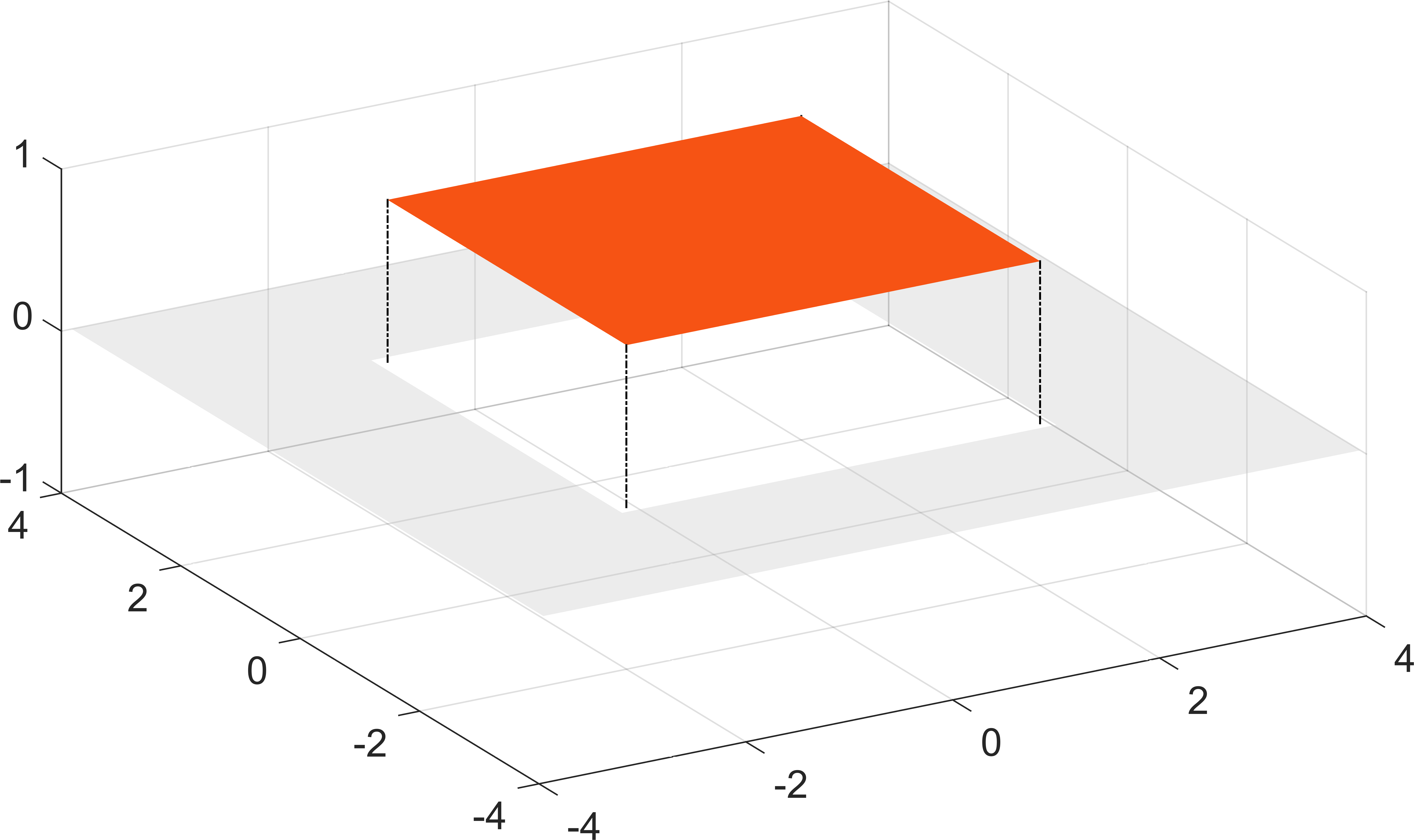}
\caption{$\phi(x)\phi(y)$}
\label{fig:Haar_1D:phi}
\end{subfigure}
\begin{subfigure}[b]{0.45\linewidth}
\includegraphics[width=\textwidth]{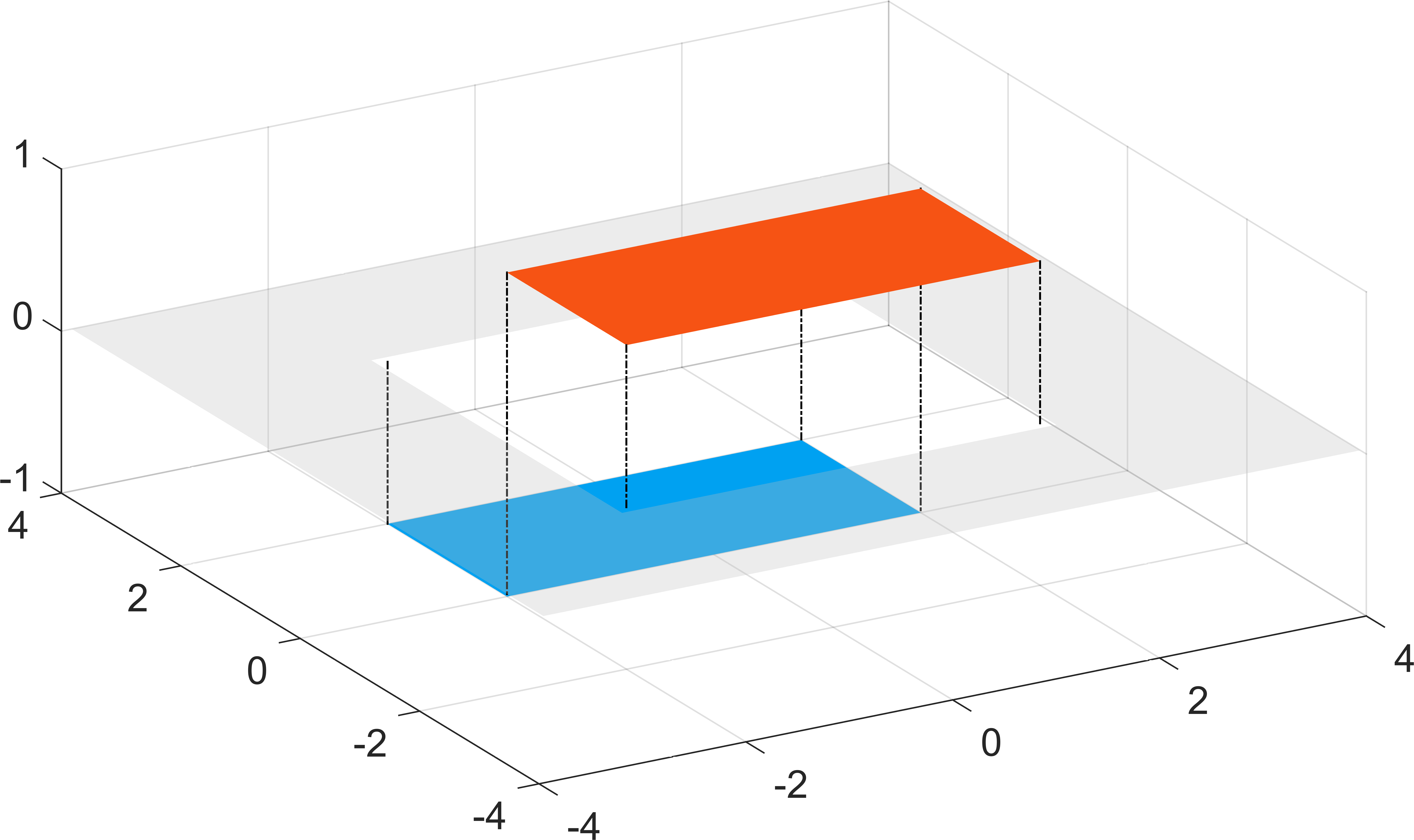}
\caption{$\phi(x)\psi_{0,0}(y)$}
\label{fig:Haar_1D:psi00}
\end{subfigure}

\begin{subfigure}[b]{0.45\linewidth}
\includegraphics[width=\textwidth]{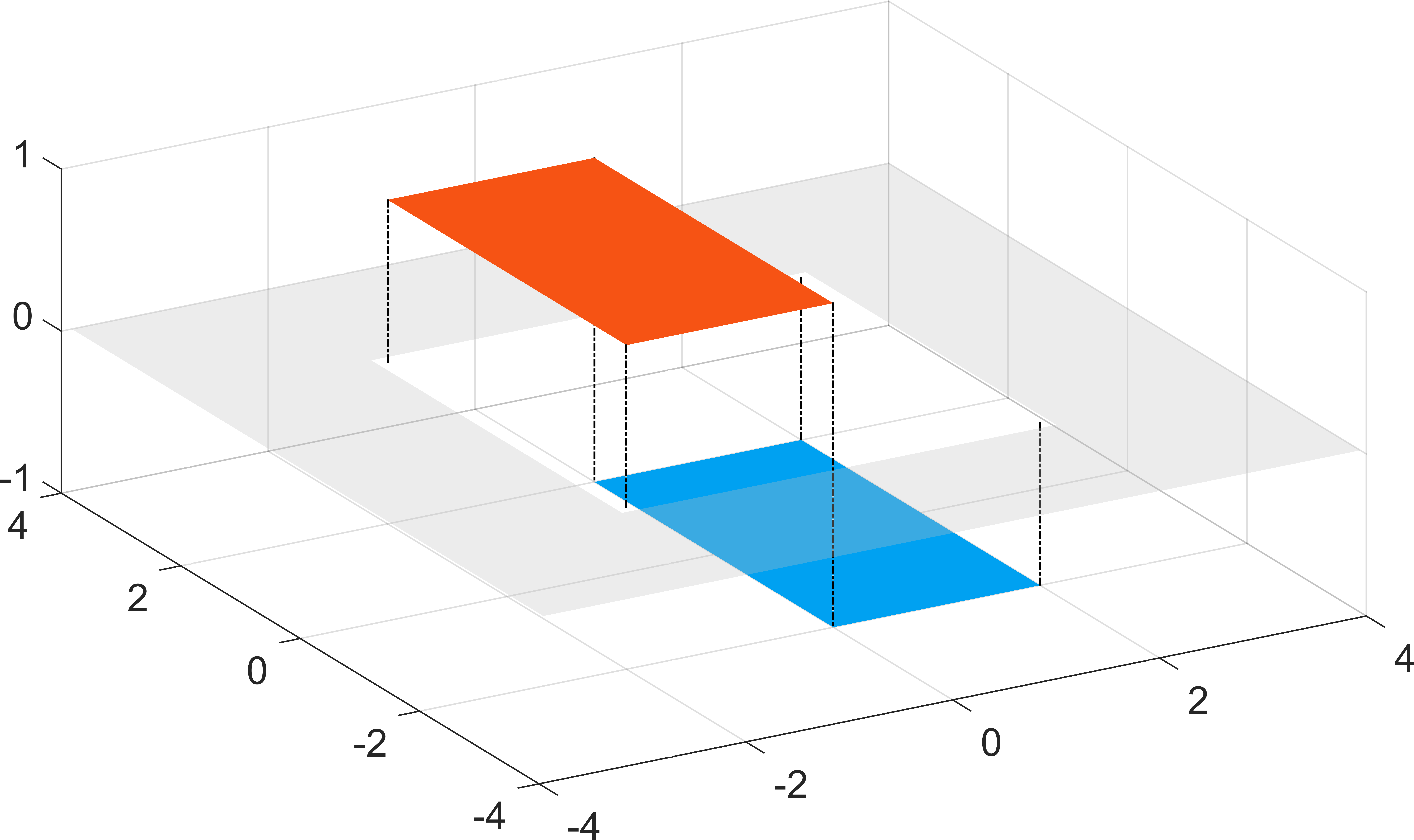}
\caption{$\psi_{0,0}(x)\phi(y)$}
\label{fig:gull}
\end{subfigure}
\begin{subfigure}[b]{0.45\linewidth}
\includegraphics[width=\textwidth]{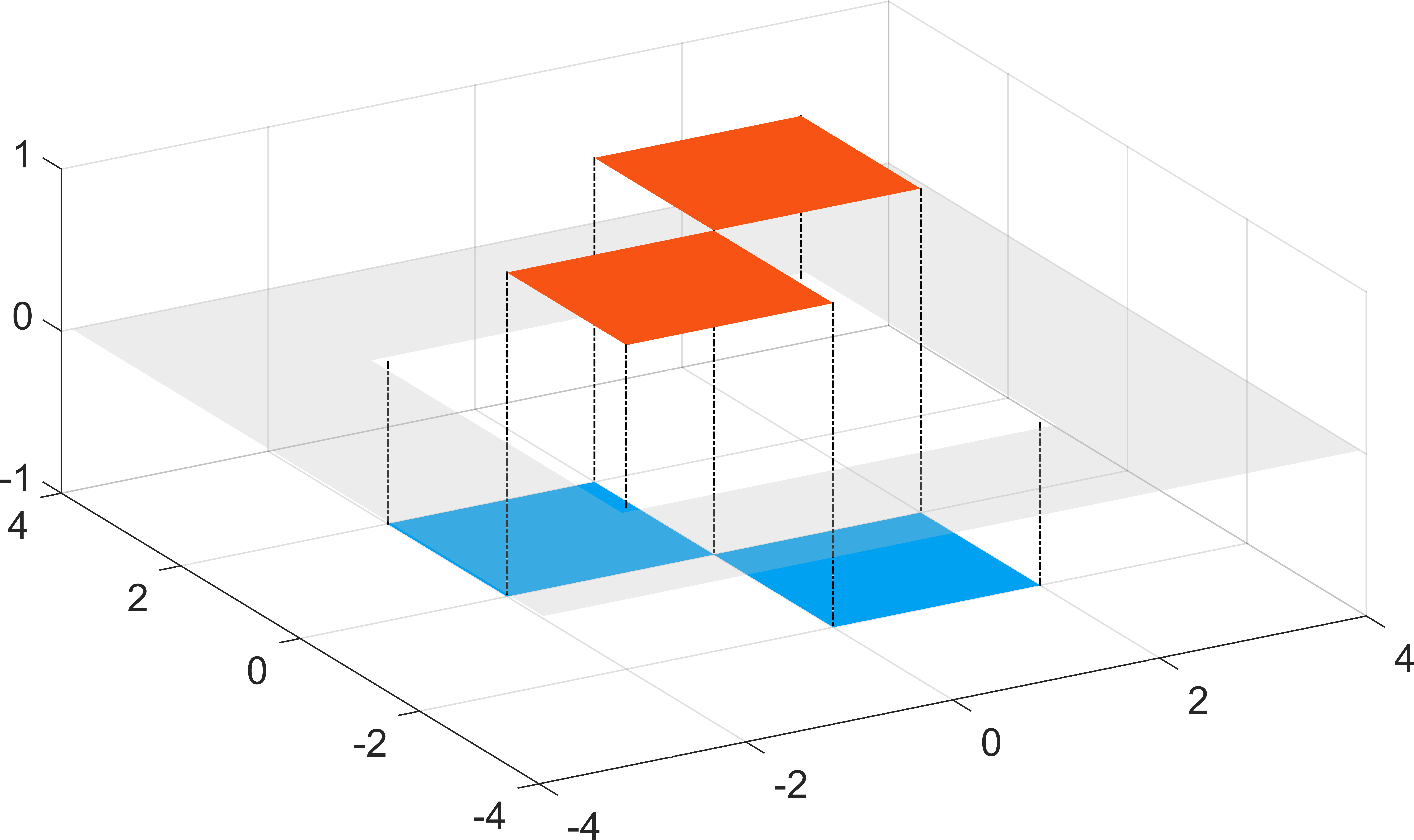}
\caption{$\psi_{0,0}(x)\psi_{0,0}(y)$}
\label{fig:Haar_1D:psi23}
\end{subfigure}
\caption{A visual illustration of 2-D Haar functions $\phi(x)\phi(y)$, $\phi(x)\psi_{0,0}(y)$, $\psi_{0,0}(x)\phi(y)$ and $\psi_{0,0}(x)\psi_{0,0}(y)$, where  $T = 2$, the red, gray and blue planes denote the values of $1$, $0$ and $-1$, respectively.}
\label{fig:Haar_2D}
\end{figure}

In computer vision and computer graphics~\cite{Crow_SIGGRAPH_1984,Tuzel_ECCV_2009,Li_TPAMI_2014}, the box filter has been used to accelerate many computation-intensive applications as it has the advantage of being fast to compute, but its adoption has been hampered by the fact that it presents serious restrictions to filter construction.
To solve the problem, we employ 2-D Haar functions to transform arbitrary spatial kernels to a set of box functions.



\begin{figure*}[t]
    \includegraphics[width=\textwidth]{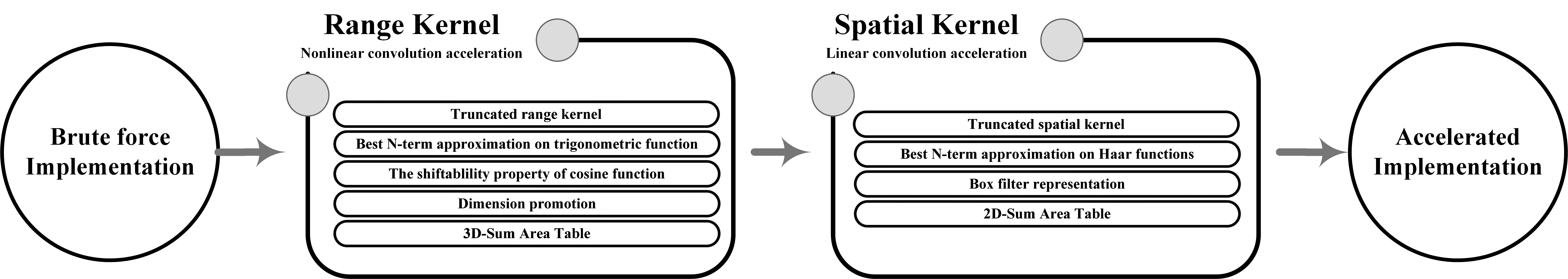}
    \caption{A compressive illustration for the flowchart of our two steps acceleration algorithm as well as the acceleration techniques used in each step. Through the nonlinear convolution acceleration step containing five techniques and the linear convolution acceleration step assembled by four techniques, our algorithm successfully reduces the computational complexity from $O(|\mathcal{N}_x||I|)$ to $O(|I|)$.  }
    \label{fig:flowchart}
\end{figure*}

As the truncated substitution $K^T_s(\|\bm{x}\|)$ of $K_s(\|\bm{x}\|)$ is a 2-D function, we can employ \eqref{eq:Haar_expansion} to decompose it into a set of 2-D Haar functions which are defined on the region $[-T_s, T_s] \times [-T_s, T_s]$. After that, we select the first $N$ largest coefficients from the set $\{c_0\} \bigcup \{c_{0,j_2, k_2}\} \bigcup \{ c_{j_1, k_1, 0} \} \bigcup \{ c_{j_1, k_1, j_2, k_2} \}$. Let $\Lambda^s_1$, $\Lambda^s_2$, $\Lambda^s_3$ and $\Lambda^s_4$ denote the selected coefficients from sets $\{c_0\}, \{c_{0,j_2, k_2}\}, \{ c_{j_1, k_1, 0} \}, \{ c_{j_1, k_1, j_2, k_2} \}$, so we have
%
\begin{IEEEeqnarray}{C}
\begin{split}
& K_s(\|\bm{x}\|) \underset{\text{\ding{192}}}{\approx} K^T_s(\|\bm{x}\|) \underset{\text{\ding{193}}}{\approx} \sum_{c_0 \in \Lambda^s_1} c_0 \phi(x)\phi(y) \\
+ & \sum_{c_{0, j_2,k_2} \in \Lambda^s_2}  c_{0, j_2,k_2} \phi(x)\psi_{j_2,k_2}(y) \\
+ & \sum_{c_{j_1,k_1,0} \in \Lambda^s_3} c_{j_1,k_1,0} \psi_{j_1,k_1}(x) \phi(y) \\
+ & \sum_{c_{j_1,k_1,j_2,k_2} \in \Lambda^s_4} c_{j_1,k_1,j_2,k_2} \psi_{j_1,k_1}(x)\psi_{j_2,k_2}(y)
\end{split}
\label{eq:Haar_N_terms}
\end{IEEEeqnarray}

The equation can be simplified further as the basis functions $\{\phi(x)\phi(y), \phi(x)\psi_{j_2,k_2}(y), \psi_{j_1,k_1}(x) \phi(y), \psi_{j_1,k_1}(x)\psi_{j_2,k_2}(y) \}$ can be effectively represented by 2-D box functions. Fig.~\ref{fig:Haar_2D} provides us a visual illustration of four 2-D Haar functions $\phi(x)\phi(y)$, $\phi(x)\psi_{0,0}(y)$, $\psi_{0,0}(x)\phi(y)$, $\psi_{0,0}(x)\psi_{0,0}(y)$ which stand for the four sets $\{\phi(x)\phi(y)\}$, $\{\phi(x)\psi_{j_2,k_2}(y)\}$, $\{\psi_{j_1,k_1}(x) \phi(y)\}$, $\{\psi_{j_1,k_1}(x)\psi_{j_2,k_2}(y) \}$ respectively. From the figure, we can verify that $\phi(x)\phi(y)$ is a 2-D box filter with the support region $[-T_s, T_s] \times [-T_s, T_s]$, and $\phi(x)\psi_{0,0}(y)$ can be represented by two 2-D box filters located at $[-T_s, T_s] \times [-T_s, 0]$ and $[-T_s, T_s] \times [0, T_s]$. Similarly, $\psi_{0,0}(x)\phi(y)$ can be reformulated as two 2-D box filters and $\psi_{0,0}(x)\psi_{0,0}(y)$ is equal to four 2-D box filters. In appendix~\ref{sec:box_filter}, we prove that the conclusion can be generalized to the set $\{\phi(x)\phi(y)\}$, $\{\phi(x)\psi_{j_2,k_2}(y)\}$, $\{\psi_{j_1,k_1}(x) \phi(y)\}$, $\{\psi_{j_1,k_1}(x)\psi_{j_2,k_2}(y) \}$.
Hence $ K_s(\|\bm{x}\|)$ can be represented by a set of 2-D box filters $\ddot{B}_{\bm{j}}(\bm{x})$: 
\begin{IEEEeqnarray}{C}
K_s(\|\bm{x}\|) \underset{\text{\ding{194}}}{\approx} \sum_{c_{\bm{j}} \in \Lambda^s } c_{\bm{j}} \ddot{B}_{\bm{j}}(\bm{x})
\label{eq:Box_N_terms}
\end{IEEEeqnarray}
where $c_{\bm{j}}$ denotes coefficient of $\ddot{B}_{\bm{j}}(\bm{x})$, and $\Lambda^s$ stands for the collection of $c_{\bm{j}}$. Further, putting $N^{\bm{j}}_{\bm{x}}$ as the support region of $\ddot{B}_{\bm{j}}(\bm{x})$, we can reformulate the convolution of $  K_s(x) $ as
\begin{IEEEeqnarray}{C}
\begin{split}
& \sum_{\bm{y} \in \mathcal{N}_{\bm{x}}} K_s(\| \bm{x}  -  \bm{y} \|) g(\bm{y}) \approx
\sum_{c_{\bm{j}} \in \Lambda^s } c_{\bm{j}} \ddot{B}_{\bm{j}}(\bm{x}  -  \bm{y}) g(\bm{y}) \\
& \underset{\text{\ding{195}}}{=} \sum_{c_{\bm{j}} \in \Lambda^s } c_{\bm{j}}  \sum_{\bm{y} \in N^{\bm{j}}_{\bm{x}}}   g(\bm{y})
\end{split}
\label{eq:2D_box_N_terms_s}
\end{IEEEeqnarray}
As is known to us, the 2-D SAT can be used to compute the box filters in \eqref{eq:2D_box_N_terms_s} in constant time by using  \eqref{eq:sum_sat_sum_region_2D}.

\subsection{Approximation for the nonlinear convolution of $K_r(x)$}

In this section, we employ the best $N$-term approximation on 1-D truncated trigonometric functions to approximate the truncated range kernel $K^T_r(x)$. Let $\Lambda^r$ represent the selected coefficients from the sets $\{ a_k\}$ with $k>0$, we have
\begin{IEEEeqnarray}{C}
K_r(x) \underset{\text{\ding{196}}}{\approx} K^T_r(x) \underset{\text{\ding{197}}}{\approx} \sum_{a_k \in \Lambda^r} a_k \cos(\frac{\pi k}{T_r} x) \dot{B}(x)
\label{eq:trigonometric_N_terms}
\end{IEEEeqnarray}
Note that $b_k = 0$ for $K^T_r(x)$ due to the symmetry of $K^T_r(x)$. Employing cosine functions's shiftability property, we have
\begin{IEEEeqnarray}{C}
\begin{split}
& \sum_{\bm{y} \in \mathcal{N}_{\bm{x}}} K_s(\| \bm{x}  -  \bm{y}\|)  K_r(I(\bm{x})  -  I(\bm{y}) ) I(\bm{y})  \\
\underset{\text{\ding{198}}}{\approx}  & g^c_k(\bm{x}) \sum_{a_k \in \Lambda^r} \sum_{\bm{y} \in \mathcal{N}_{\bm{x}}}  a_k K_s(\| \bm{x}  -  \bm{y}\|) \dot{B}( I(\bm{x}) - I(\bm{y}) ) G_k^c(\bm{y}) \\
+ & g^s_k(\bm{x}) \sum_{a_k \in \Lambda^r} \sum_{\bm{y} \in \mathcal{N}_{\bm{x}}}  a_k K_s(\| \bm{x}  -  \bm{y}\|) \dot{B}(I(\bm{x}) - I(\bm{y}) ) G_k^s(\bm{y})
\end{split}
\label{eq:1D_N_terms}
\end{IEEEeqnarray}
where $G_k^c(\bm{y}) = g^c_k(\bm{y}) I(\bm{y})$, $G_k^s(\bm{y}) = g^s_k(\bm{y}) I(\bm{y})$, $g^c_k(\bm{x}) = \cos(\frac{\pi k}{T_r} I(\bm{x}))$ and $g^s_k(\bm{x}) = \sin(\frac{\pi k}{T_r} I(\bm{x}))$.
The computational complexity of \eqref{eq:1D_N_terms} depends on the size of $\mathcal{N}_{\bm{x}}$. This dependency can be eliminated by the dimension promotion technique. Let $\mathcal{N}_{I(\bm{x})}$ denote the interval $[I(\bm{x})-T_r, I(\bm{x})+T_r]$, $F_c(\bm{y}, z) = G^c_k(\bm{y}) \delta_{I(\bm{y})}(z)$, $F_s(\bm{y}, z) = G^s_k(\bm{y}) \delta_{I(\bm{y})}(z)$, we reformulate \eqref{eq:1D_N_terms} as
\begin{IEEEeqnarray}{C}
\begin{split}
& \sum_{\bm{y} \in \mathcal{N}_{\bm{x}}} K_s(\| \bm{x}  -  \bm{y} \|)  K_r(I(\bm{x})  -  I(\bm{y}) ) I(\bm{y}) \\
\underset{\text{\ding{199}}}{\approx} & g^c_k(\bm{x}) \sum_{a_k \in \Lambda^r} a_k \sum_{z \in \mathcal{N}_{I(\bm{x})}} \sum_{\bm{y} \in \mathcal{N}_{\bm{x}}} K_s(\| \bm{x}  -  \bm{y} \|) F_c(\bm{y}, z) \\
+ & g^s_k(\bm{x}) \sum_{a_k \in \Lambda^r} a_k \sum_{z \in \mathcal{N}_{I(\bm{x})}} \sum_{\bm{y} \in \mathcal{N}_{\bm{x}}} K_s(\| \bm{x}  -  \bm{y} \|)  F_s(\bm{y}, z)
\end{split}
\label{eq:1D_box_N_terms}
\end{IEEEeqnarray}
which only involves the linear convolution of $K_s(x)$.


\subsection{3-D box filter based approximation for BF }
\label{sec:3DFBF}

Now we are able to decompose the nonlinear convolution of $K_r(x)$  into a set of linear convolutions of $K_s(x)$ \eqref{eq:1D_box_N_terms} as well as to fast compute the linear convolution of $K_s(x)$ \eqref{eq:2D_box_N_terms_s}. Putting \eqref{eq:2D_box_N_terms_s} into \eqref{eq:1D_box_N_terms}, we can further transform the convolution $f(\bm{x}) =\sum_{\bm{y} \in \mathcal{N}_{\bm{x}}} K_s(\|{\bm{x} - \bm{y}}\|) K_r(I(\bm{x}) -  I(\bm{y})) I(\bm{y})$ the numerator of BF to
\begin{IEEEeqnarray}{C}
\begin{split}
& f(\bm{x})\underset{\text{\ding{200}}}{\approx}  g^c_k(\bm{x}) \sum_{a_k \in \Lambda^r} \sum_{c_{\bm{j}} \in \Lambda^s } a_k c_{\bm{j}} \sum_{z \in \mathcal{N}_{I(\bm{x})}} \sum_{\bm{y} \in N^{\bm{j}}_{\bm{x}}}  F_c(\bm{y}, z) \\
+ & g^s_k(\bm{x}) \sum_{a_k \in \Lambda^r} \sum_{c_{\bm{j}} \in \Lambda^s } a_k c_{\bm{j}} \sum_{z \in \mathcal{N}_{I(\bm{x})}} \sum_{\bm{y} \in N^{\bm{j}}_{\bm{x}}}   F_s(\bm{y}, z)
\end{split}
\label{eq:numerator_2D_box_N_terms}
\end{IEEEeqnarray}

$\sum_{z \in \mathcal{N}_{I(\bm{x})}} \sum_{\bm{y} \in N^{\bm{j}}_{\bm{x}}} F_c(\bm{y}, z)$ and $\sum_{z \in \mathcal{N}_{I(\bm{x})}} \sum_{\bm{y} \in N^{\bm{j}}_{\bm{x}}} F_s(\bm{y}, z)$ can be interpreted as the sum of $F_c(\bm{y}, z)$ and $F_s(\bm{y}, z)$ in the cube $N^{\bm{j}}_{\bm{x}} \times \mathcal{N}_{I(\bm{x})}$. Hence, \eqref{eq:numerator_2D_box_N_terms} denotes a linear combination of 3-D box filters that performed on the auxiliary images $F_c(\bm{y}, z)$, $F_s(\bm{y}, z)$. So, using \eqref{eq:sat_sum_3D_1} \eqref{eq:sat_sum_3D_2} to compute 3-D box filtering results, we can reduce the complexity of $\sum_{\bm{y} \in \mathcal{N}_{\bm{x}}} K_s(\|{\bm{x} - \bm{y}}\|) K_r(I(\bm{x}) -  I(\bm{y})) I(\bm{y})$ down to $O(|I|)$. In addition, similar discussion can be applied to BF's denominator $\sum_{\bm{y} \in \mathcal{N}_{\bm{x}}} K_s(\|{\bm{x} - \bm{y}}\|) K_r(I(\bm{x}) -  I(\bm{y}))$. Therefore, the filtering result $\hat{I}(\bm{x})$ of BF in~\eqref{eq:BF} can be figured out with linear complexity $O(|I|)$.

Finally, we plot Fig~\ref{fig:flowchart} to outline the flowchart of our algorithm as well as the acceleration techniques exploited in each step as math symbols stated above may have overshadowed the underlying ideas. Generally speaking, our acceleration algorithm can be divided into two parts (\ie the linear/nonlinear convolution acceleration steps) which employ following speeding up techniques:
\begin{enumerate}
    \item[\ding{192}] Truncated spatial kernel $K_s^T(\bm{x})$.
    \item[\ding{193}] Best $N$-term approximation for $K_s^T(\bm{x})$ on Haar functions.
    \item[\ding{194}] Box filter representation for Haar functions.
    \item[\ding{195}] 2-D SAT (2-D box filtering).
    \item[\ding{196}] Truncated range kernel $K_r^T(\bm{x})$.
    \item[\ding{197}] Best $N$-term approximation for $K_r^T(\bm{x})$ on truncated trigonometric functions
    \item[\ding{198}] The shiftability property of cosine functions.
    \item[\ding{199}] Dimension promotion.
    \item[\ding{200}] 3-D SAT (3-D box filtering).
\end{enumerate}

Specifically, in the linear convolution acceleration step for the spatial kernel, Eq~\eqref{eq:Haar_N_terms} takes the truncated spatial kernel $K_s^T(\bm{x})$ and best $N$-term approximation for $K_s^T(\bm{x})$ on Haar functions to approximate original spatial kernel $K_s(\bm{x})$ in steps \ding{192} \ding{193}, respectively. Moreover, step \ding{194} in \eqref{eq:Box_N_terms} holds due to the reason that Haar functions can be represented by the linear combination of box filters as Fig~\ref{fig:Haar_2D} indicated. Finally, applying 2-D SAT to step \ding{195}, we can figure out \eqref{eq:2D_box_N_terms_s} in linear computational complexity. Next, we accelerate the nonlinear convolution of the range kernel. Similar to the linear convolution acceleration step, Eq~\eqref{eq:trigonometric_N_terms} of the nonlinear convolution acceleration step adopts
the truncated range kernel $K_r^T(\bm{x})$ and best $N$-term approximation for $K_r^T(\bm{x})$ on trigonometric functions to approximate original range kernel $K_r(\bm{x})$ in steps \ding{196} \ding{197}, respectively. Sequentially, the shiftability property of cosine functions is exploited by step \ding{198} to eliminate the nonlinearity in the trigonometric best $N$-term approximation for $K_r^T(\bm{x})$. In addition, the nonlinearity in the box $\dot{B}(I(\bm{x}) - I(\bm{y}))$ is removed by the dimension promotion technique used by step \ding{199} in \eqref{eq:1D_box_N_terms}. At last, putting \eqref{eq:2D_box_N_terms_s} into \eqref{eq:1D_box_N_terms} and employing 3-D SAT, we can compute step \ding{200} in \eqref{eq:numerator_2D_box_N_terms} in linear time.


\section{Comparison with previous methods}
\label{sec:comparison}

Based on best $N$-term approximation, we employ 2-D Haar functions and 1-D truncated trigonometric functions to speed up the convolution of $K_s(x)$ and $K_r(x)$, and we put the two acceleration techniques together to compose our 3-D box filter based acceleration method. In this section, we provide some results on synthetic and natural data as well as a detailed analysis to understand the improvements of our proposal and the differences between our approach and existing methods.

\subsection{Comparison with the acceleration techniques of $K_s(x)$ }
\label{sec:comparison_Ks}

Although FFT~\cite{Durand_TOG_2002} can fast compute $\sum_{\bm{y} \in \mathcal{N}_{\bm{x}}} K_s(\| \bm{x}  -  \bm{y} \|) g(\bm{y})$, its complexity is not linear. Kernel separation based methods~\cite{Getreuer_ipol_2013} can complete the task in linear complexity. However, they are limited to the Gaussian function. Even worse, the approximation error will significantly degrade the smoothing quality of texture regions. To the best of our knowledge, the 2-D box filtering based algorithms are the first kind of techniques that can reduce the complexity $O(|\mathcal{N}_{\bm{x}}||I|)$ of the linear convolution of arbitrary $K_s(x)$ down to $O(|I|)$. Our Haar based fast computation method~\eqref{eq:2D_box_N_terms_s} also belongs to this kind of techniques. In this section, we will highlight the improvements and advantages of our approach, compared with other methods.

\subsubsection{Coefficients and support regions} Employing the linear combination $\sum_{i = 1}^{M_s} \sum_{\bm{y}} \beta^s_i \ddot{B}_i(\bm{x}  -  \bm{y}) g(\bm{y})$ of box filters to approximate $\sum_{\bm{y} \in \mathcal{N}_{\bm{x}}} K_s(\| \bm{x}  -  \bm{y} \|) g(\bm{y})$ is the key idea of box filtering based algorithms. In the acceleration literature~\cite{Zhang_TIP_2012, Gunturk_TIP_2011, Pan_MPE_2014}, the optimization approach is utilized to determine the coefficients and support regions of box functions $\ddot{B}_i(\bm{x})$. Compared with previous algorithms, the procedure of our method to determine the two parameters is much simpler. This is because we choose the first $N$ largest coefficients of Haar functions (\ie the best $N$-term approximation on Haar functions). Meanwhile, the support regions of box functions are predefined by Haar functions $\{ \phi(x), \psi_{j,k}(x) \}$ with the size of $\frac{T_s}{2^{j-1}}$, where $j \in \{ 0, \ldots, \infty \}$. Table~\ref{tab:time_cs} plots the run time of four methods used to determine the coefficients and support regions of box filters. It can be seen that our method spends the least time for acquiring these parameters.
\begin{table}[b]
\caption{The time of computing the coefficients and support regions of box filters, where the number of used box filters is 3. Note that although the method of Zhang is comparable with ours, this method can only accelerate the Gaussian function.}
\label{tab:time_cs}
\centering
\begin{tabularx}{\linewidth}{@{}YYYYY@{}}
\hline
 & Zhang~\cite{Zhang_TIP_2012}  & Gunturk~\cite{Gunturk_TIP_2011} & Pan~\cite{Pan_MPE_2014} &  Ours \\ \hline
Time & 0.1s & 2s & 3s & 0.05s \\ \hline
\end{tabularx}
\end{table}

\subsubsection{The computational complexity equivalence between the linear combinations of Haar functions and box filters}

\begin{figure}[t]
    \centering
	\begin{subfigure}[b]{0.48\linewidth}
		\includegraphics[width=\textwidth ]{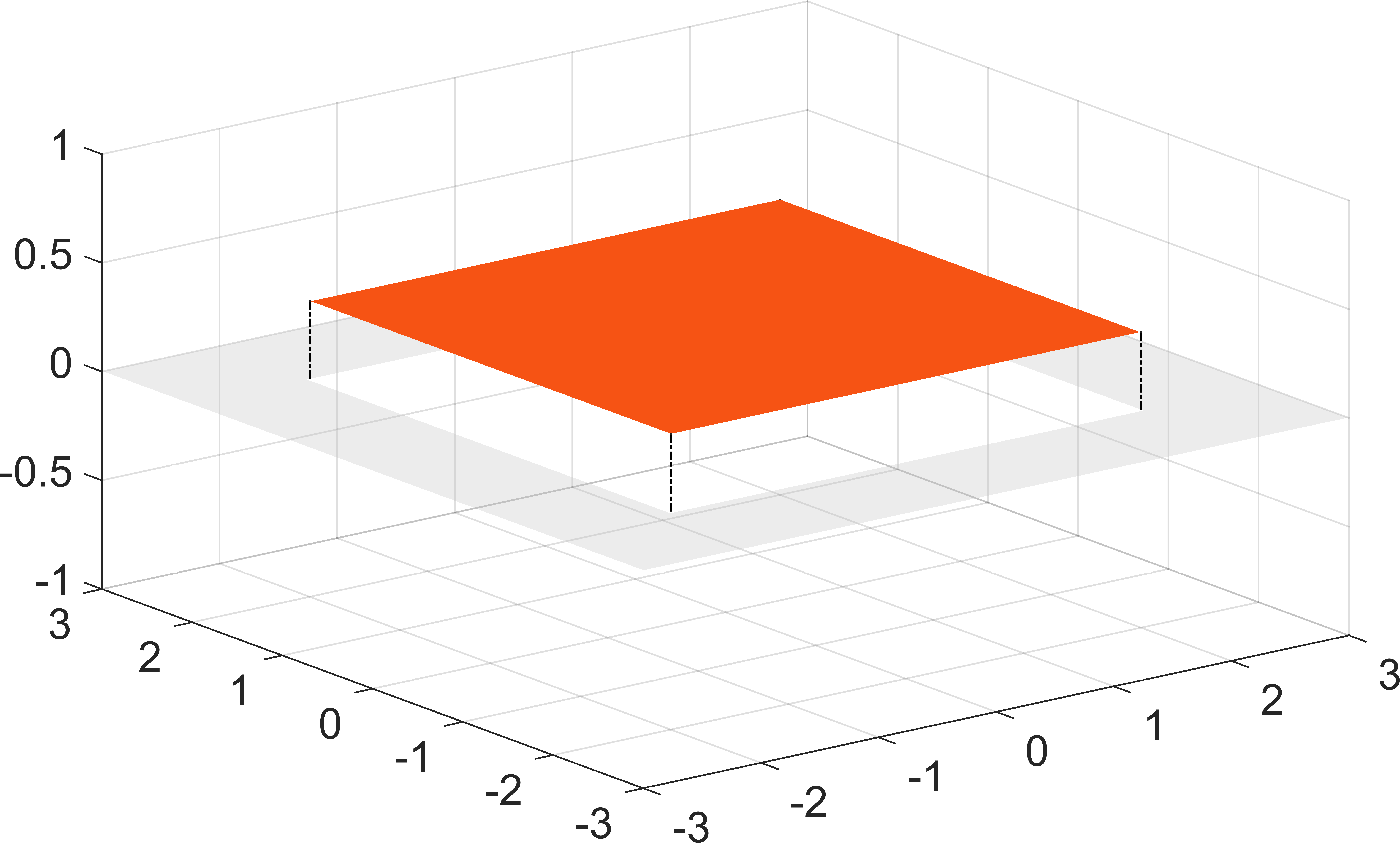}
		\caption{$f_1(x,y)$}	
		\label{fig:linear_combinations:1}
	\end{subfigure}
    \begin{subfigure}[b]{0.48\linewidth}
		\includegraphics[width=\textwidth ]{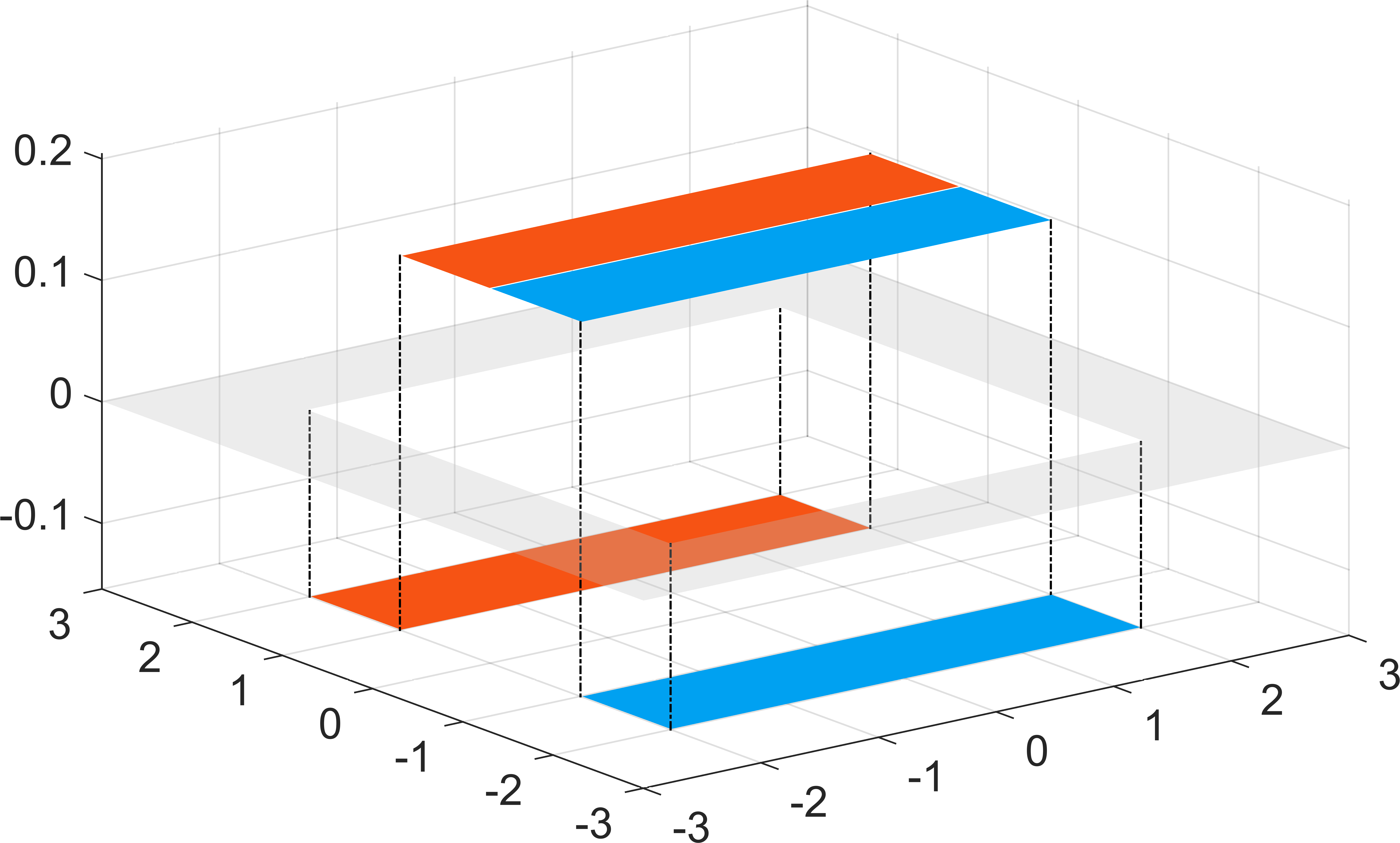}
		\caption{$f_2(x,y)$}	
		\label{fig:linear_combinations:2}
	\end{subfigure}

    \begin{subfigure}[b]{0.48\linewidth}
		\includegraphics[width=\textwidth ]{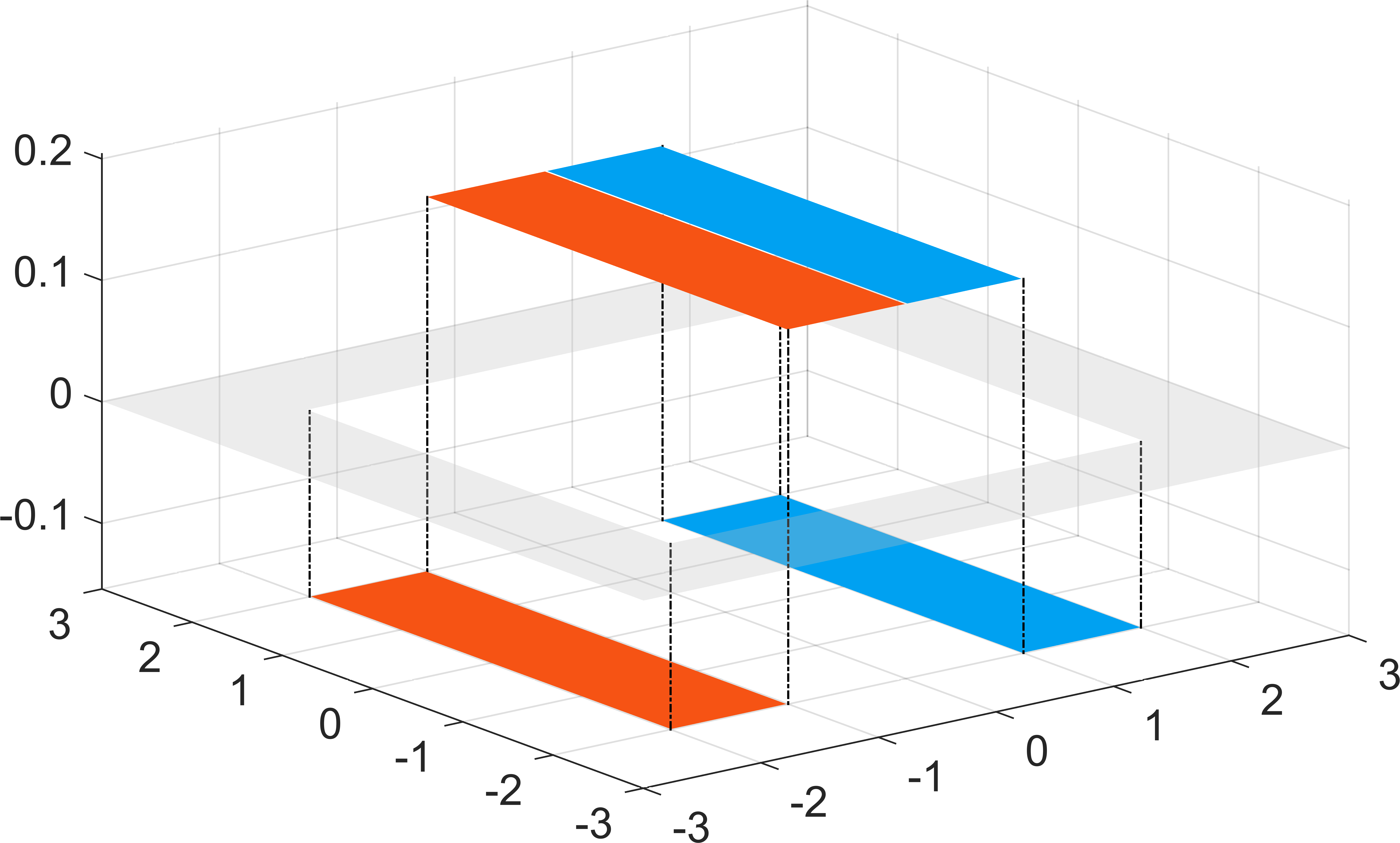}
		\caption{$f_3(x,y)$}	
		\label{fig:linear_combinations:3}
	\end{subfigure}
    \begin{subfigure}[b]{0.48\linewidth}
		\includegraphics[width=\textwidth ]{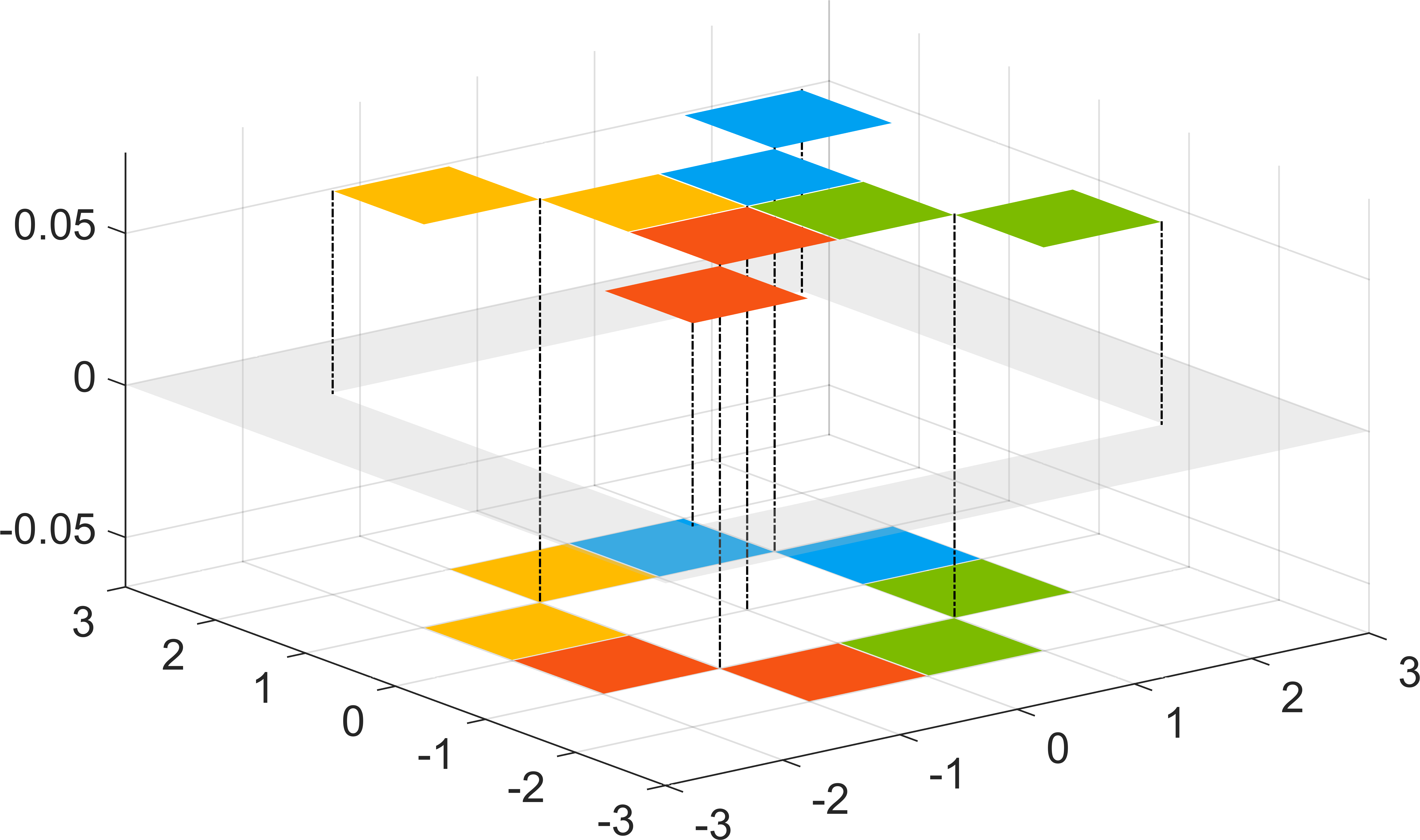}
		\caption{$f_4(x,y)$}	
		\label{}
	\end{subfigure}
    \caption{Visual illustration for the linear combinations on the 2-D basis functions $\phi(x)\phi(y)$, $\phi(x)\psi_{1,k_2}(y)$, $\psi_{1,k_1}(x) \phi(y)$, $\psi_{1, k_1}(x) \psi_{1, k_2}(y)$. Here, we denote the four linear combinations as $f_1(x,y)$, $f_2(x,y)$, $f_3(x,y)$, $f_4(x,y)$, respectively, and different color implies different basis functions exploited in the linear combinations. }	
	\label{fig:linear_combinations}
\end{figure}

\begin{figure*}[t]
    \centering
	\begin{subfigure}[b]{0.24\linewidth}
		\includegraphics[width=\textwidth ]{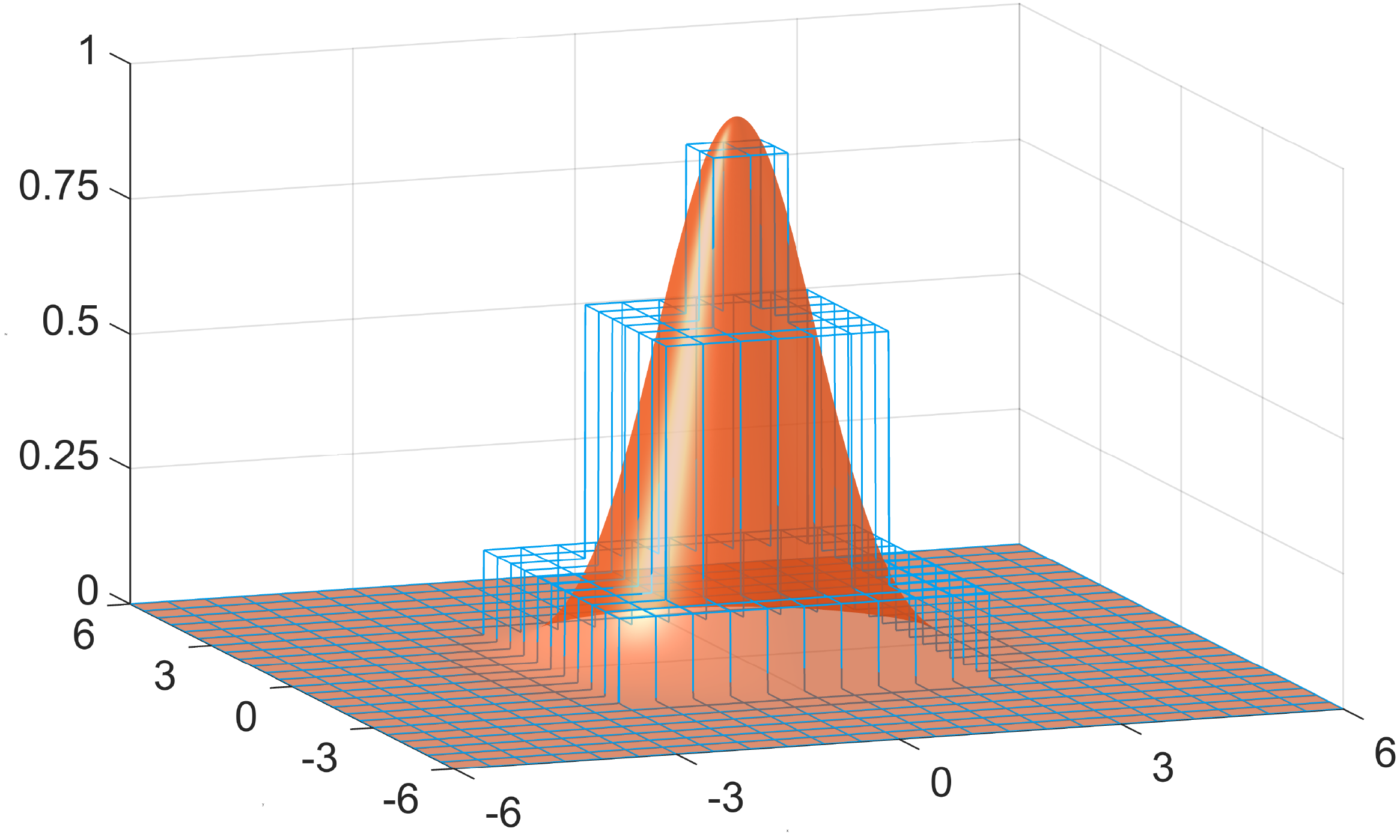}
	\end{subfigure}
    \begin{subfigure}[b]{0.24\linewidth}
		\includegraphics[width=\textwidth ]{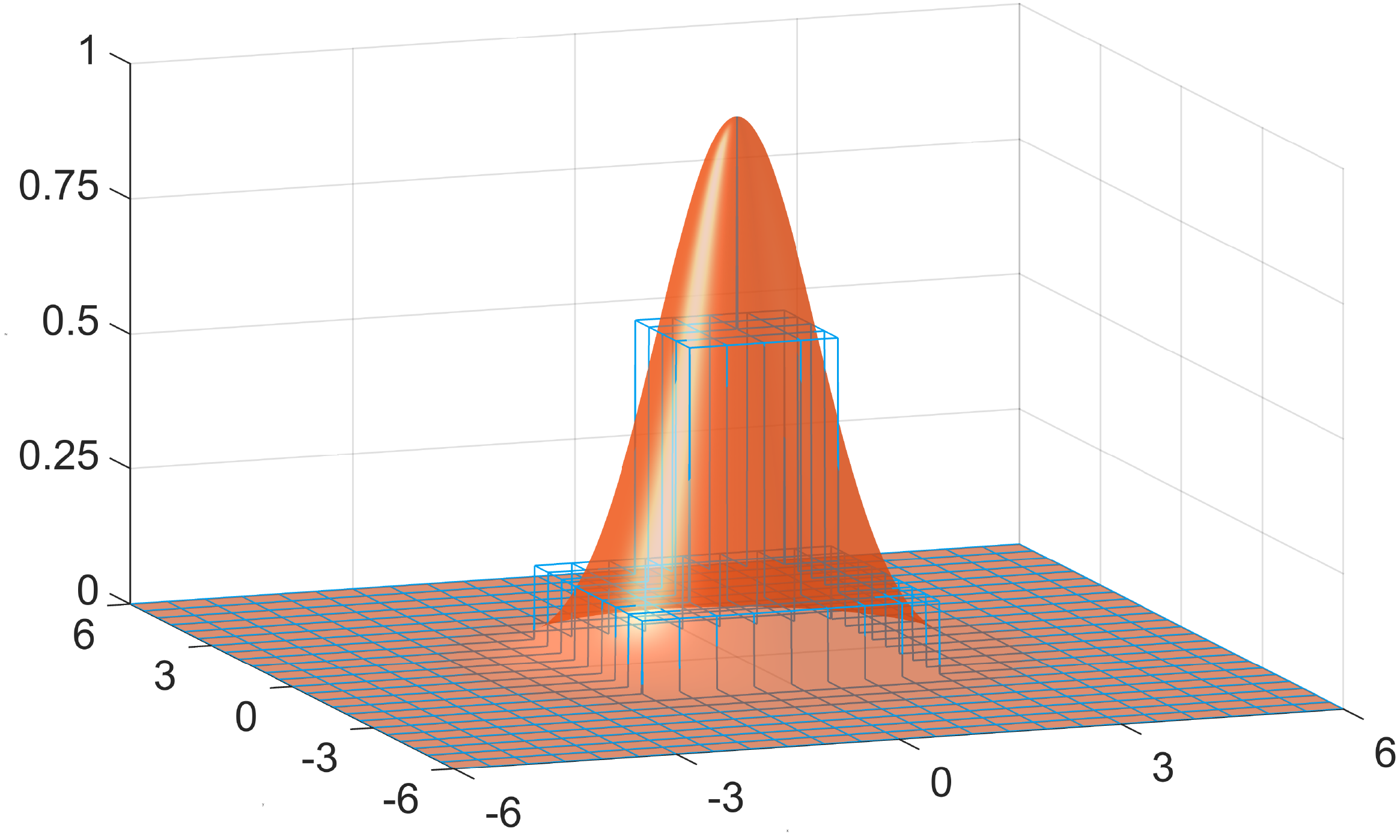}
	\end{subfigure}
    \begin{subfigure}[b]{0.24\linewidth}
		\includegraphics[width=\textwidth ]{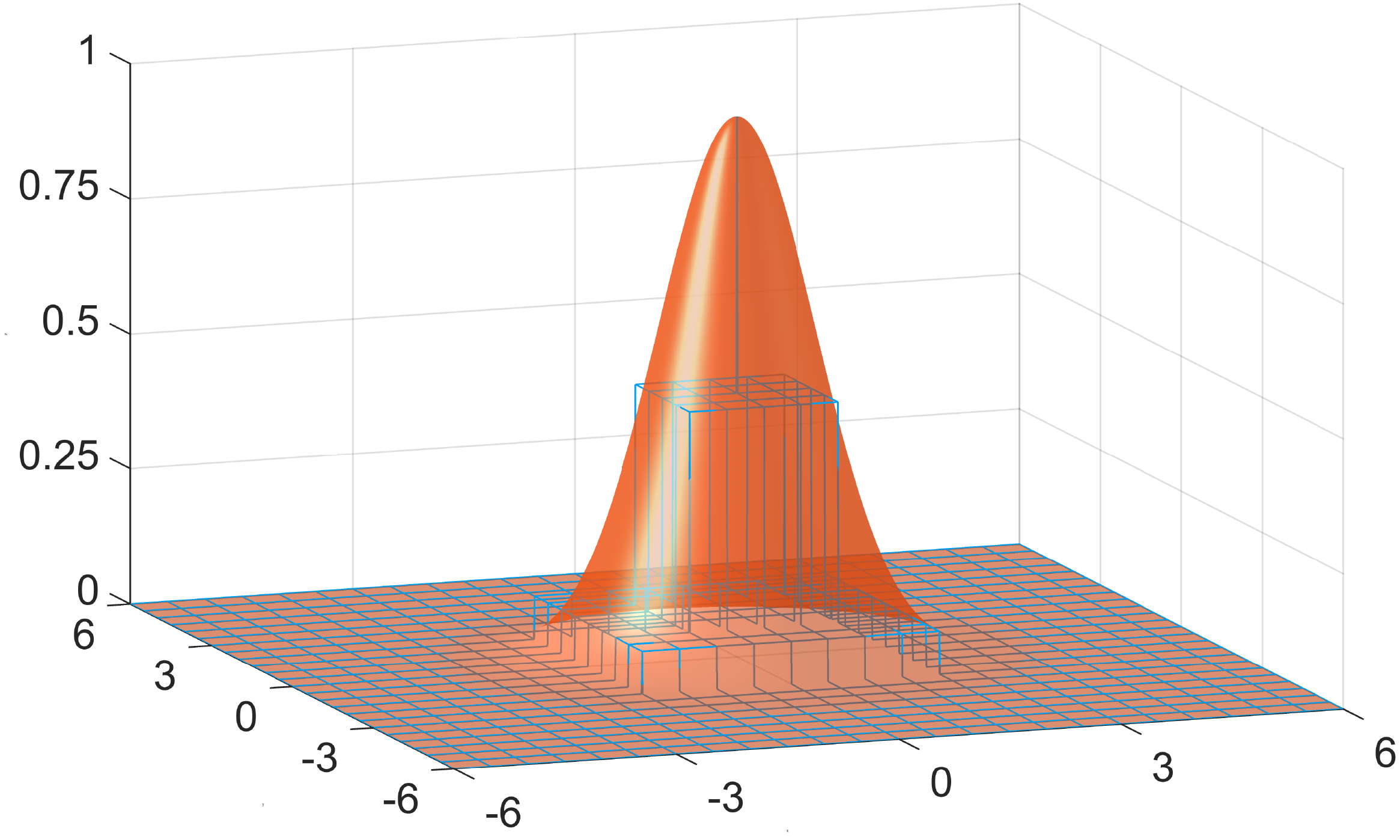}
	\end{subfigure}
    \begin{subfigure}[b]{0.24\linewidth}
		\includegraphics[width=\textwidth ]{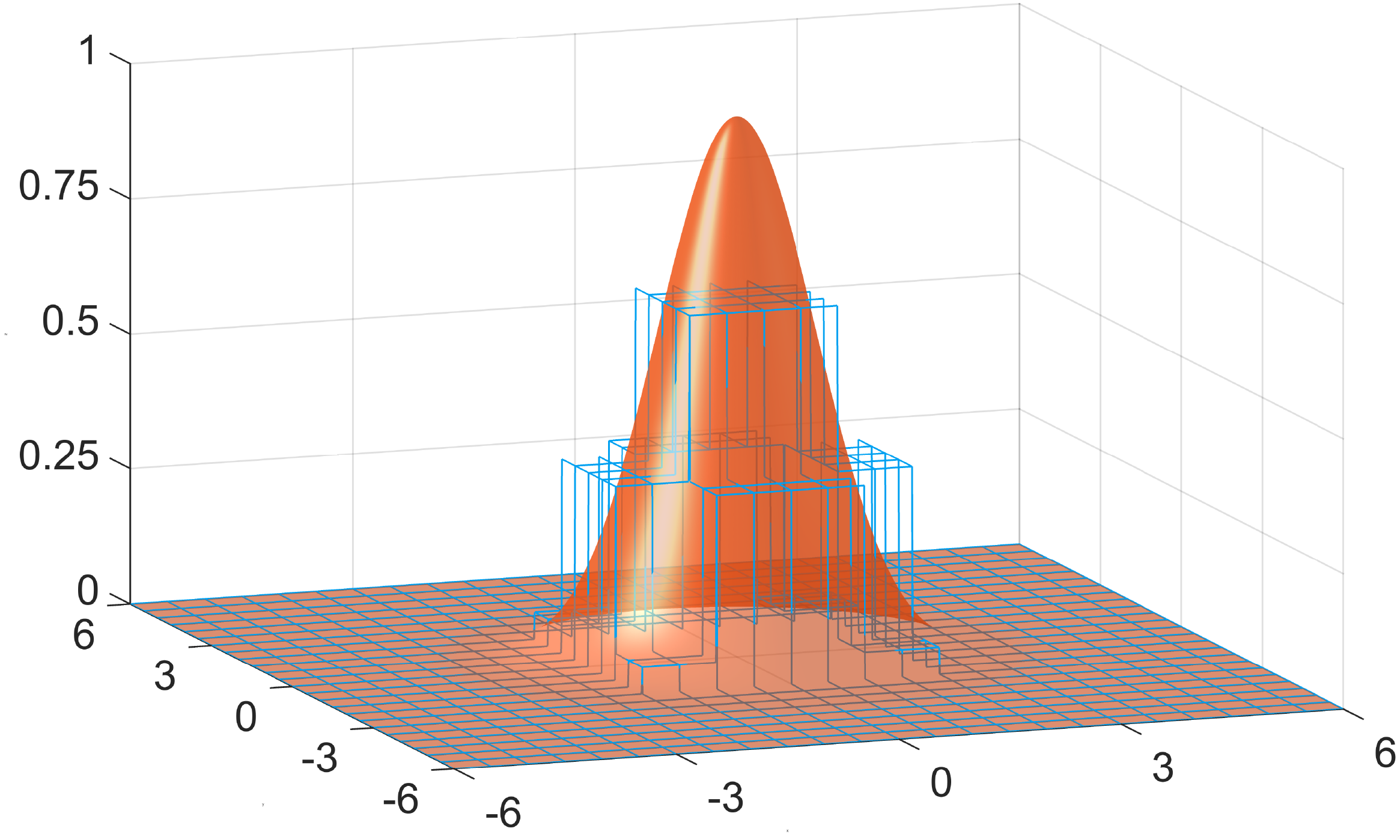}
	\end{subfigure}

    \begin{subfigure}[b]{0.24\linewidth}
		\includegraphics[width=\textwidth ]{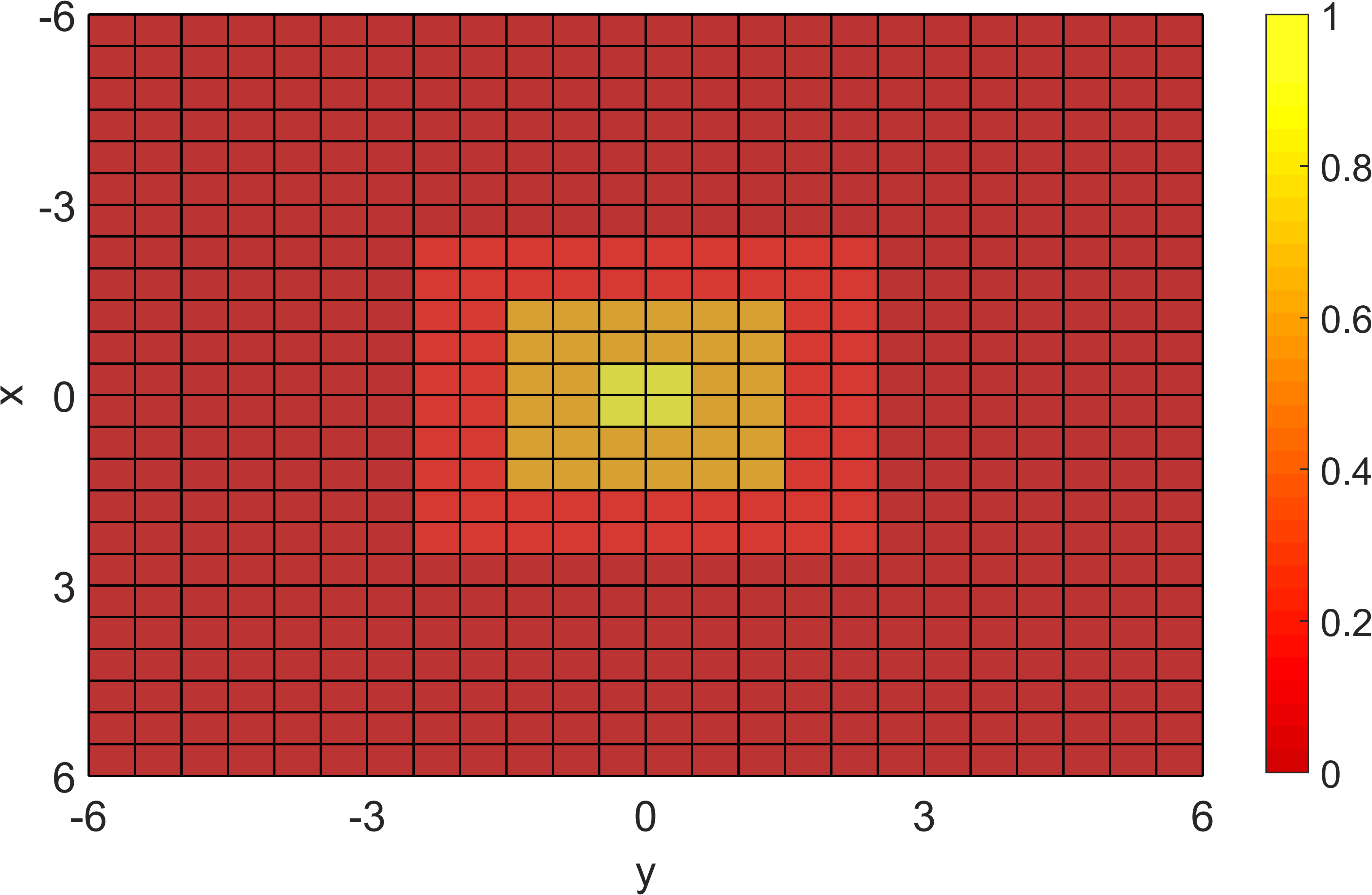}
		\caption{Zhang~\cite{Zhang_TIP_2012}}	
		\label{fig:box_approxiamtion:zhang}
	\end{subfigure}
    \begin{subfigure}[b]{0.24\linewidth}
		\includegraphics[width=\textwidth ]{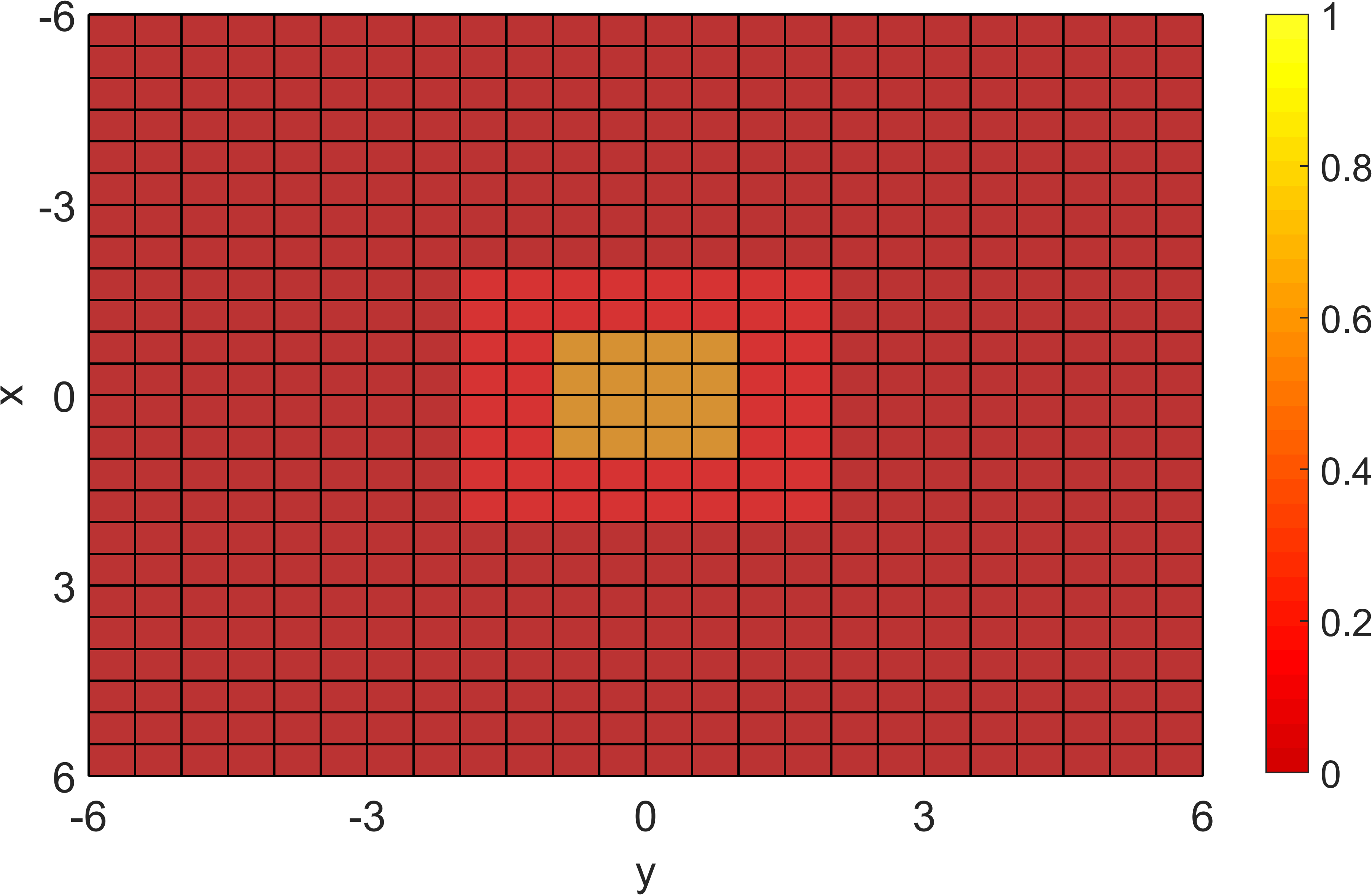}
		\caption{Gunturk~\cite{Gunturk_TIP_2011}}	
		\label{fig:box_approxiamtion:gunturk}
	\end{subfigure}
    \begin{subfigure}[b]{0.24\linewidth}
		\includegraphics[width=\textwidth ]{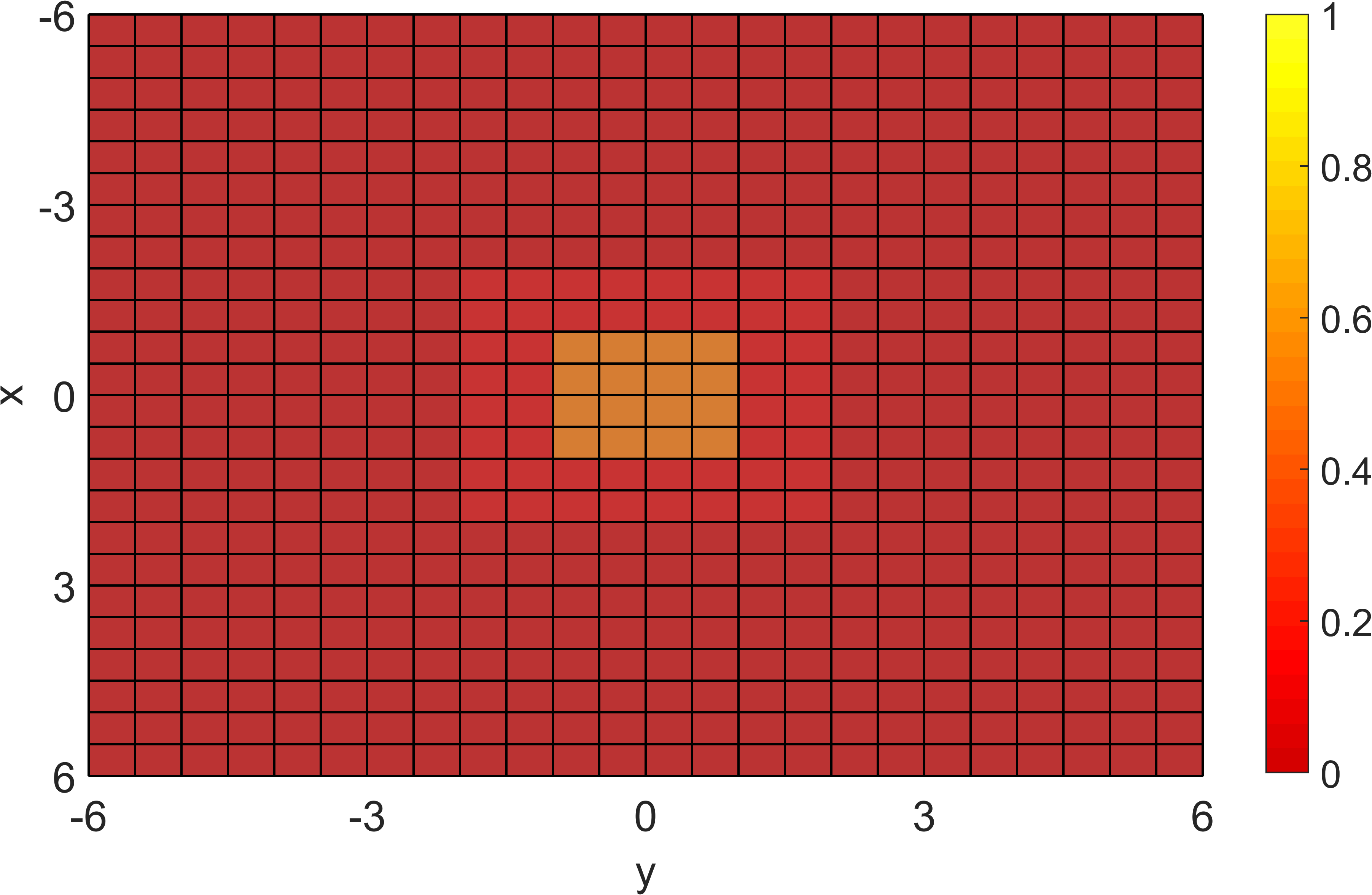}
		\caption{Pan~\cite{Pan_MPE_2014}}	
		\label{fig:box_approxiamtion:pan}
	\end{subfigure}
    \begin{subfigure}[b]{0.24\linewidth}
		\includegraphics[width=\textwidth ]{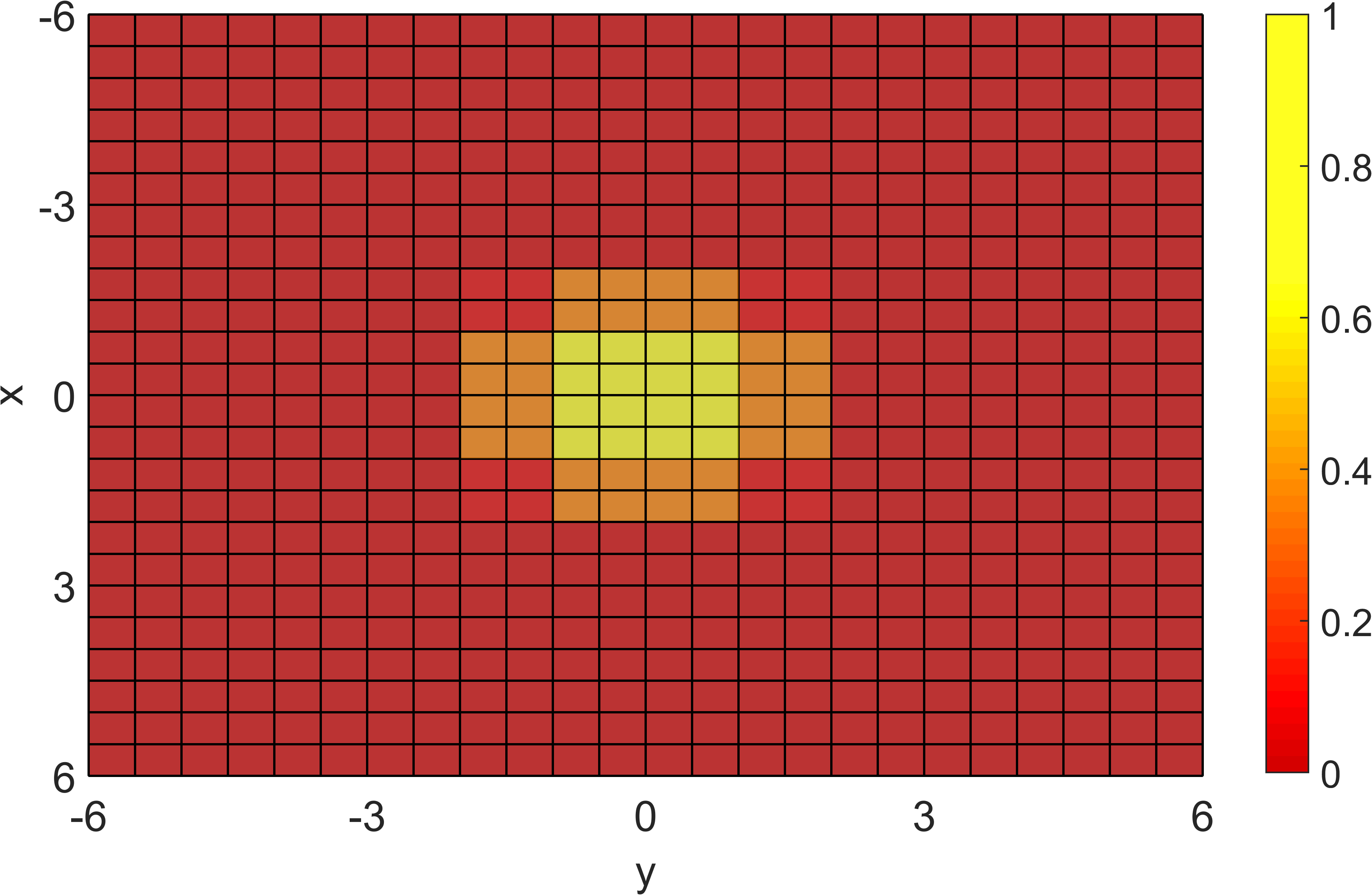}
		\caption{Ours}	
		\label{fig:box_approxiamtion:ours}
	\end{subfigure}
    \caption{Visual comparison of four different box filter approximations for the Gaussian function $G_1(x)$, where the first row shows the Gaussian function $G_1(x)$ and the four methods' approximations. In order to disclose the difference of the four approximations more clearly, the second row illustrates the density map plots of the four approximations. Note that Zhang~\cite{Zhang_TIP_2012}, Gunturk~\cite{Gunturk_TIP_2011} and Pan~\cite{Pan_MPE_2014} cascade three box filters together, our method compute the linear combination $f_1(x,y)+f_2(x,y)+f_3(x,y)$ which can be synthesized by three box filters. }	
	\label{fig:box_approxiamtion}
\end{figure*}

To approximate the spatial kernel $K_s(\| \bm{x}  -  \bm{y} \|)$ centered at pixel $\bm{x}$, all previous methods assume $\ddot{B}_i(\bm{x} - \bm{y}) $ must be centered at the given pixel $\bm{x}$ too. Our Haar functions based method breaks this assumption and jointly uses a set of box functions deviated from the pixel $\bm{x}$ as well as the box functions centered at $\bm{x}$ to approximate the kernel $K_s(\| \bm{x}  -  \bm{y} \|)$. The major advantage of our approach is that it reuses box filtering results and therefore can employ fewer box filters to obtain more accurate approximation results. For instance, although $\psi_{j_1, k_1}(x) \psi_{j_2, k_2}(y)$ is consisted of four box functions (or four 2-D box filters), the size of the support regions of these box functions is same to $2^{-j_1+1} T_s \times 2^{-j_2+1} T_s $. This can be demonstrated in  Fig~\ref{fig:Haar_2D} which implies that for the convolution kernel $\psi_{j_1, k_1}(x) \psi_{j_2, k_2}(y)$, the filtering result of $\bm{x}$ is a linear composition of the box filtering result of four pixels around pixel $\bm{x}$. Hence,
the computational complexity of the convolution with the kernel $ \psi_{j_1, k_1}(x) \psi_{j_2, k_2}(y)$ is same to the computational complexity of the box filter with the support region $2^{-j_1+1} T_s \times 2^{-j_2+1} T_s $. Similar discussion can also be applied to the basis functions $\phi(x)\phi(y), \phi(x)\psi_{j_2,k_2}(y), \psi_{j_1,k_1}(x) \phi(y)$ according to the discussion in section~\ref{sec:Ks}. 

Employing box filters, we can synthesis the linear combinations of 2-D Haar functions. For instance,
Fig~\ref{fig:linear_combinations} demonstrates four 2-D Haar functions' linear combinations $f_1(x, y) = c_0 \phi(x)\phi(y)$, $f_2(x, y) = \sum_{k_2 \in \mathcal{K} } c_{0,1,k_2}\phi(x)\psi_{1,k_2}(y)$, $f_3(x, y) = \sum_{k_1 \in \mathcal{K}} c_{1,k_1,0} \psi_{1,k_1}(x) \phi(y)$ and $f_4(x, y) = \sum_{ k_1, k_2 \in \mathcal{K}} c_{1,k_1,1,k_2} \psi_{1, k_1}(x) \psi_{1, k_2}(y)$, where $\mathcal{K} = \{ -1, 1\}$. The coefficients $c_0$, $c_{0,1,k_2}$, $c_{1,k_1,0}$, $c_{1,k_1,1,k_2}$  are computed from \eqref{eq:Haar_2D_coefficients} based on the Gaussian function $G_1(x)$ as illustrated in Fig~\ref{fig:box_approxiamtion}. From Fig~\ref{fig:linear_combinations}, we can find that each linear combination can be decomposed into several box functions with the same support regions. This conclusion can be generalized further.
According to appendix~\ref{sec:box_filter}, we could reasonably conclude that given $j_1, j_2$, the linear combinations of $\phi(x)\phi(y)$, $\phi(x)\psi_{j_2,k_2}(y)$, $\psi_{j_1,k_1}(x) \phi(y)$, $\psi_{j_1, k_1}(x) \psi_{j_2, k_2}(y)$ can be synthesized from four box filters with support regions $2 T_s \times 2 T_s $, $2 T_s \times 2^{-j_2+1} T_s $, $2^{-j_1+1} T_s \times 2 T_s $, $2^{-j_1+1} T_s \times 2^{-j_2+1} T_s $, respectively. So it is rational to equally treat the four linear combinations of Haar functions and the four box filters from the computational complexity perspective. Moreover, the computational complexity of the linear combinations $f_1(x, y) + f_2(x, y) + f_3(x, y)$ equals to three box filters. In the next section, we will take it to approximate the spatial Gaussian kernel function.


\begin{figure}[t]
\begin{subfigure}[c]{0.31\linewidth}
\includegraphics[width = \linewidth]{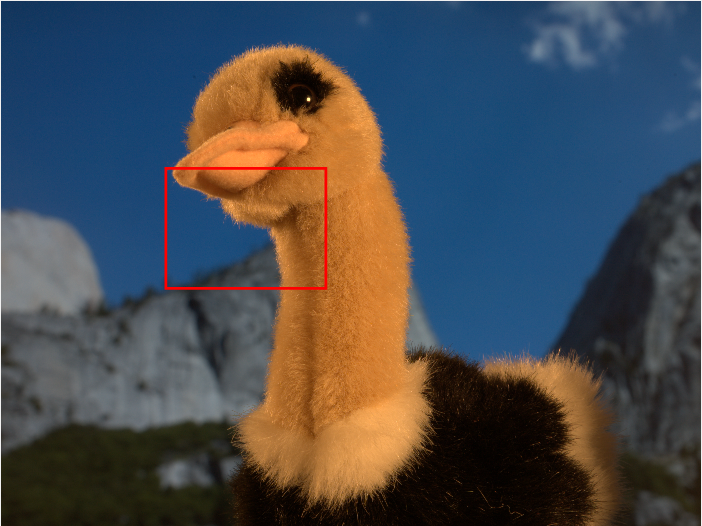}
\end{subfigure}
\vspace{0.02cm}
\begin{subfigure}[c]{0.31\linewidth}
\includegraphics[width = \linewidth]{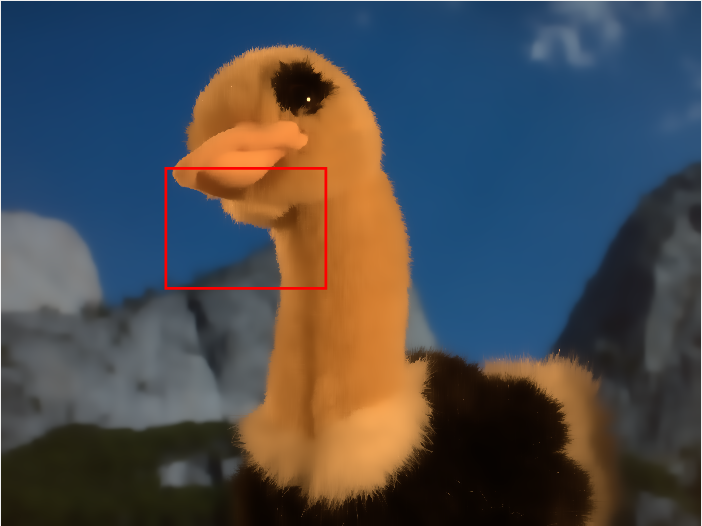}
\end{subfigure}
\vspace{0.02cm}
\begin{subfigure}[c]{0.31\linewidth}
\includegraphics[width = \linewidth]{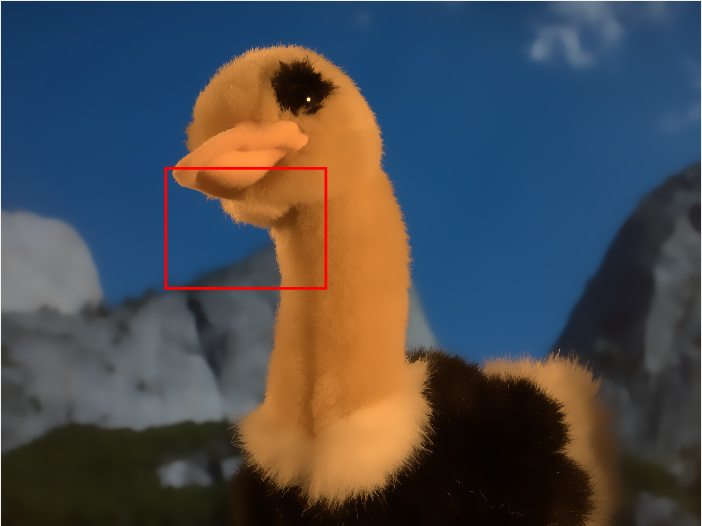}
\end{subfigure}
\vspace{0.02cm}

\begin{subfigure}[c]{0.31\linewidth}
\includegraphics[width = \linewidth]{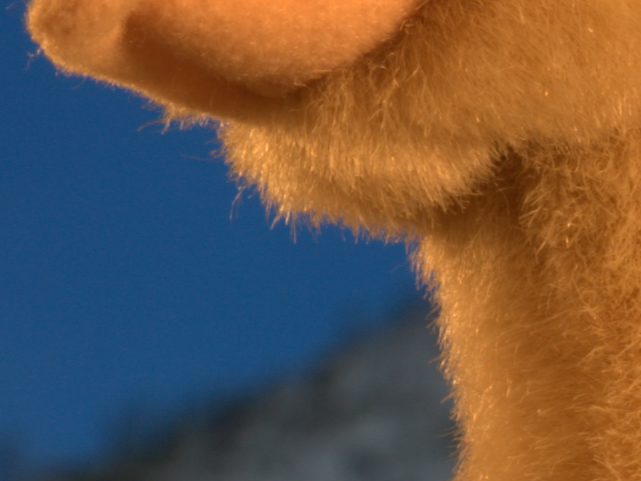}
\end{subfigure}
\vspace{0.04cm}
\begin{subfigure}[c]{0.31\linewidth}
\includegraphics[width = \linewidth]{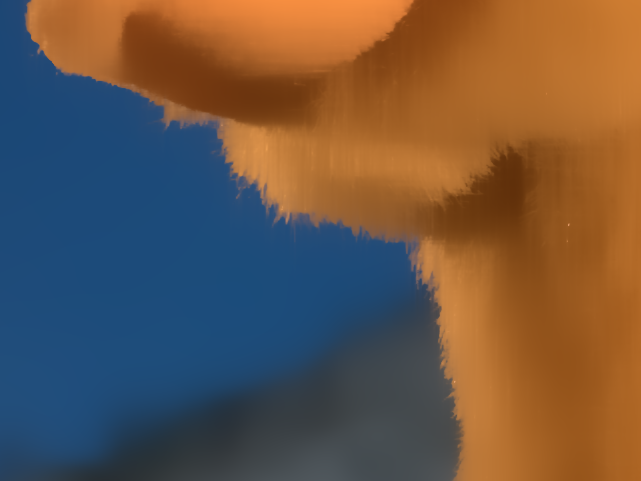}
\end{subfigure}
\vspace{0.04cm}
\begin{subfigure}[c]{0.31\linewidth}
\includegraphics[width = \linewidth]{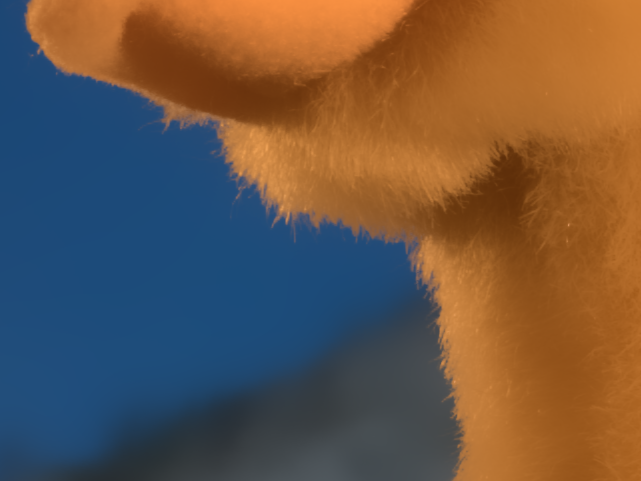}
\end{subfigure}
\vspace{0.04cm}

\begin{subfigure}[c]{0.31\linewidth}
\includegraphics[width = \linewidth]{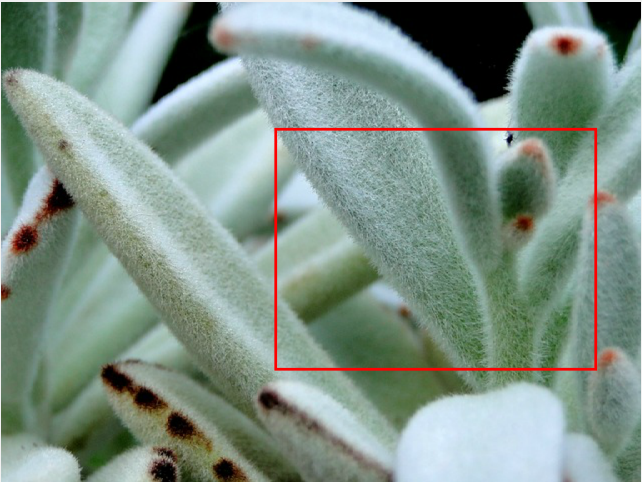}
\end{subfigure}
\vspace{0.02cm}
\begin{subfigure}[c]{0.31\linewidth}
\includegraphics[width = \linewidth]{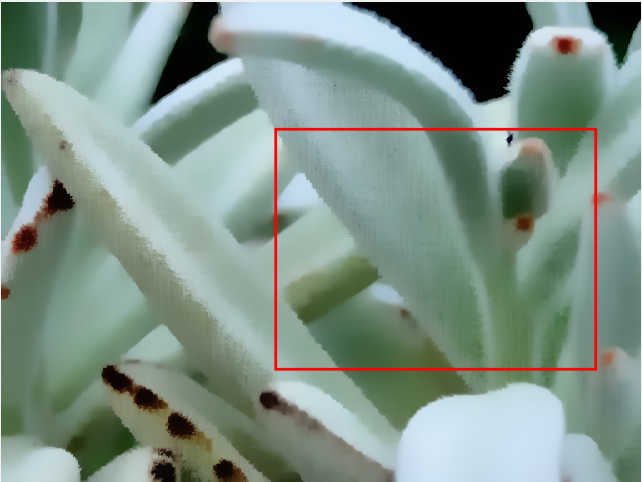}
\end{subfigure}
\vspace{0.02cm}
\begin{subfigure}[c]{0.31\linewidth}
\includegraphics[width = \linewidth]{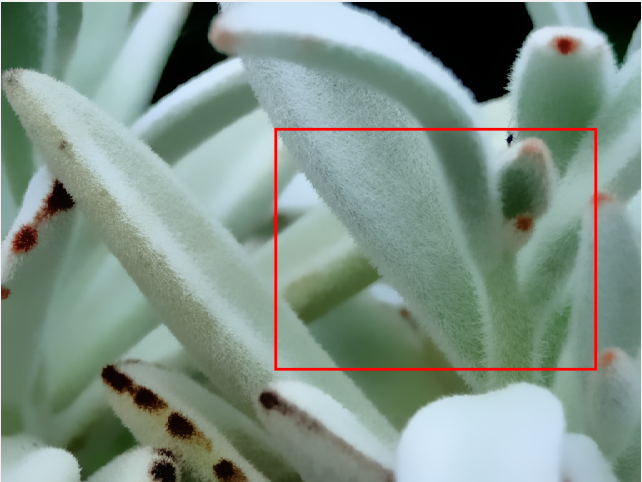}
\end{subfigure}
\vspace{0.02cm}

\begin{subfigure}[c]{0.31\linewidth}
\includegraphics[width = \linewidth]{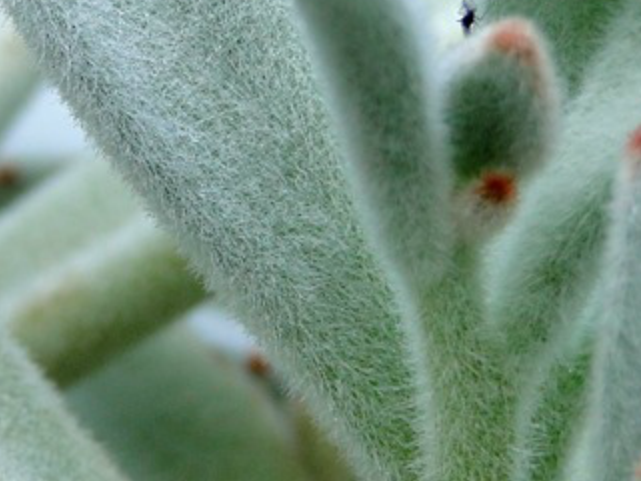}
\caption{Input}	
\end{subfigure}
\begin{subfigure}[c]{0.31\linewidth}
\includegraphics[width = \linewidth]{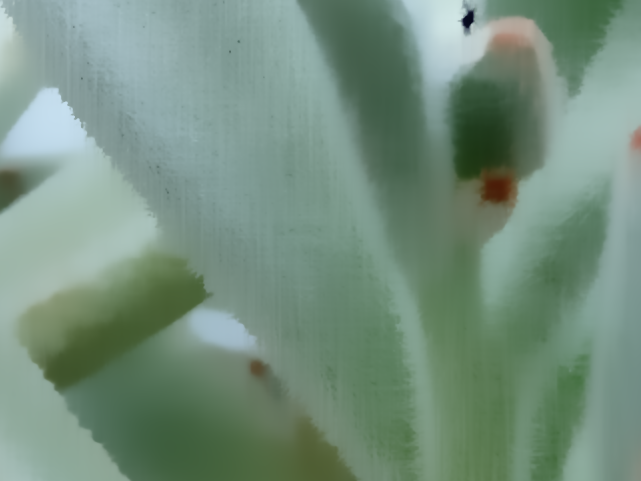}
\caption{Kernel separation}	
\end{subfigure}
\begin{subfigure}[c]{0.31\linewidth}
\includegraphics[width = \linewidth]{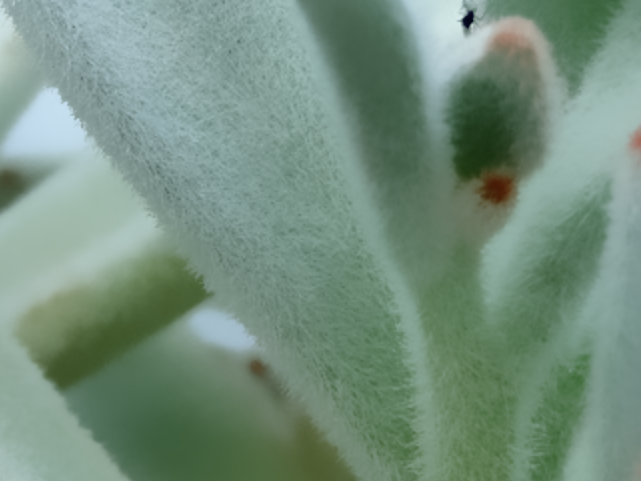}
\caption{Ours}	
\end{subfigure}

\caption{Comparisons of smoothing result in texture regions. (a) is the input image. (b) is the smoothing result produced by the kernel separation method~\cite{Rachid_TR_1993}. (c) is the smoothing result of ours. We can observe that the kernel separation method tends to over smooth the image along the vertical and horizontal edges due to 1-D handling of spatial domain. }
\label{fig:Duck}
\end{figure}

\renewcommand{\tabcolsep}{0.1cm}
\begin{figure*}[t]
\begin{tabular}{ccccc}
\multirow{2}{*}{
\begin{adjustbox}{valign=m}
\begin{subfigure}[c]{0.3\linewidth}
\vspace{-0.5cm}
\includegraphics[width = \linewidth]{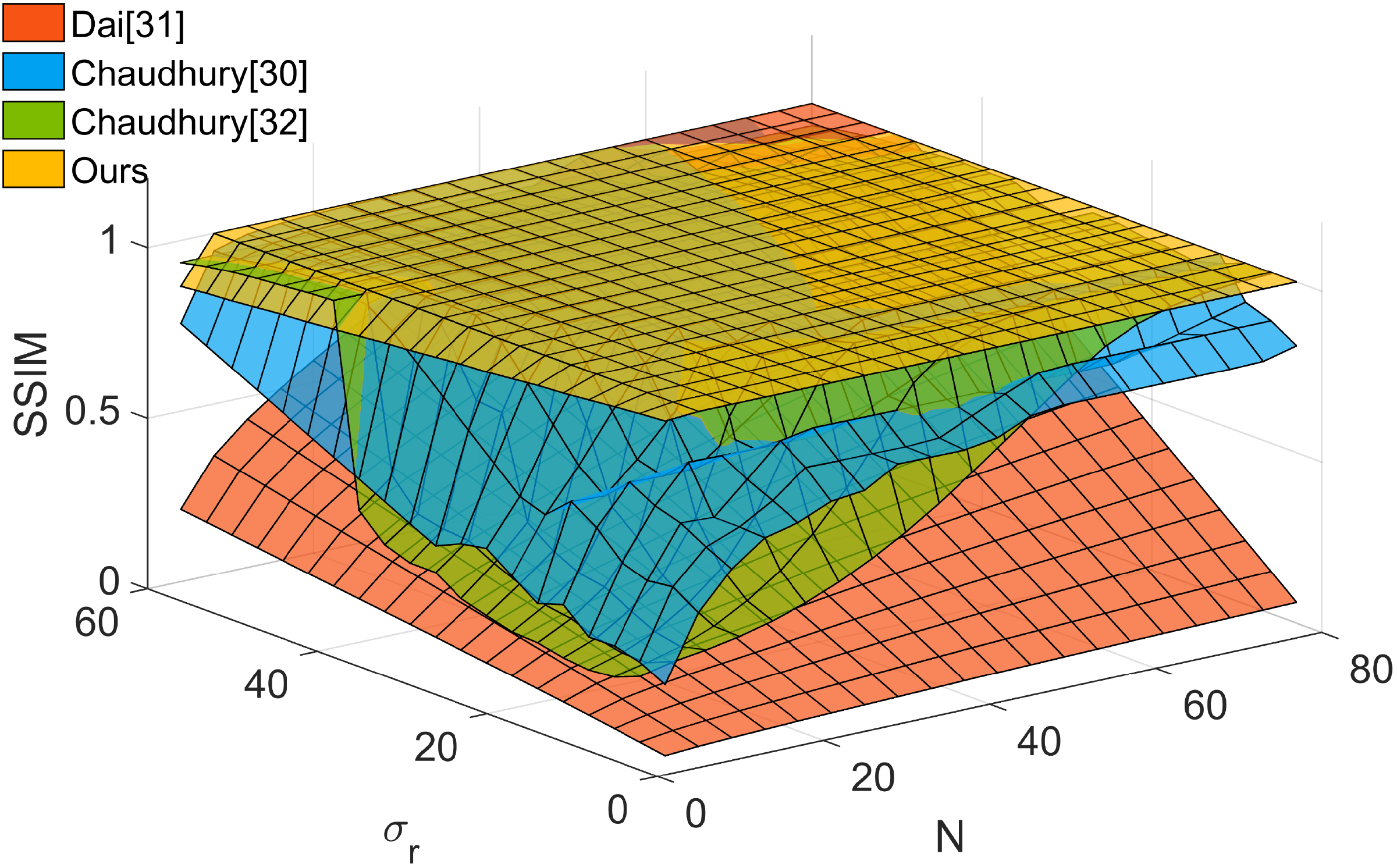}
\caption{SSIM evaluation}
\label{fig:shiftable_approxiamtion:SSIM}
\end{subfigure}
\end{adjustbox}
}
&
\begin{adjustbox}{valign=m}
\begin{subfigure}[c]{0.15\linewidth}
\includegraphics[width = \linewidth]{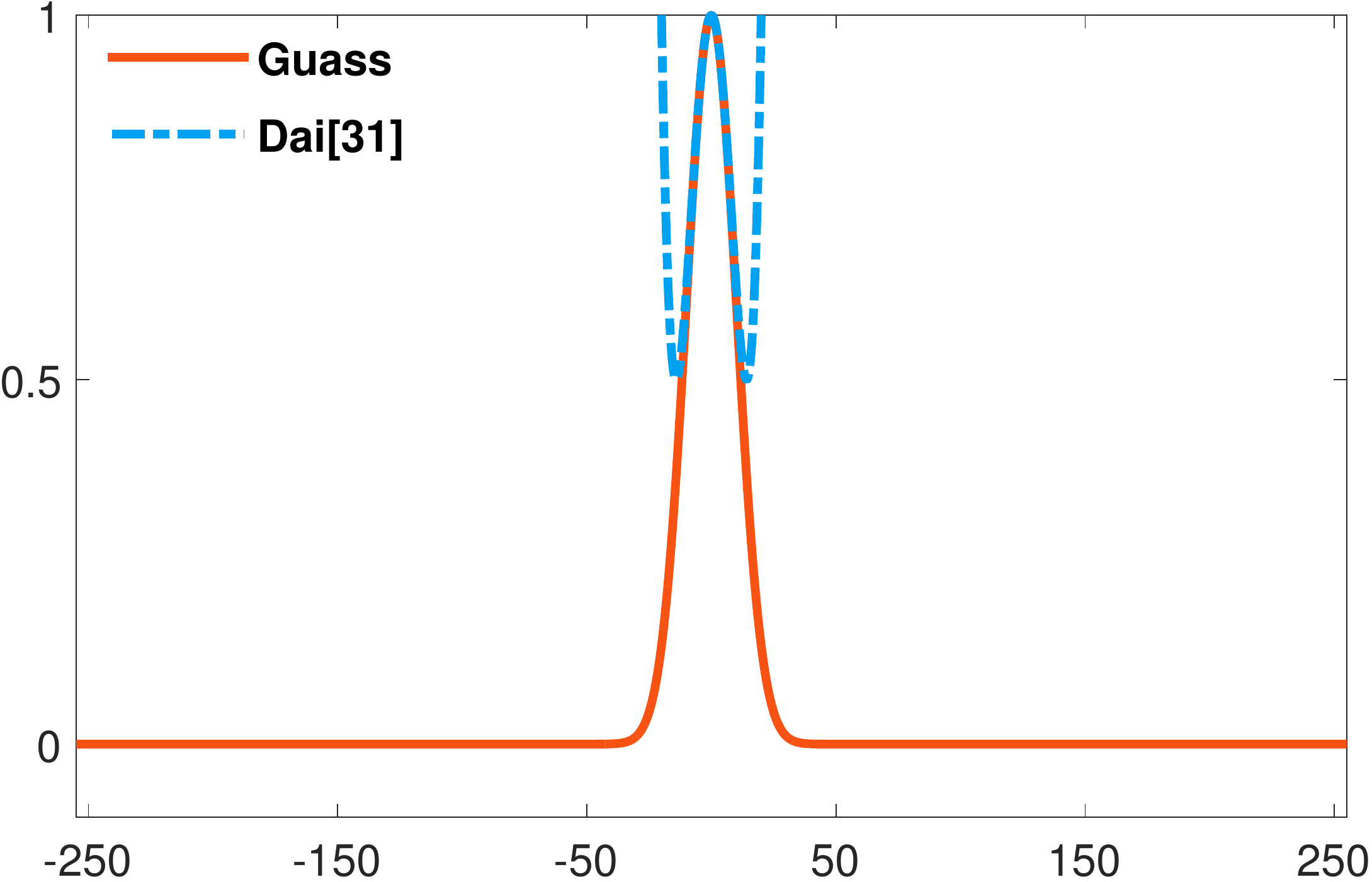}
\end{subfigure}
\end{adjustbox}
&
\begin{adjustbox}{valign=m}
\begin{subfigure}[c]{0.15\linewidth}
\includegraphics[width = \linewidth]{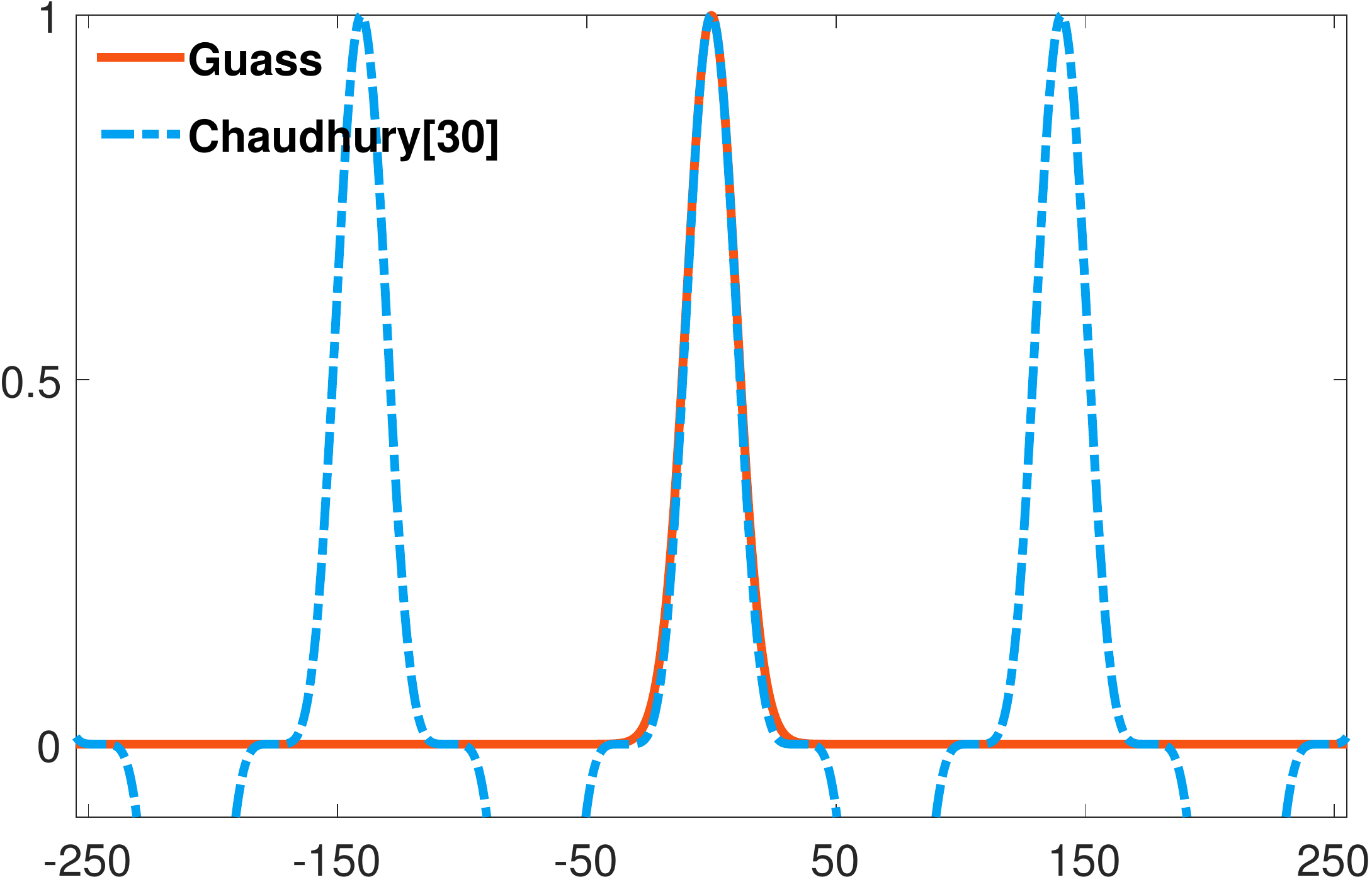}
\end{subfigure}
\end{adjustbox}
&
\begin{adjustbox}{valign=m}
\begin{subfigure}[c]{0.15\linewidth}
\includegraphics[width = \linewidth]{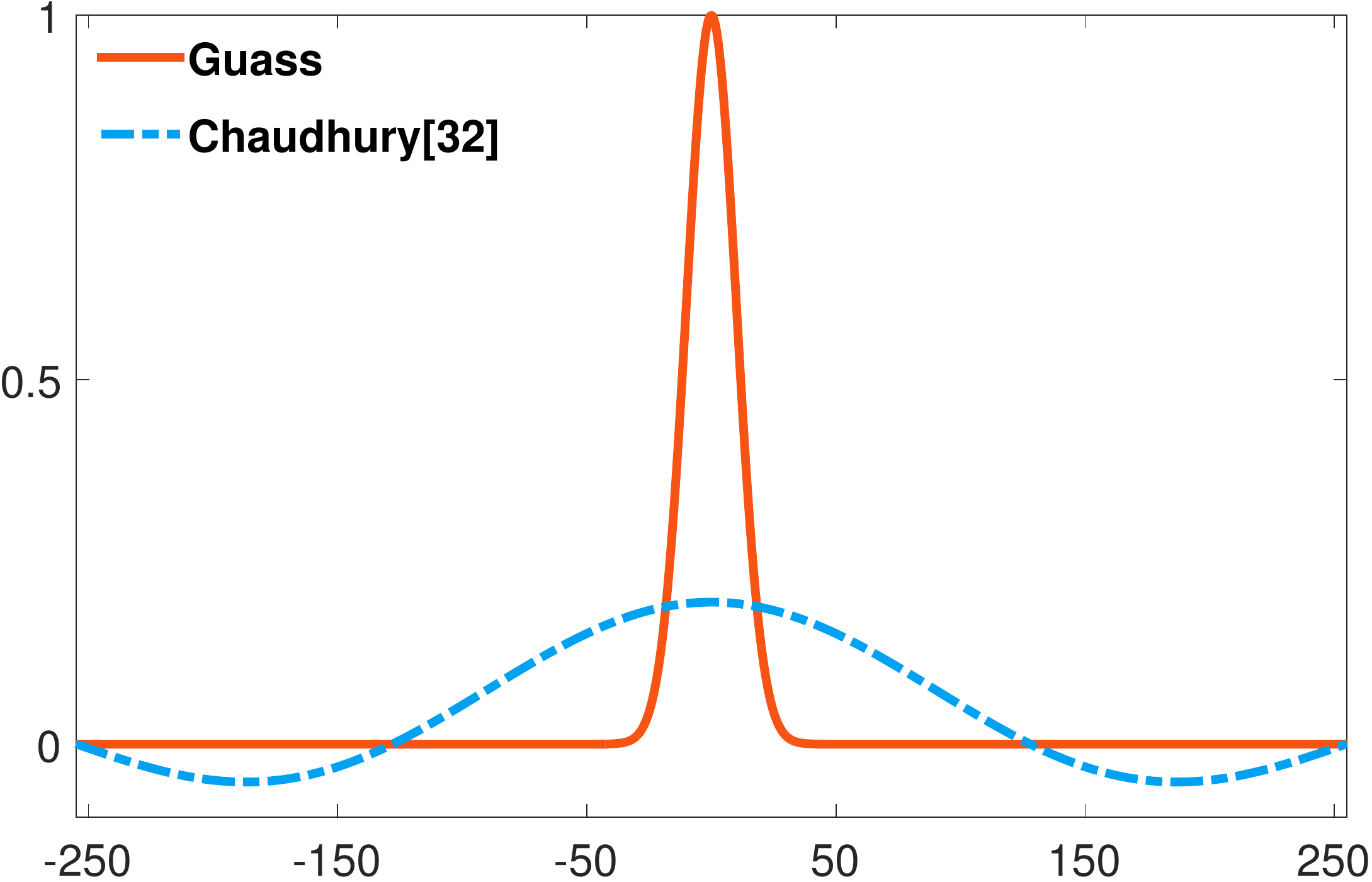}
\end{subfigure}
\end{adjustbox}
&
\begin{adjustbox}{valign=m}
\begin{subfigure}[c]{0.15\linewidth}
\includegraphics[width = \linewidth]{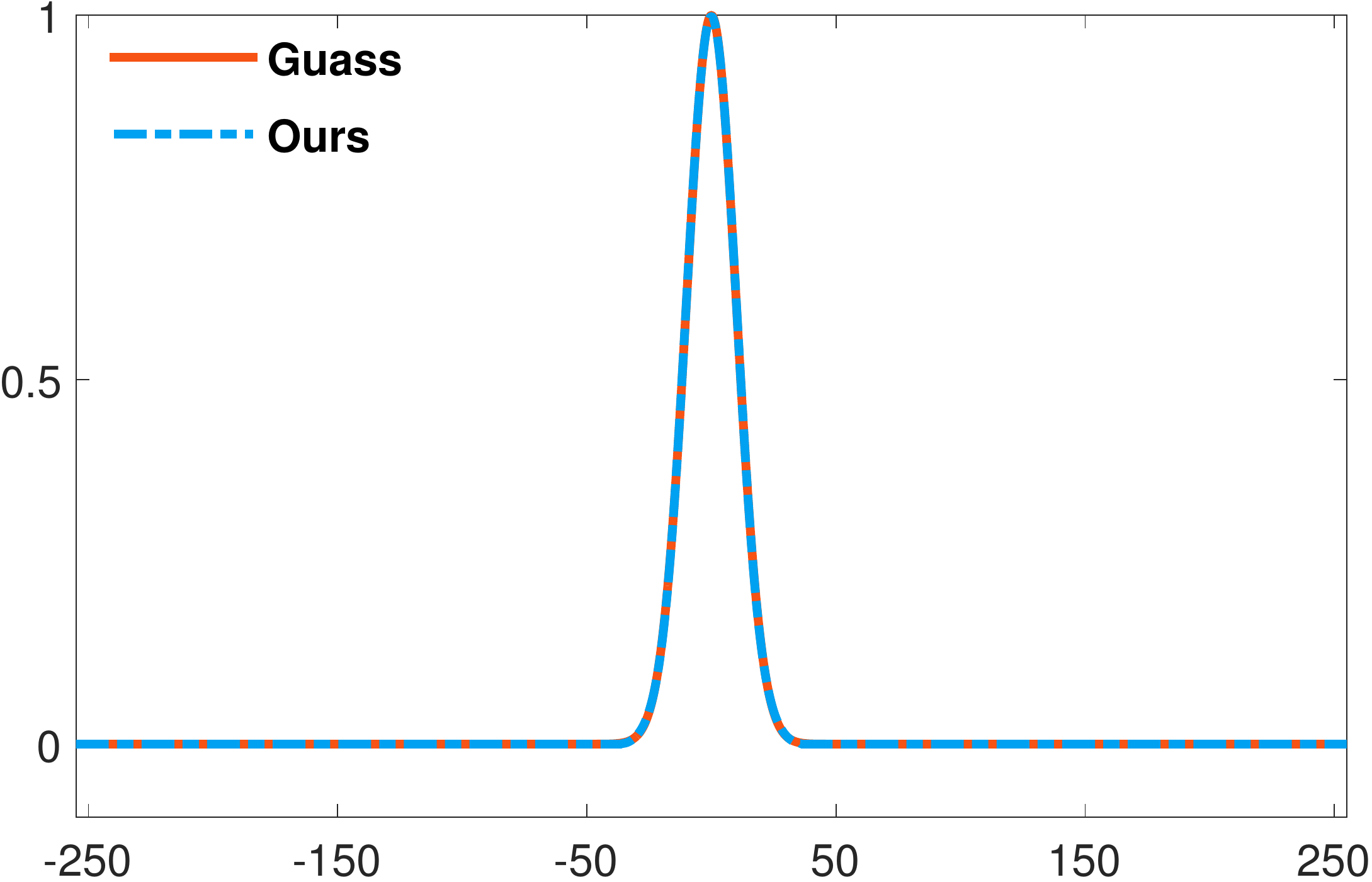}
\end{subfigure}
\end{adjustbox}
\\

&
\begin{adjustbox}{valign=m}
\begin{subfigure}[c]{0.15\linewidth}
\includegraphics[width = \linewidth]{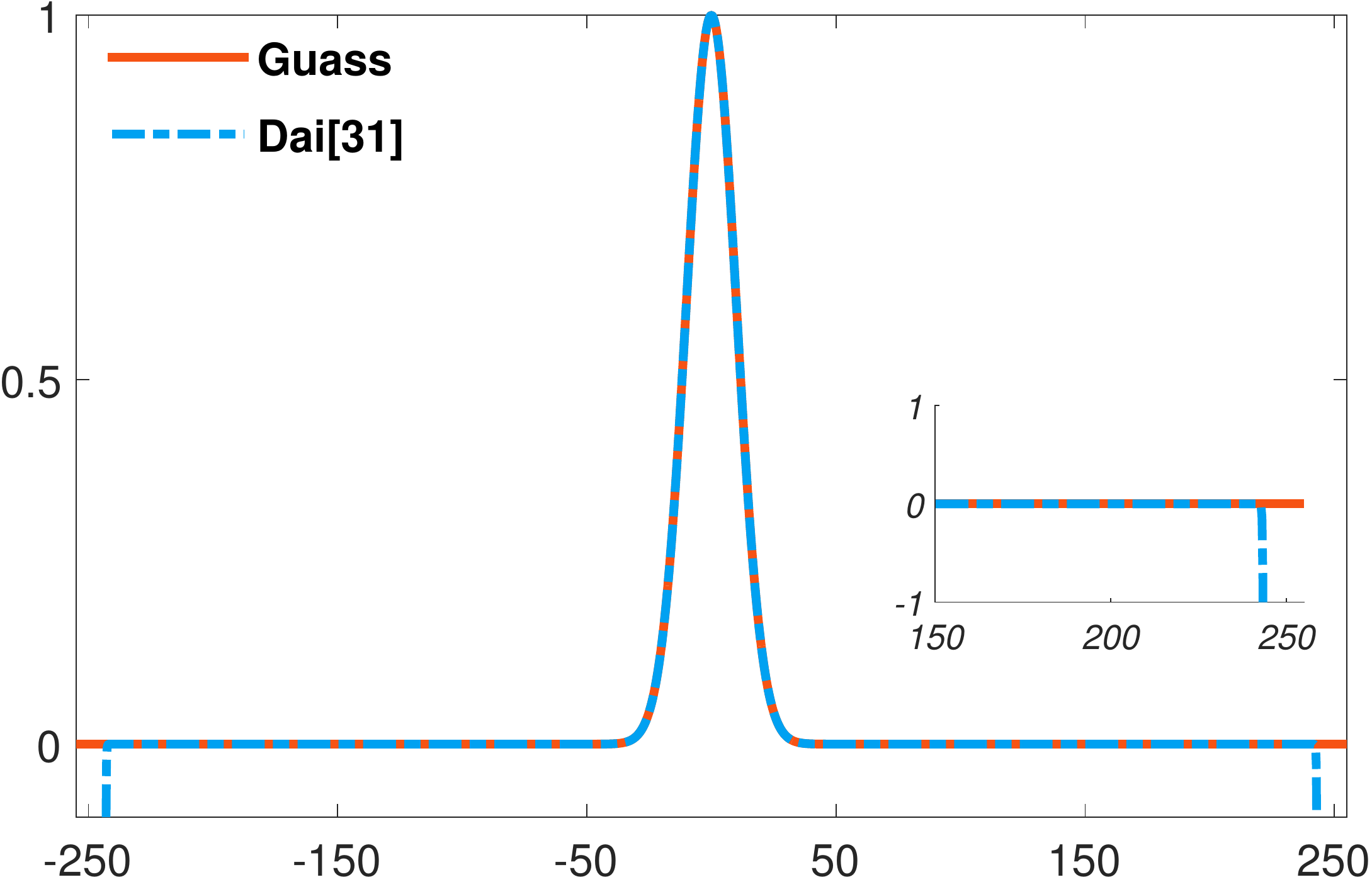}
\caption{Dai~\cite{Dai_EL_2014}}	
\label{fig:shiftable_approxiamtion:Dai}
\end{subfigure}
\end{adjustbox}
&
\begin{adjustbox}{valign=m}
\begin{subfigure}[c]{0.15\linewidth}
\includegraphics[width = \linewidth]{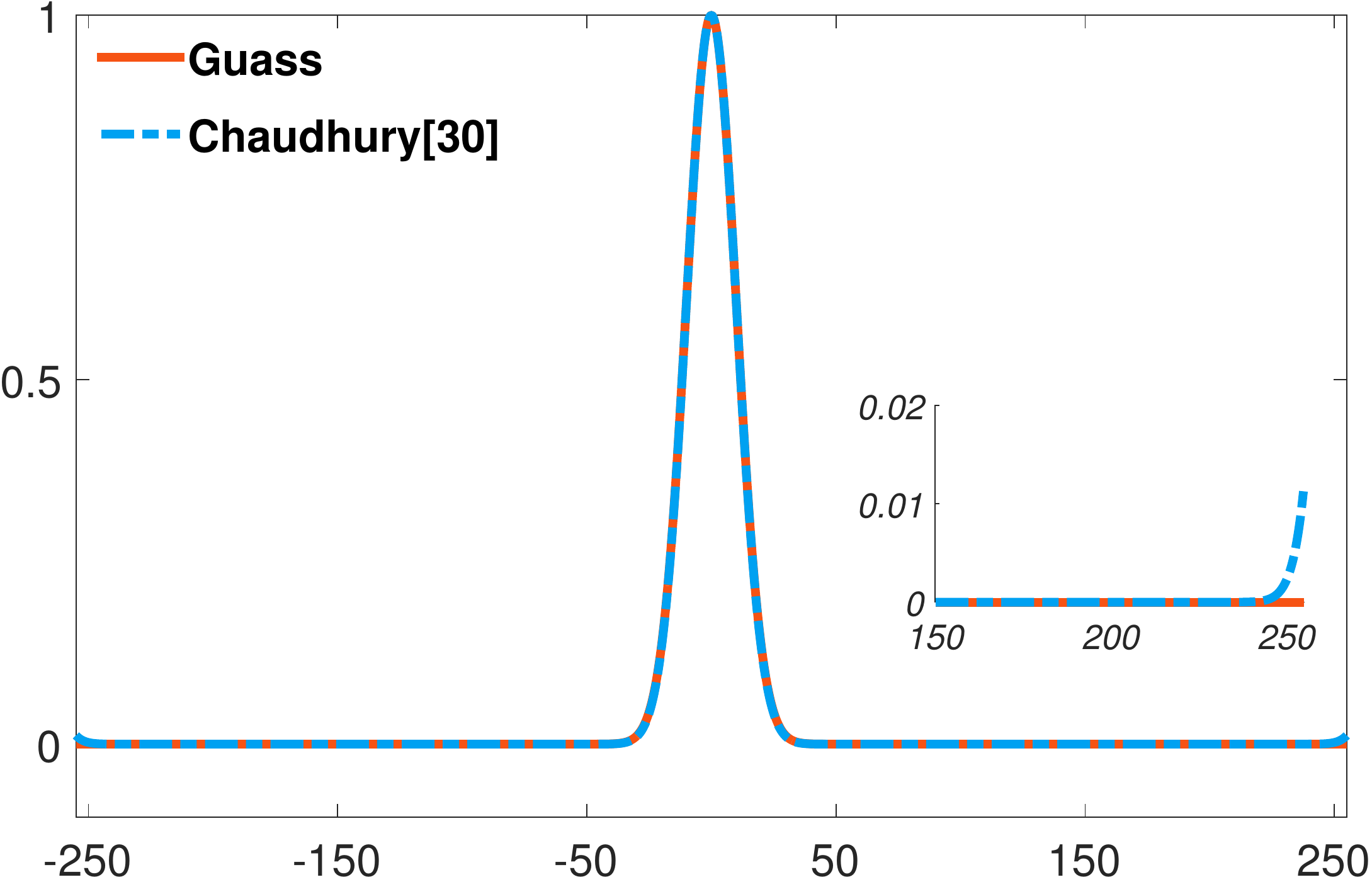}
\caption{Chaudhury~\cite{Chaudhury_TIP_2011}}
\label{fig:shiftable_approxiamtion:Chaudhury1}
\end{subfigure}
\end{adjustbox}
&
\begin{adjustbox}{valign=m}
\begin{subfigure}[c]{0.15\linewidth}
\includegraphics[width = \linewidth]{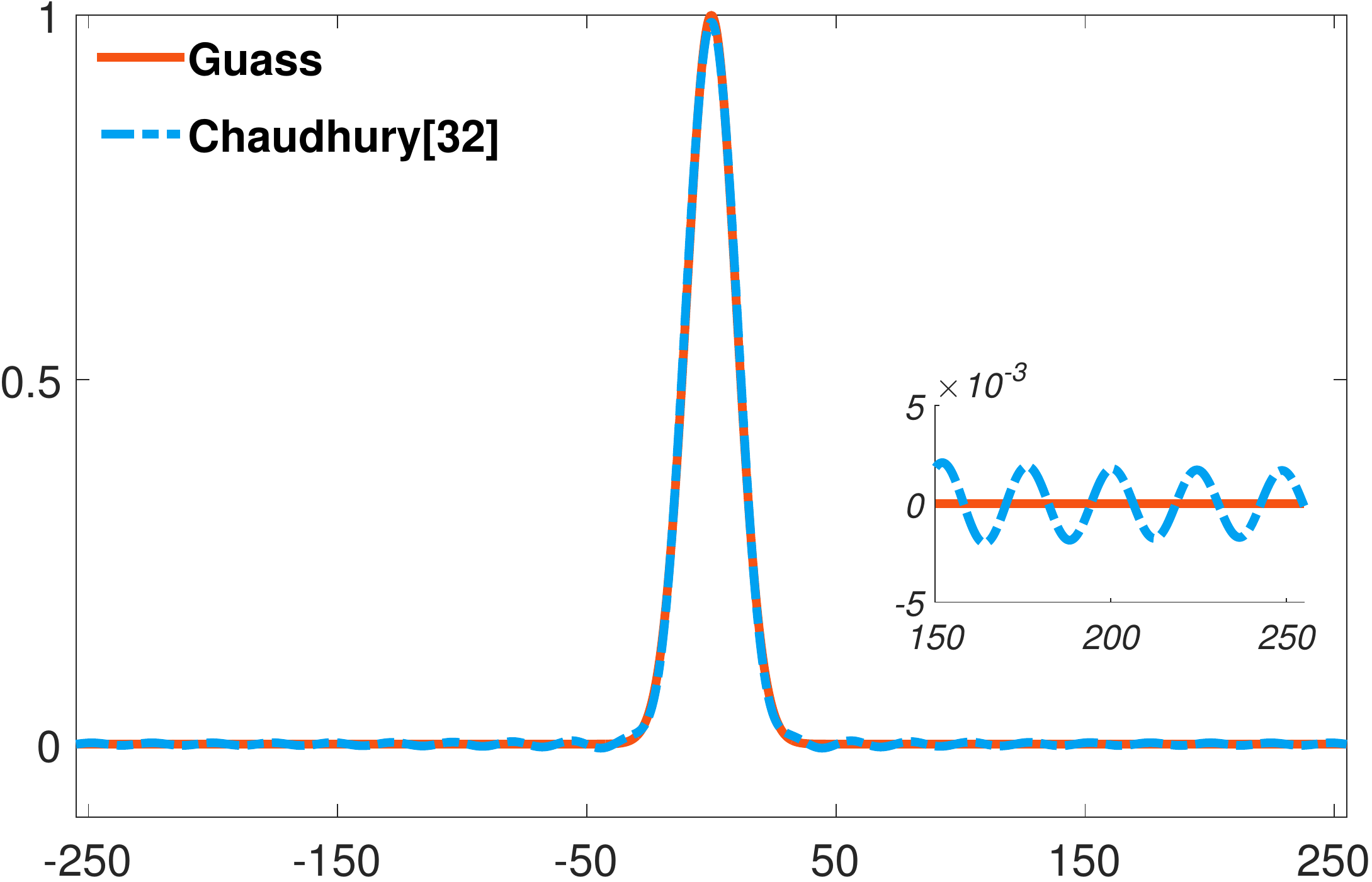}
\caption{Chaudhury~\cite{Chaudhury_TIP_2013}}	
\label{fig:shiftable_approxiamtion:Chaudhury2}
\end{subfigure}
\end{adjustbox}
&
\begin{adjustbox}{valign=m}
\begin{subfigure}[c]{0.15\linewidth}
\includegraphics[width = \linewidth]{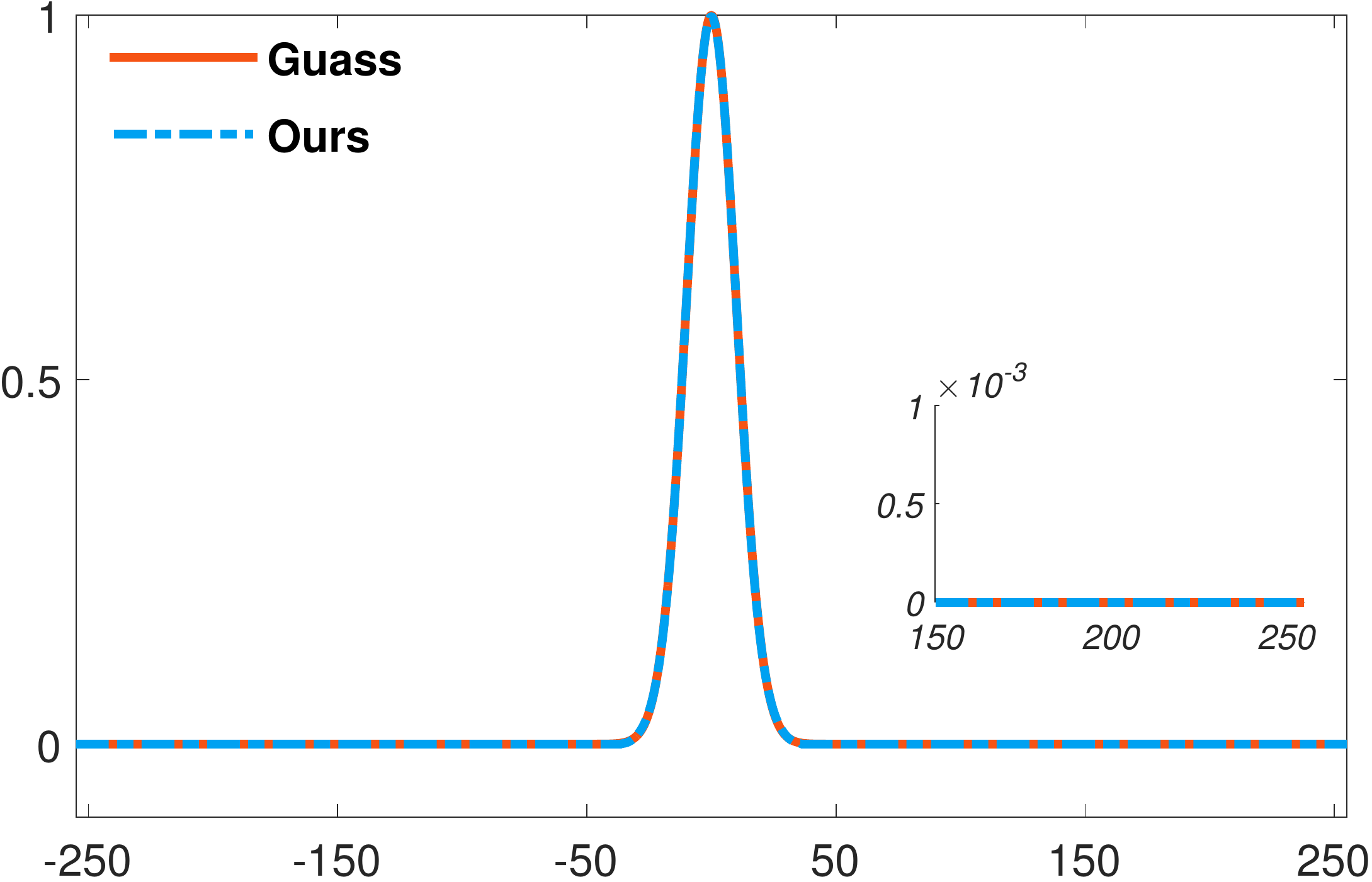}
\caption{Ours}	
\label{fig:shiftable_approxiamtion:Ours}
\end{subfigure}
\end{adjustbox}
\end{tabular}
\caption{Visual and quantitative approximation comparison for the Gaussian range kernel $G_{10}(x)$. (a) illustrates the approximation error measured by SSIM, where x-axis denotes the bandwidth parameter $\sigma_r$ of the Gaussian range kernel, y-axis stands for the number $N$ of approximation terms and z-axis presents the value of SSIM. From (b) to (c), the first row shows the approximation curves of Dai~\cite{Dai_EL_2014},  Chaudhury~\cite{Chaudhury_TIP_2011}, Chaudhury~\cite{Chaudhury_TIP_2013} and ours with $5$ approximation terms for the Gaussian kernel. In the second rows, we keep the SSIM indices of the four methods as 0.99 to find the smallest approximation number $N$ of each method. The  number $N$  of Dai~\cite{Dai_EL_2014},  Chaudhury~\cite{Chaudhury_TIP_2011}, Chaudhury~\cite{Chaudhury_TIP_2013} and ours are 1700, 81, 42 and 5 respectively. }
\label{fig:shiftable_approxiamtion}
\end{figure*}

\subsubsection{Approximation comparison}

Fig~\ref{fig:box_approxiamtion} shows the Gaussian approximation result of our method as well as box filters based approximations such as Zhang~\cite{Zhang_TIP_2012}, Gunturk~\cite{Gunturk_TIP_2011}, Pan~\cite{Pan_MPE_2014}. To approximate the Gaussian function, we use five Haar functions illustrated in Figs~\ref{fig:linear_combinations:1}-\ref{fig:linear_combinations:3}  whereas  Zhang~\cite{Zhang_TIP_2012}, Gunturk~\cite{Gunturk_TIP_2011}, Pan~\cite{Pan_MPE_2014} take three box filters. Unlike the box filters based methods~\cite{Zhang_TIP_2012,Gunturk_TIP_2011,Pan_MPE_2014} employing three box filters to approximate the Gaussian function, our method adopts the linear combination  $f_1(x,y)+f_2(x,y)+f_3(x,y)$ which contains five Haar functions to fulfill the same task. Note that the linear combination of five Haar functions can be synthesized by three box filters and therefore the complexity of them are the same. From Fig~\ref{fig:box_approxiamtion:zhang}-\ref{fig:box_approxiamtion:pan}, we can observe that Zhang,  Gunturk and Pan simply take three cubes and cascade them together to approximate the Gaussian function. In contrast, our approximation in Fig~\ref{fig:box_approxiamtion:ours} is different from theirs. The approximation surface is more complicated than them and the approximation error is the smallest.  Table~\ref{tab:accuracy_box} illustrates the relationship between the approximation quality and the computational complexity. We choose the structural similarity (SSIM) and Peak signal-to-noise ratio (PSNR)  to evaluate the approximation error. In addition, the number $N$ of box filters denotes the computational complexity. Note that it is rational to use the number of box filters to indicate the computational complexity because for fixed $j_1, j_2$, the filtering result of the linear combination of Haar functions is a linear combination of the values of box filtering result at different points. Table~\ref{tab:accuracy_box} clearly indicates that our approximation error is the smallest among the four methods and the decay rate of our approximation is the fastest with respect to the number of approximation terms.

\begin{table}[t]
\caption{Accuracy comparison for box filter based acceleration methods, where the approximation error (or accuracy error)  is measured by SSIM and PSNR and $N$ denotes the number of box filters.}
\label{tab:accuracy_box}
\centering
\begin{tabularx}{\linewidth}{@{}|c|c|Y|Y|Y|Y|@{}}
\hline
\multicolumn{2}{|c|}{}        & Zhang~\cite{Zhang_TIP_2012} & Gunturk~\cite{Gunturk_TIP_2011} & Pan~\cite{Pan_MPE_2014} & Ours        \\ \hline
\multirow{4}{*}{\rotatebox{90}{SSIM}} & $N=1$ & 0.80                        & 0.83                            & 0.84                    & $\bm{0.85}$ \\ \cline{2-6}
                      & $N=2$ & 0.81                        & 0.85                            & 0.87                    & $\bm{0.89}$ \\ \cline{2-6}
                      & $N=3$ & 0.83                        & 0.87                            & 0.88                    & $\bm{0.93}$ \\ \cline{2-6}
                      & $N=4$ & 0.86                        & 0.92                            & 0.94                    & $\bm{0.97}$ \\ \hline
\multirow{4}{*}{\rotatebox{90}{PSNR}} & $N=1$ &       23.16       &        24.69              &    25.01          &     $\bm{25.33 }$        \\ \cline{2-6}
                      & $N=2$ &    23.31                    &    25.32                        &   27.01              &    $\bm{27.93}$        \\ \cline{2-6}
                      & $N=3$ &    24.68                    &    27.03                             &   27.23              &    $\bm{35.22}$         \\ \cline{2-6}
                      & $N=4$ &    25.91                    &     33.12                            &   36.76              &    $\bm{42.16}$         \\ \hline
\end{tabularx}
\end{table}

Other than box filter based acceleration methods, we also compare our approach with other spatial kernel acceleration algorithms such as Rachid~\cite{Rachid_TR_1993} which belongs to the kernel separation method. However, this kind of methods cannot perform satisfactorily on texture regions. We can observe these artifacts produced by the kernel separation method in Fig~\ref{fig:Duck}.



\subsection{Comparison with the acceleration techniques of $K_r(x)$}
\label{sec:comparison_Kr}

Decomposing the nonlinear convolution of $K_r(x)$ into a set of linear convolutions is the key step to speed up BF. In this section, we compare our acceleration approach with previous techniques. Without loss of generality, we assume $K_s(\|\bm{x}\|)$ is a 2-D box function in the following paragraphs. 

\begin{table}[b]
\caption{Computational complexity and accuracy for comparison between the dimension promotion based method (Porikli) in~\cite{Porikli_CVPR_2008} and our method}
\label{tab:promotion_ca}
\centering
\begin{tabularx}{\linewidth}{@{}|Y|Y|Y|Y|Y|@{}}
\hline
\multirow{2}{*}{} & \multirow{2}{*}{Porikli} & \multicolumn{3}{c|}{Ours} \\ \cline{3-5}
                  &                            & 3   & 4   & 5   \\ \hline
Add          &         $256|I|$         &  $12T_r|I|$   &  $16T_r|I|$   &  $20T_r|I|$      \\ \hline
Mul    &         $256|I|$         &   $6|I|$  &   $8|I|$  &    $10|I|$   \\ \hline
SSIM              &         0.96                 &  0.95   &  0.97   &  0.99   \\ \hline
PSNR              &         40.21                 &  38.97   &  40.32   &  44.53   \\ \hline
Time              &         1.63s                 &   1.35s  &  1.36s   &  1.37s   \\ \hline
\end{tabularx}
\end{table}

\begin{table*}[t]
\caption{Technique Summary for BF accelerating methods.}
\label{tab:summary}
\centering
\begin{tabularx}{\linewidth}{@{}ccYcYcYY@{}}
\hline
 &  Kernel separation &  Box filtering & Dimension promotion & PBFIC &  Shiftability property &   Best $N$-term  & Truncated kernels \\ \hline
Porikli~\cite{Porikli_CVPR_2008}  & \ding{56} & \ding{52} & \ding{52} & \ding{56} & \ding{56} & \ding{56} & \ding{56}\\
Yang~\cite{Yang_CVPR_2009} & \ding{52} & \ding{56} & \ding{56} & \ding{52} & \ding{56} & \ding{56} & \ding{56}\\
Gunturk~\cite{Gunturk_TIP_2011}  & \ding{56} & \ding{52} & \ding{52} & \ding{56} & \ding{56} & \ding{56} & \ding{56}\\
Pan~\cite{Pan_MPE_2014}  & \ding{56} & \ding{52} & \ding{52} & \ding{56} & \ding{56} & \ding{56} & \ding{56}\\
Chaudhury~\cite{Chaudhury_TIP_2013}  & \ding{56} & \ding{56} & \ding{56} & \ding{56} & \ding{52} & \ding{56} & \ding{56}\\
Ours  & \ding{56} & \ding{52} & \ding{52} & \ding{56} & \ding{52} & \ding{52} & \ding{52}\\ \hline
\end{tabularx}
\end{table*}

\subsubsection{Dimension promotion} Both our method and the dimension promotion based algorithm employ the dimension promotion technique to eliminate the dependency between $I(\bm{y})$ and $I(\bm{x})$ in the kernel $K_r(I(\bm{x}) - I(\bm{y}))$ and $\dot{B}(I(\bm{x}) - I(\bm{y}))$ according to \eqref{eq:BF_promotion}  \eqref{eq:sat_sum_3D_1}~\eqref{eq:sat_sum_3D_2} \eqref{eq:numerator_2D_box_N_terms}. Here $\dot{B}(\bm{x})$ is a box function with the support region $[-T_r, T_r]$. The response of BF can be computed by first performing  box filtering on the auxiliary images $F(\bm{y}, z)$ (or $F_c(\bm{y}, z)$, $F_s(\bm{y}, z)$) for each fixed $z$ and then calculating the sum weighted by $K_r(I(\bm{y}) - z)$ (or $\dot{B}(I(\bm{y}) - z)$) along $z$.

The major difference between the dimension promotion based algorithms and ours is the computational complexity of the second step as all these methods perform box filtering at the first step. Specifically, let $g(z)$ be a scalar function.  For an 8-bit image, $\sum_{z=0}^{255} K_r(I(\bm{x}) - z) g(z)$ needs $256$ multiplications and $255$ additions. In contrast, employing the 1-D SAT along the $z$ axis, we only require $255$ additions and $1$ subtraction to compute $\sum_{z=0}^{255} \dot{B}(I(\bm{x}) - z) g(z) = \sum_{z \in \mathcal{N}_{I(\bm{x})}} g(z)$.  Note that the floating point addition or subtraction requires $6$ clock cycles, multiplication or division require 30-44 clock cycles on the Intel x86 processor. Due to the high running cost for multiplication operation, our method can significantly decrease the run time. Indeed, we can cut down the cost further because for each fixed $\bm{x}$, we only need to compute the box filtering result at $I(\bm{x})$ along the z axis. Hence in this situation the 1-D SAT degrade to the sliding window summation and therefore we can reduce the running cost to $2T_r$ additions.

Our approach is not problem-free. The biggest shortcoming is that our method can only give an approximation filtering result. Fortunately, the approximation error is very small and the accuracy will be improved when we take more terms to approximate the range kernel $K_r(x)$. In Table~\ref{tab:promotion_ca}, we list the overall computational cost of filtering an image $I$ as well as the SSIM and PSNR indices that measure the similarity between the filtering image and the ground truth, where $|I|$ represents the pixel number of the image $I$. Note that the multiplications in our method are caused by the multiplication of $a_k$ in \eqref{eq:1D_box_N_terms} and Table~\ref{tab:promotion_ca} does not take into account the complexity of linear convolution of $K_s(x)$  because both our approach and the dimension promotion based algorithm perform 255 linear convolutions. Compared with the dimension promotion based algorithm, the number of multiplications is very small and thus can be neglected. The approximation error almost vanishes when we use three approximation terms. We can verify this from the SSIM and PSNR indices and the approximation error map shown in Fig~\ref{fig:shiftable_approxiamtion}. More importantly, the run time is still smaller than the dimension promotion based algorithm.


%


\renewcommand{\tabcolsep}{0.05cm}
\begin{figure}[t]
\begin{tabular}{cccc}
\multirow{2}{*}{\begin{adjustbox}{valign=m}
\begin{subfigure}[c]{0.25\linewidth}
\vspace{-0.7cm}
\includegraphics[width = \linewidth]{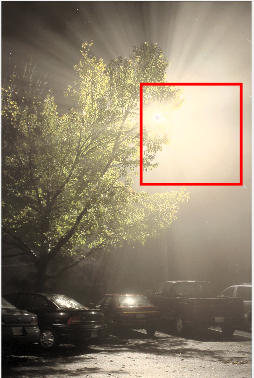}
\caption{Input}	
\end{subfigure}
\end{adjustbox}} & \multirow{2}{*}{\begin{adjustbox}{valign=m}
\begin{subfigure}[c]{0.25\linewidth}
\vspace{-0.7cm}
\includegraphics[width = \linewidth]{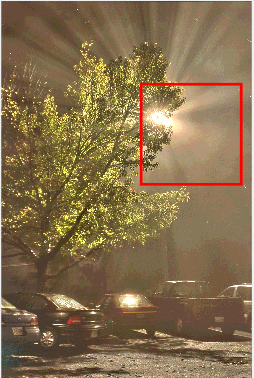}
\caption{PBFIC}	
\end{subfigure}
\end{adjustbox}} & \multirow{2}{*}{\begin{adjustbox}{valign=m}
\begin{subfigure}[c]{0.25\linewidth}
\vspace{-0.7cm}
\includegraphics[width = \linewidth]{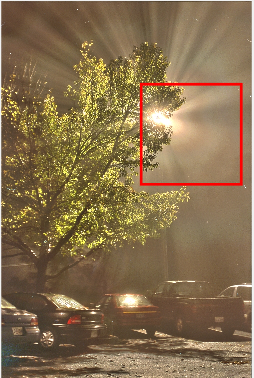}
\caption{Ours}	
\end{subfigure}
\end{adjustbox}} &
\begin{adjustbox}{valign=m}
\begin{subfigure}[c]{0.18\linewidth}
\includegraphics[width = \linewidth]{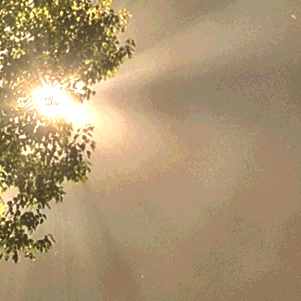}
\end{subfigure}
\end{adjustbox}\vspace{0.1cm}
\\
                  &                   &                   &
\begin{adjustbox}{valign=m}
\begin{subfigure}[c]{0.18\linewidth}
\includegraphics[width = \linewidth]{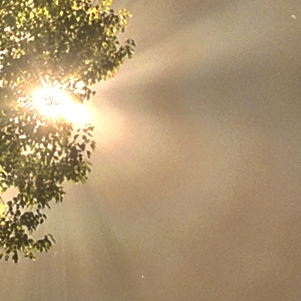}
\caption{Close up}	
\end{subfigure}
\end{adjustbox}
\end{tabular}
\caption{Quantization artifacts demonstration. (a) Input HDR image (32 bit float, displayed by linear scaling). (b) Compressed image using 32 PBFIC (32 bins). (c) Compressed image using our method. (d) Zoom-in of the square in (b) as the upper image and that in (c) as the lower one.}
\label{fig:HDR}
\end{figure}

\subsubsection{Principle bilateral filtered image component (PBFIC)} This kind of methods is equivalent to the dimension promotion based methods if we compute 256 PBFICs as they do not need to interpolate missing values in this situation. Hence it is reasonable to say that these methods speed up BF by employing downsampling and interpolation operation to reduce the computational burden. The major shortcoming is the large approximation error and it is the origin of stepwise artifacts which degrade the quality of filtering results for HDR image intensively as illustrated in Fig~\ref{fig:HDR}. In contrast, our method does not suffer from the problem.

\begin{figure*}[t]
\centering
\begin{subfigure}[b]{0.23\linewidth}
     \includegraphics[width=\textwidth ]{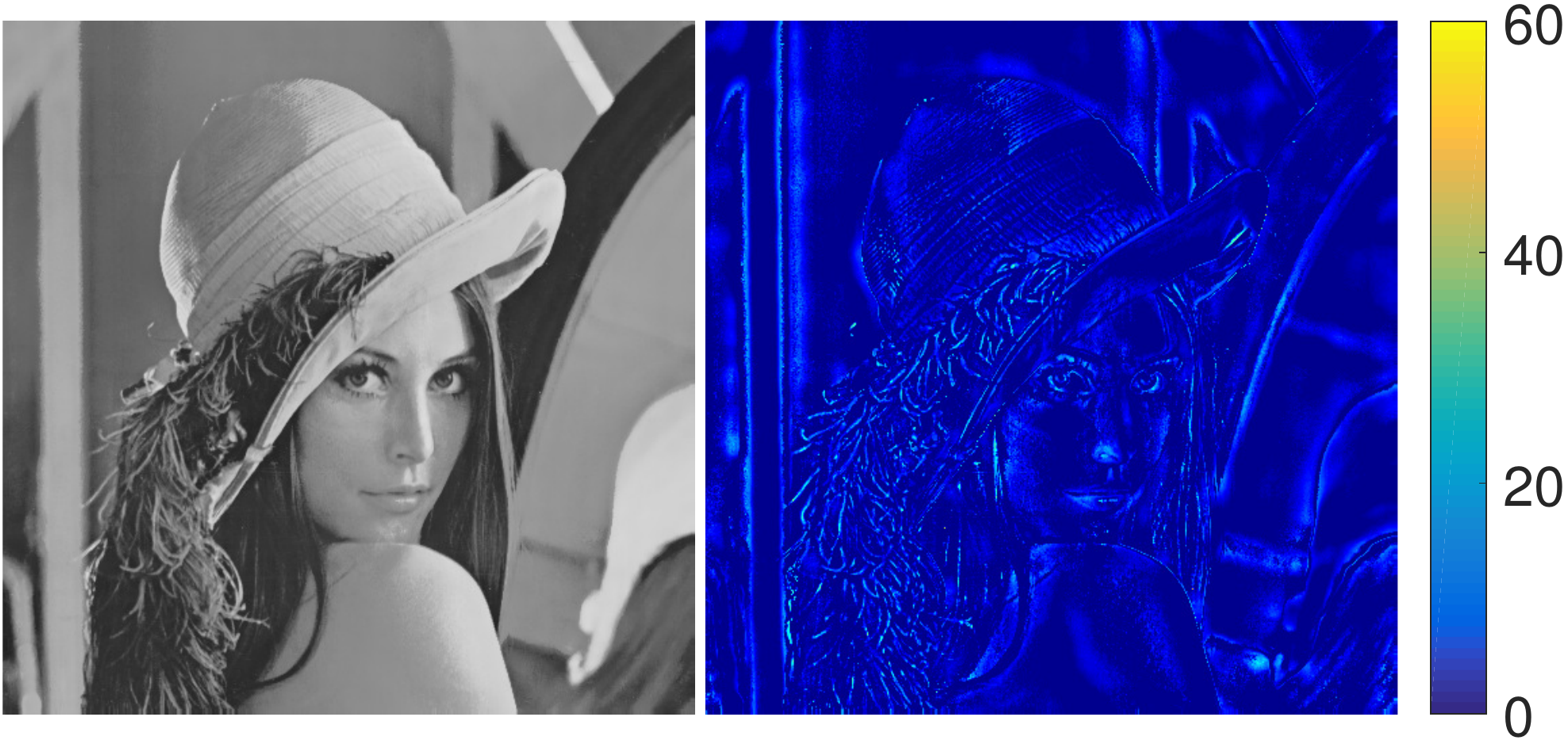}
\end{subfigure}
\begin{subfigure}[b]{0.23\linewidth}
     \includegraphics[width=\textwidth ]{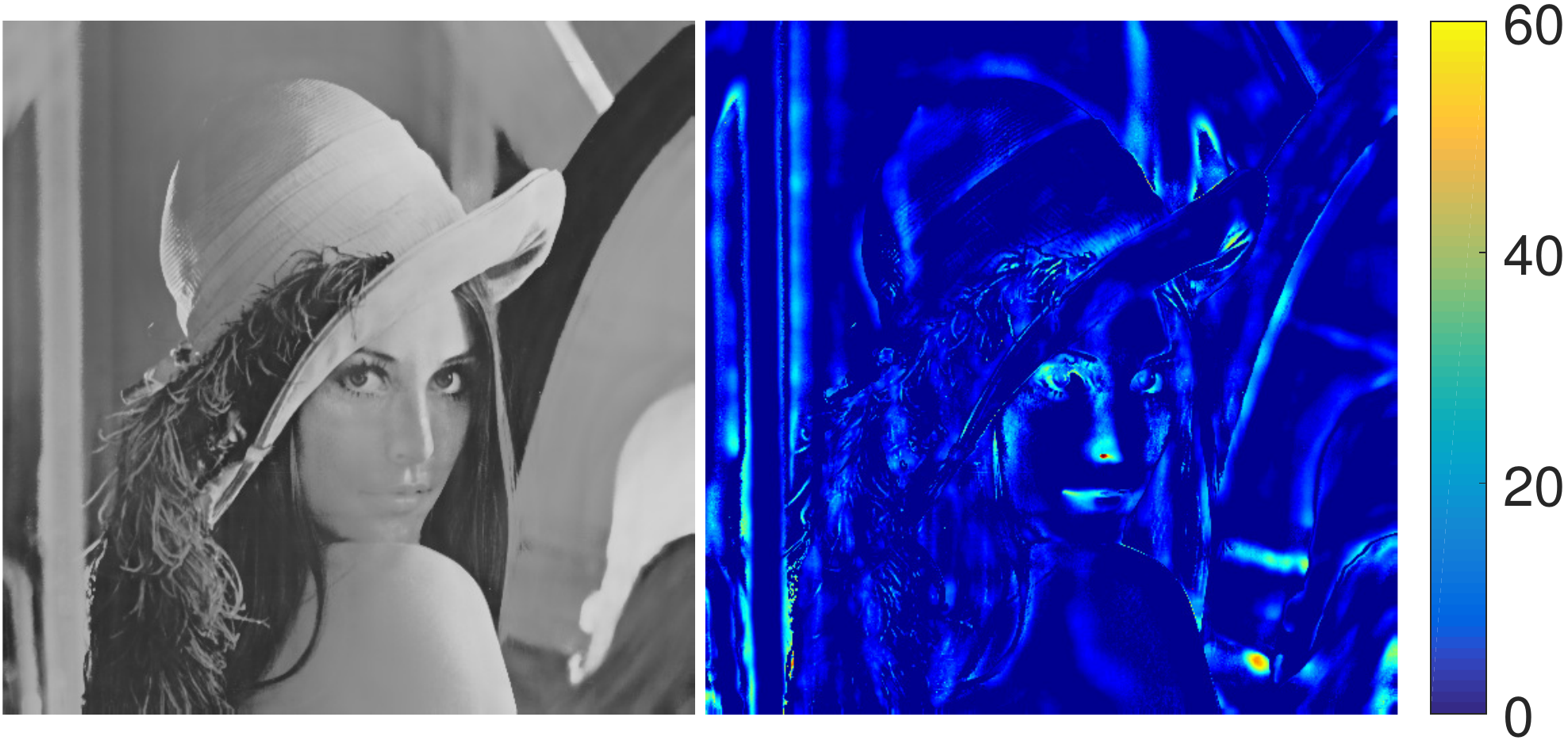}
\end{subfigure}
\begin{subfigure}[b]{0.23\linewidth}
     \includegraphics[width=\textwidth ]{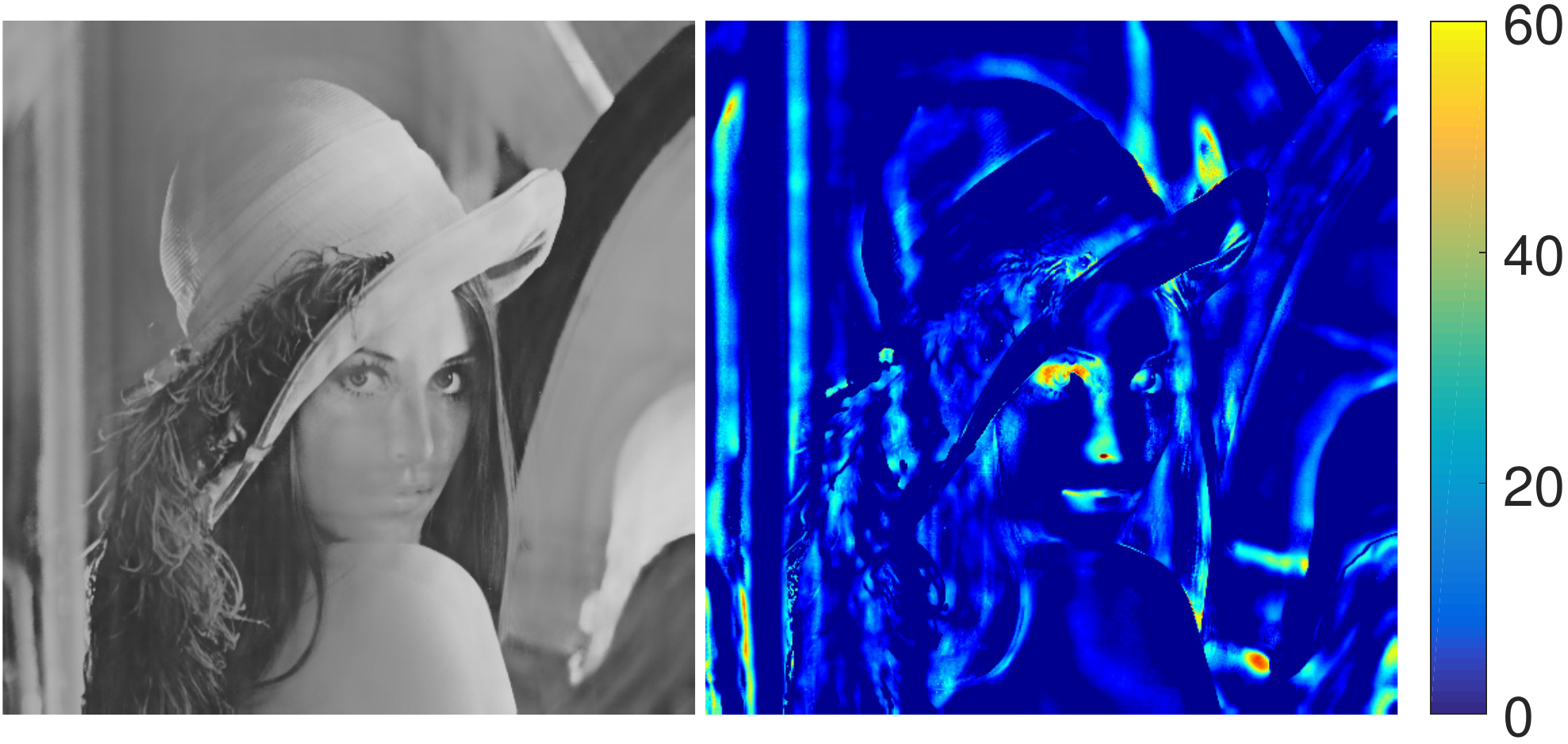}
\end{subfigure}
\begin{subfigure}[b]{0.23\linewidth}
     \includegraphics[width=\textwidth ]{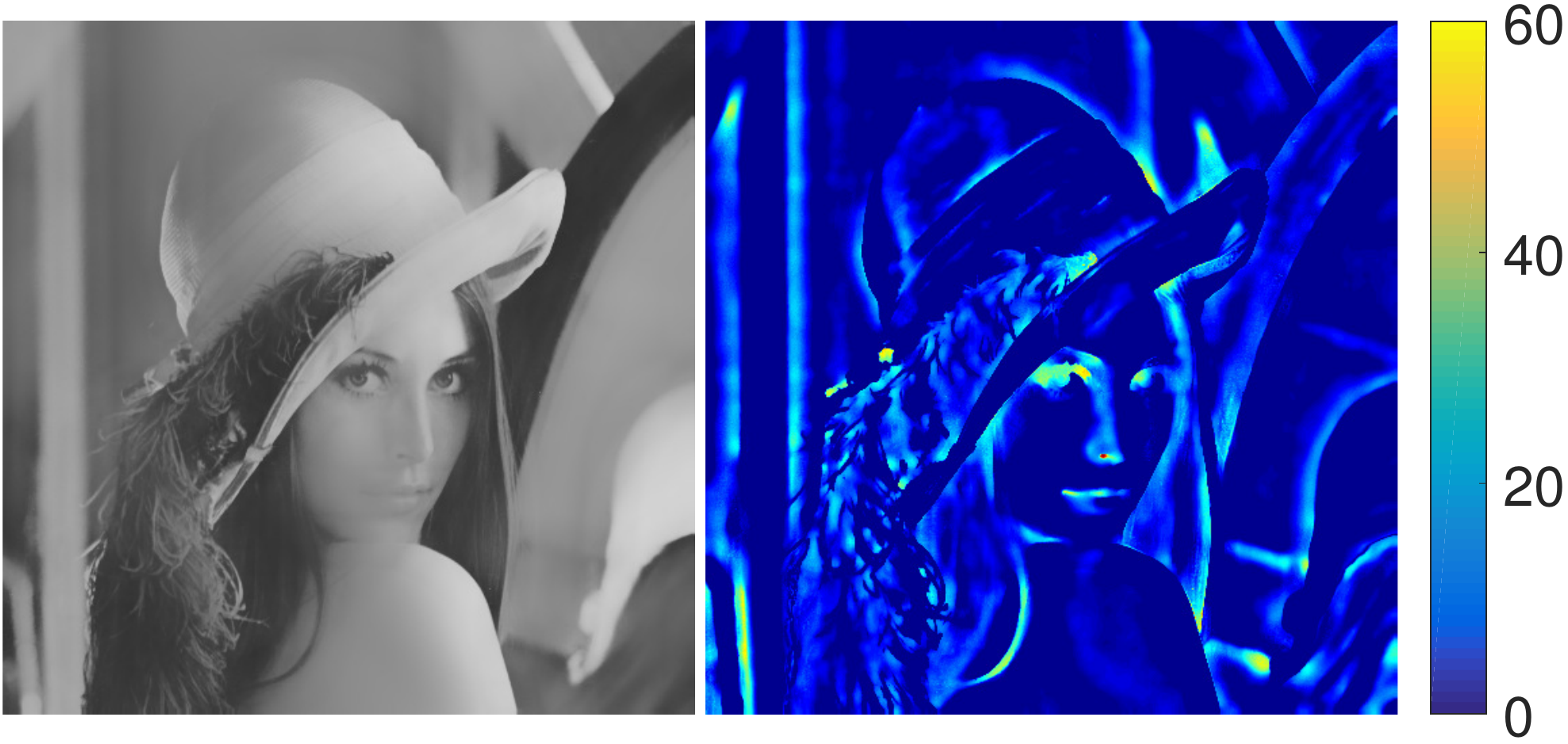}
\end{subfigure}
\rotatebox{90}{\footnotesize \  Porikli~\cite{Porikli_CVPR_2008}}

\begin{subfigure}[b]{0.23\linewidth}
     \includegraphics[width=\textwidth ]{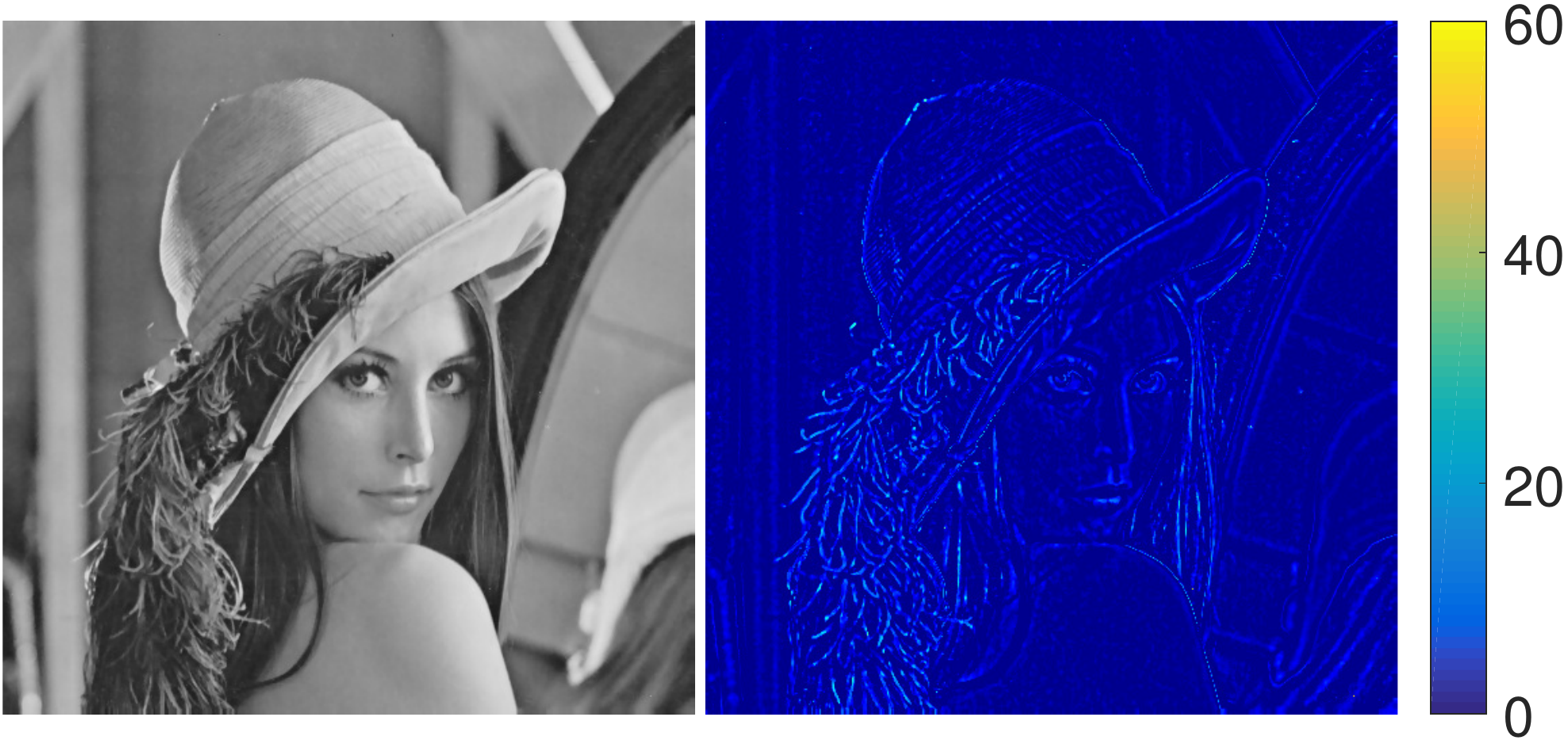}
\end{subfigure}
\begin{subfigure}[b]{0.23\linewidth}
     \includegraphics[width=\textwidth ]{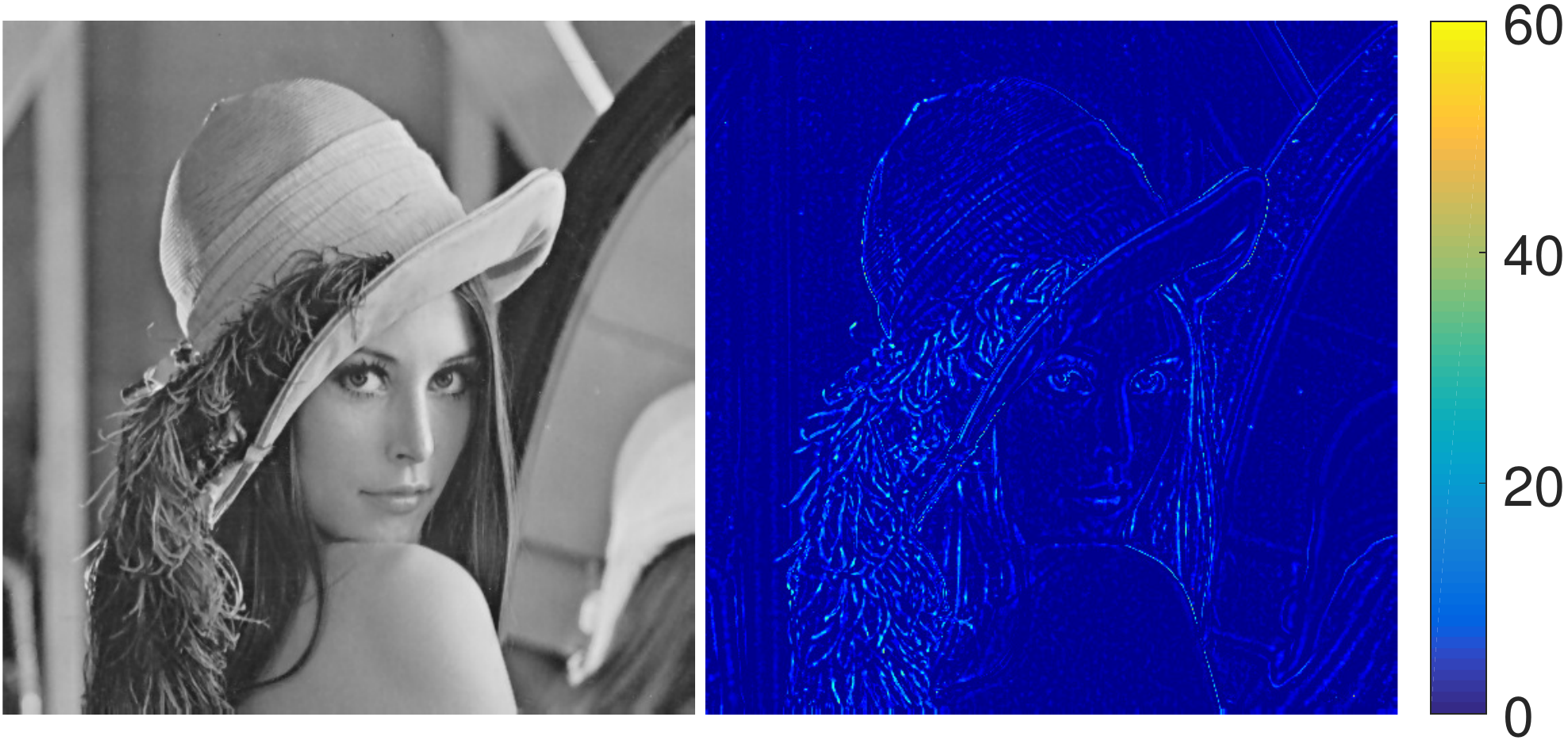}
\end{subfigure}
\begin{subfigure}[b]{0.23\linewidth}
     \includegraphics[width=\textwidth ]{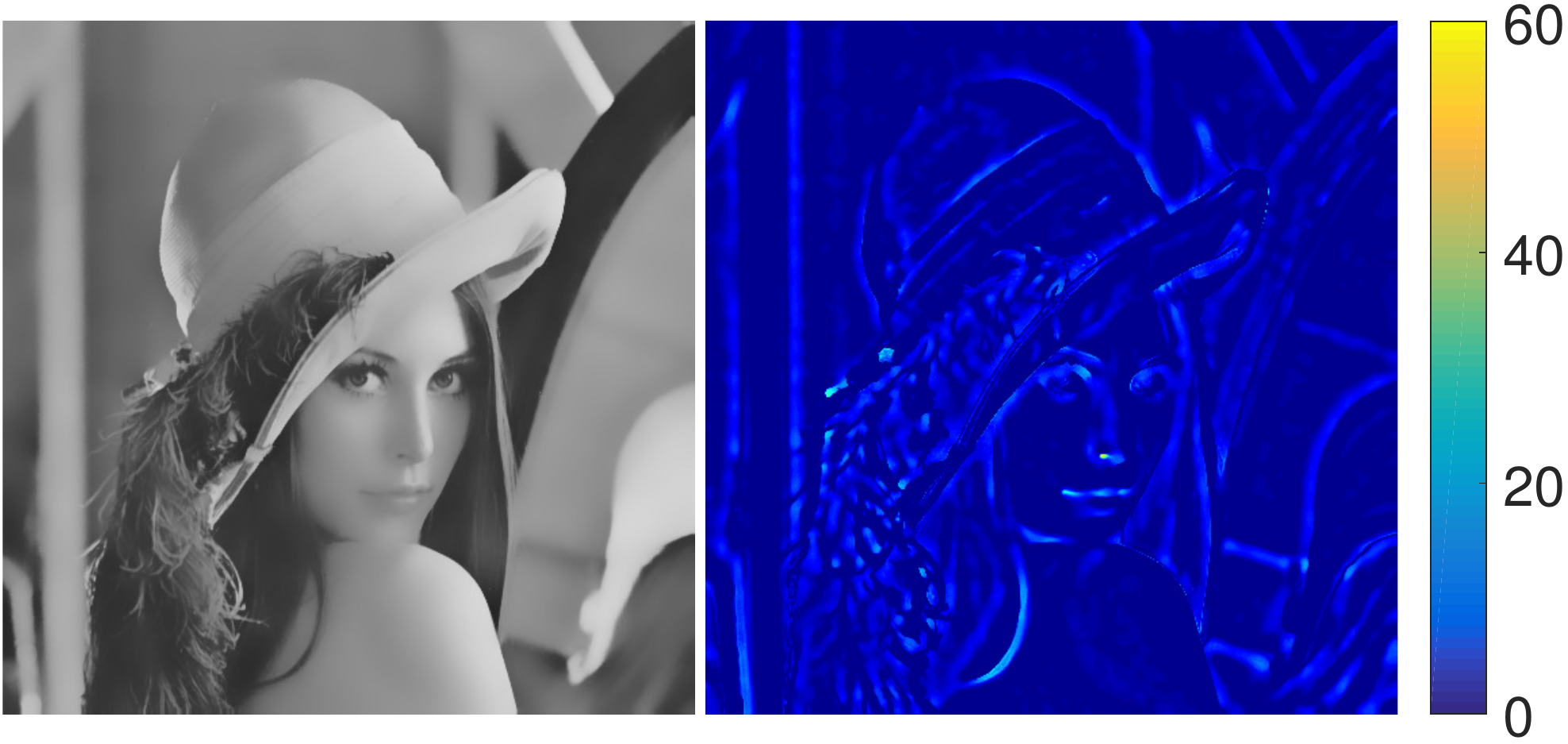}
\end{subfigure}
\begin{subfigure}[b]{0.23\linewidth}
     \includegraphics[width=\textwidth ]{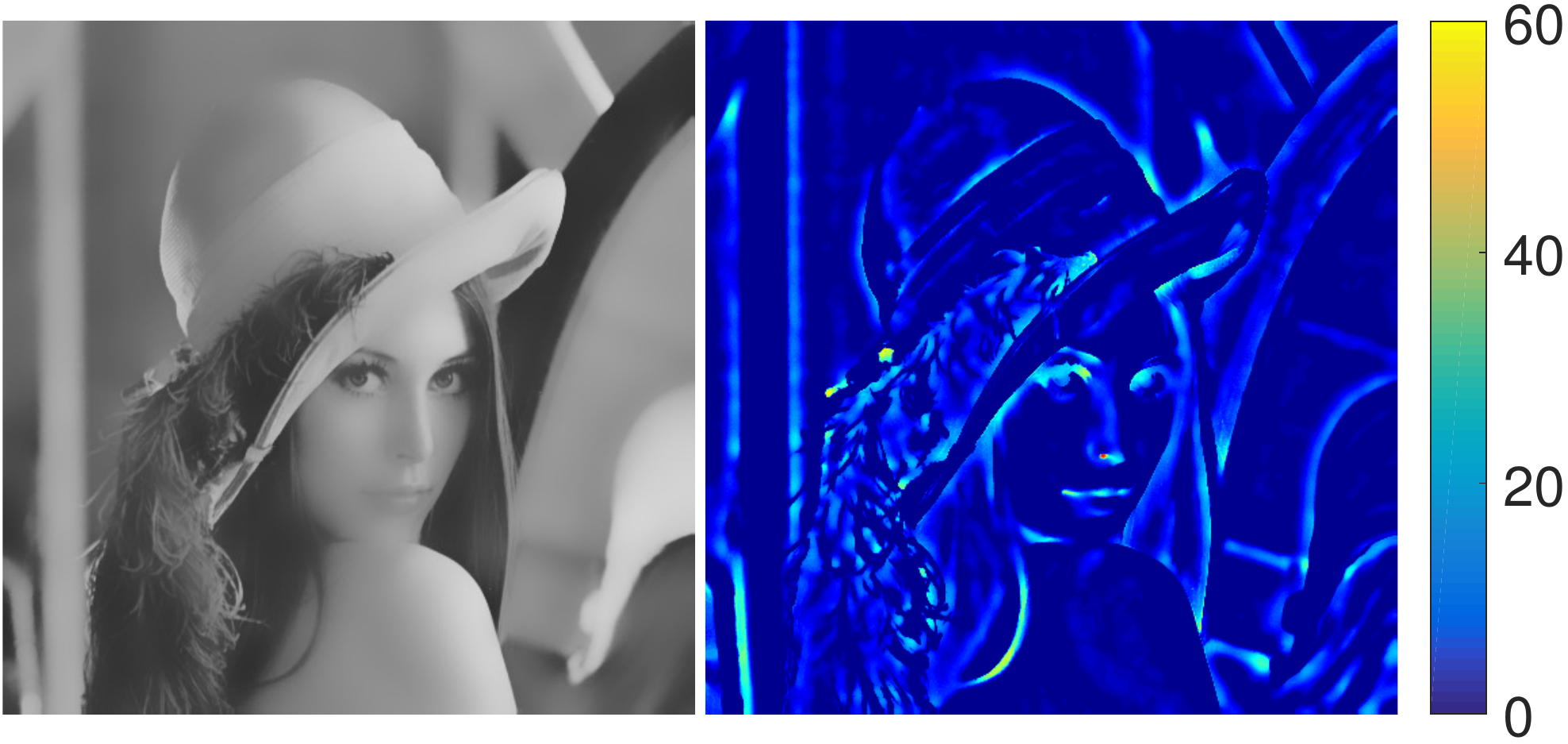}
\end{subfigure}
\rotatebox{90}{\footnotesize \ \ \   Yang~\cite{Yang_CVPR_2009}}

\begin{subfigure}[b]{0.23\linewidth}
     \includegraphics[width=\textwidth ]{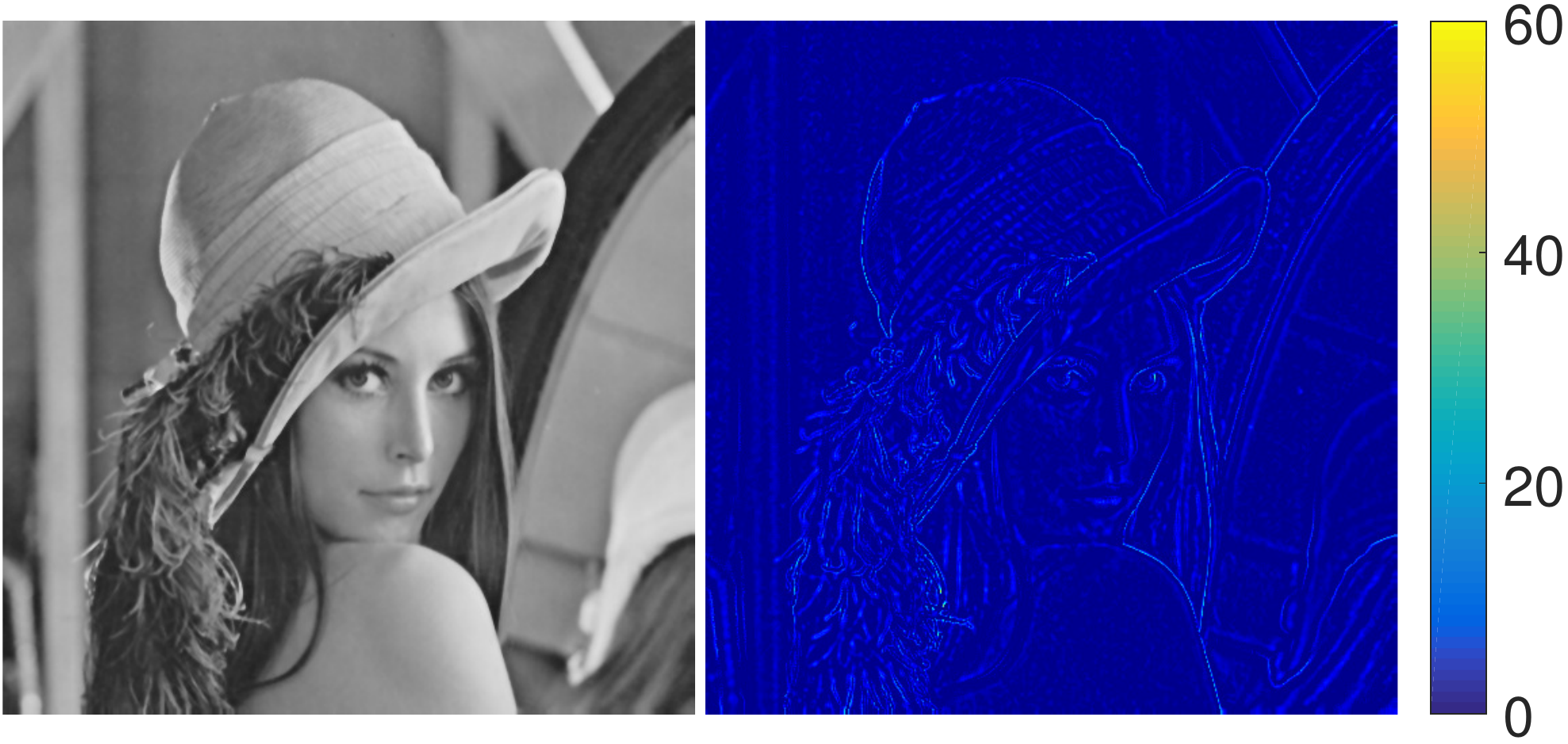}
\end{subfigure}
\begin{subfigure}[b]{0.23\linewidth}
     \includegraphics[width=\textwidth ]{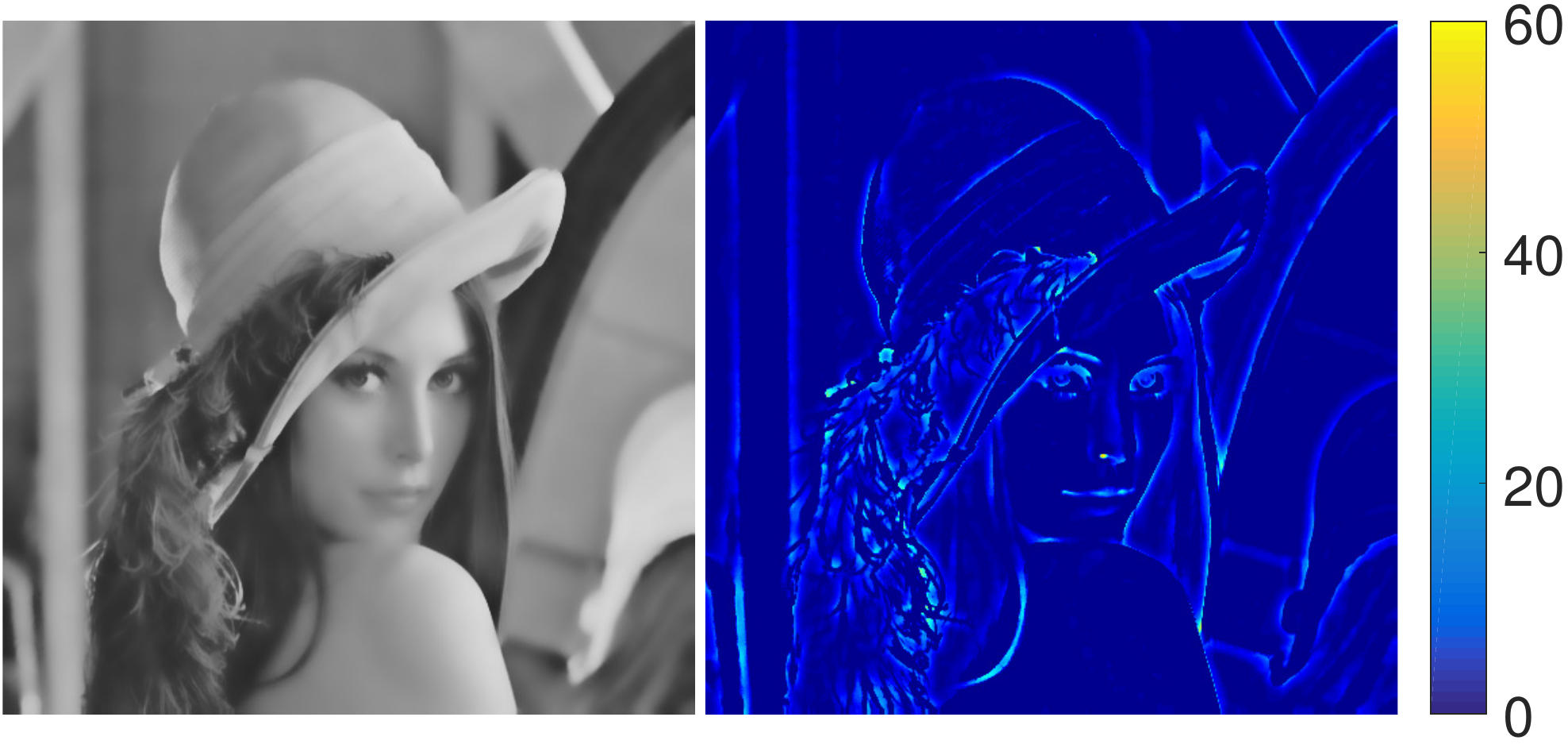}
\end{subfigure}
\begin{subfigure}[b]{0.23\linewidth}
     \includegraphics[width=\textwidth ]{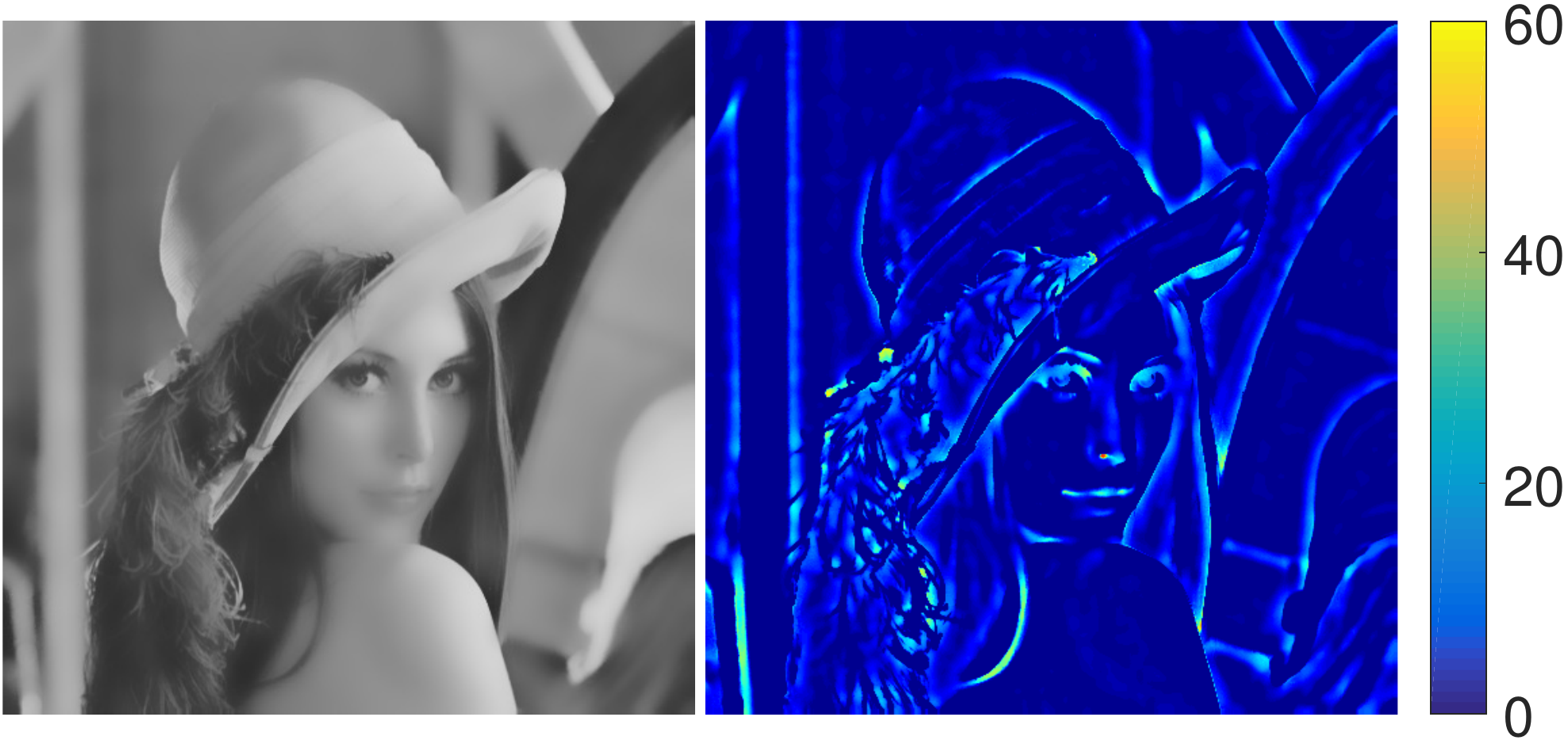}
\end{subfigure}
\begin{subfigure}[b]{0.23\linewidth}
     \includegraphics[width=\textwidth ]{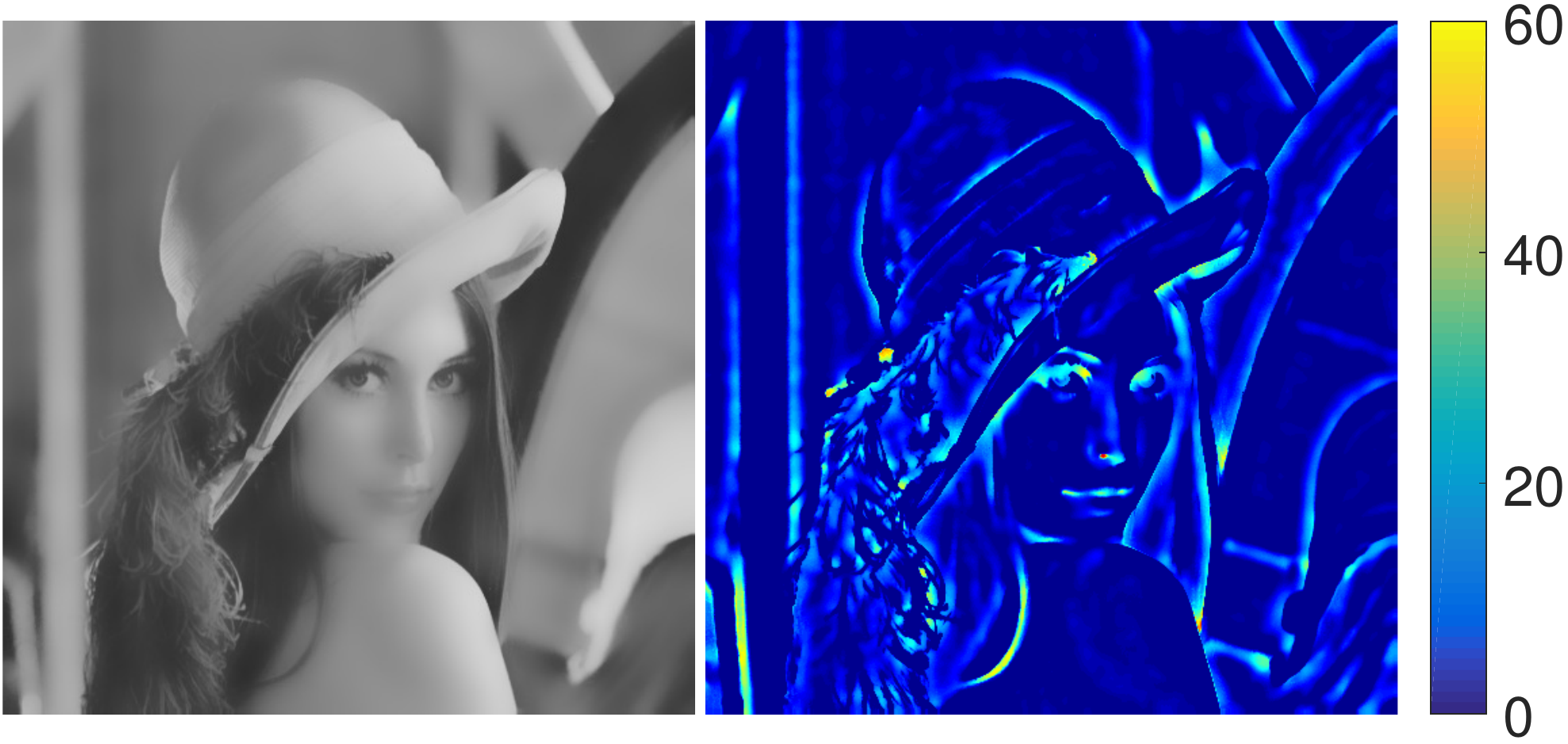}
\end{subfigure}
\rotatebox{90}{\footnotesize \ \ \   Gunturk~\cite{Gunturk_TIP_2011}}

\begin{subfigure}[b]{0.23\linewidth}
     \includegraphics[width=\textwidth ]{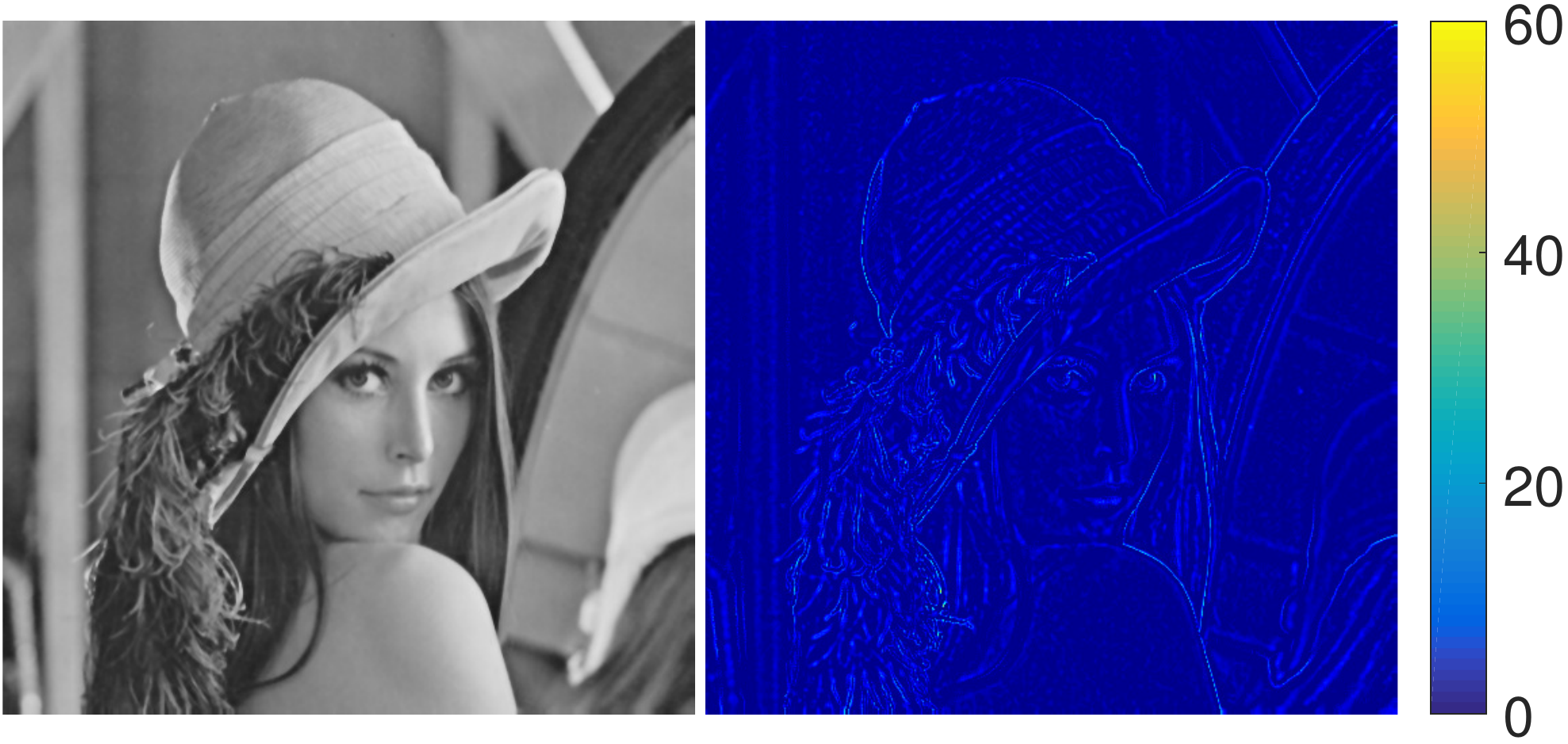}
\end{subfigure}
\begin{subfigure}[b]{0.23\linewidth}
     \includegraphics[width=\textwidth ]{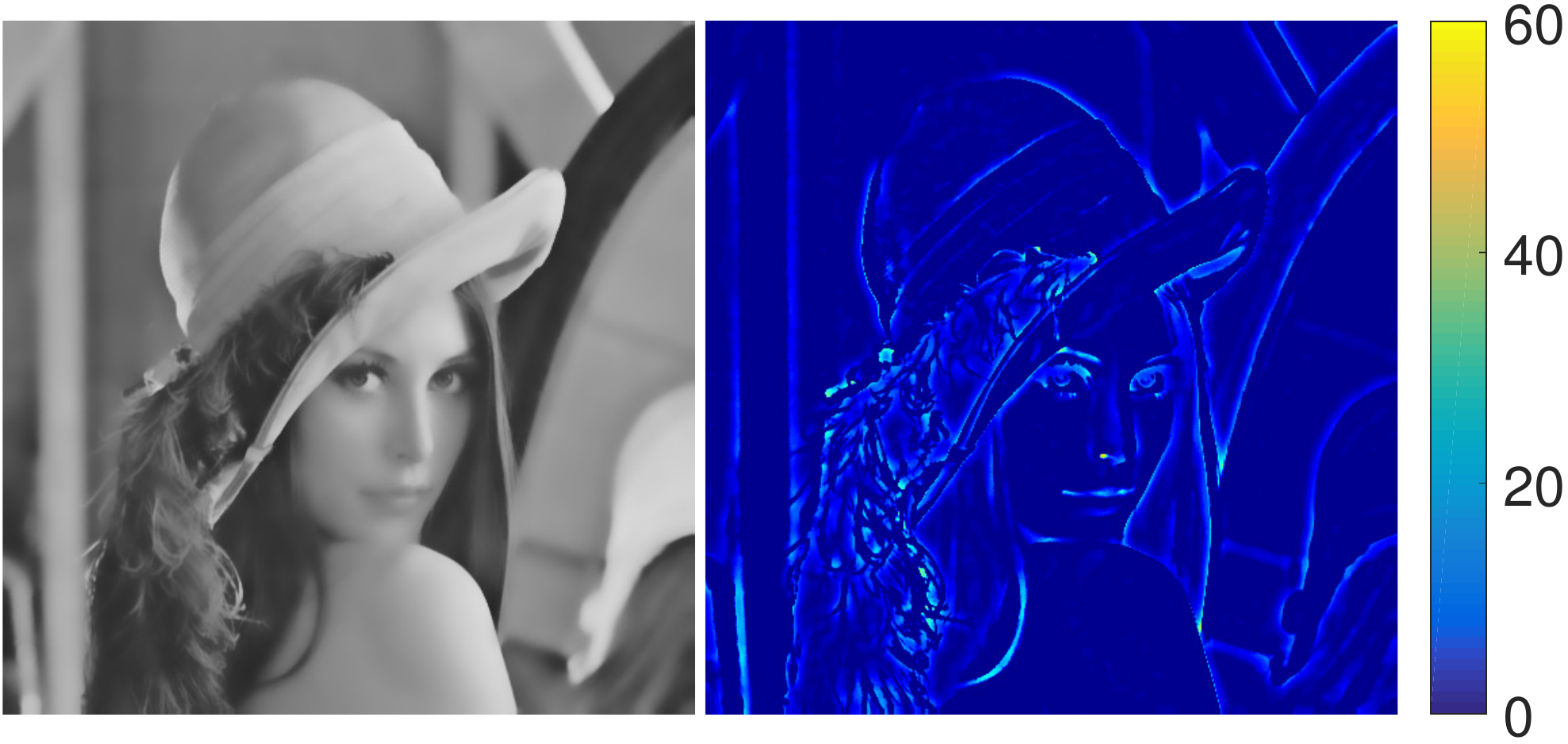}
\end{subfigure}
\begin{subfigure}[b]{0.23\linewidth}
     \includegraphics[width=\textwidth ]{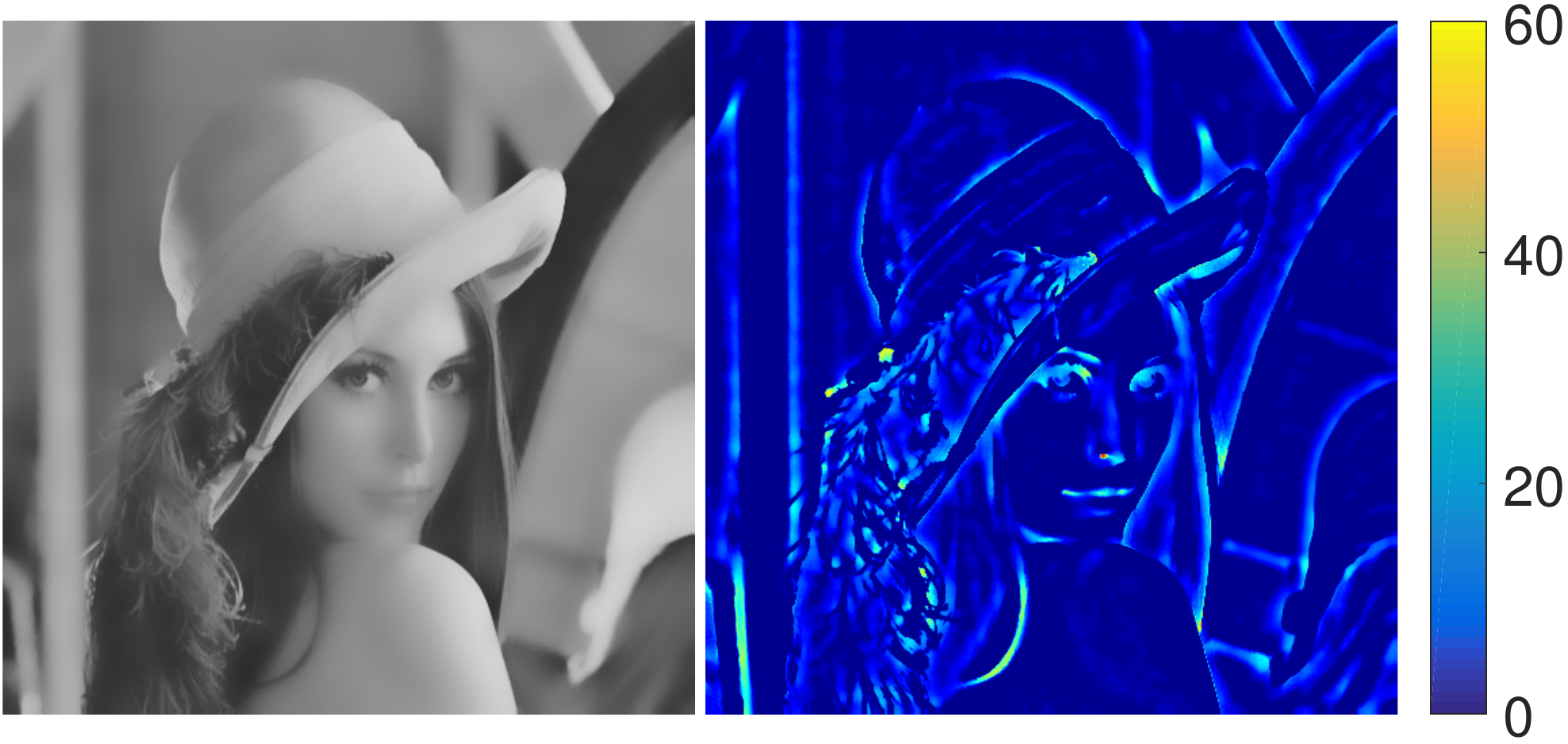}
\end{subfigure}
\begin{subfigure}[b]{0.23\linewidth}
     \includegraphics[width=\textwidth ]{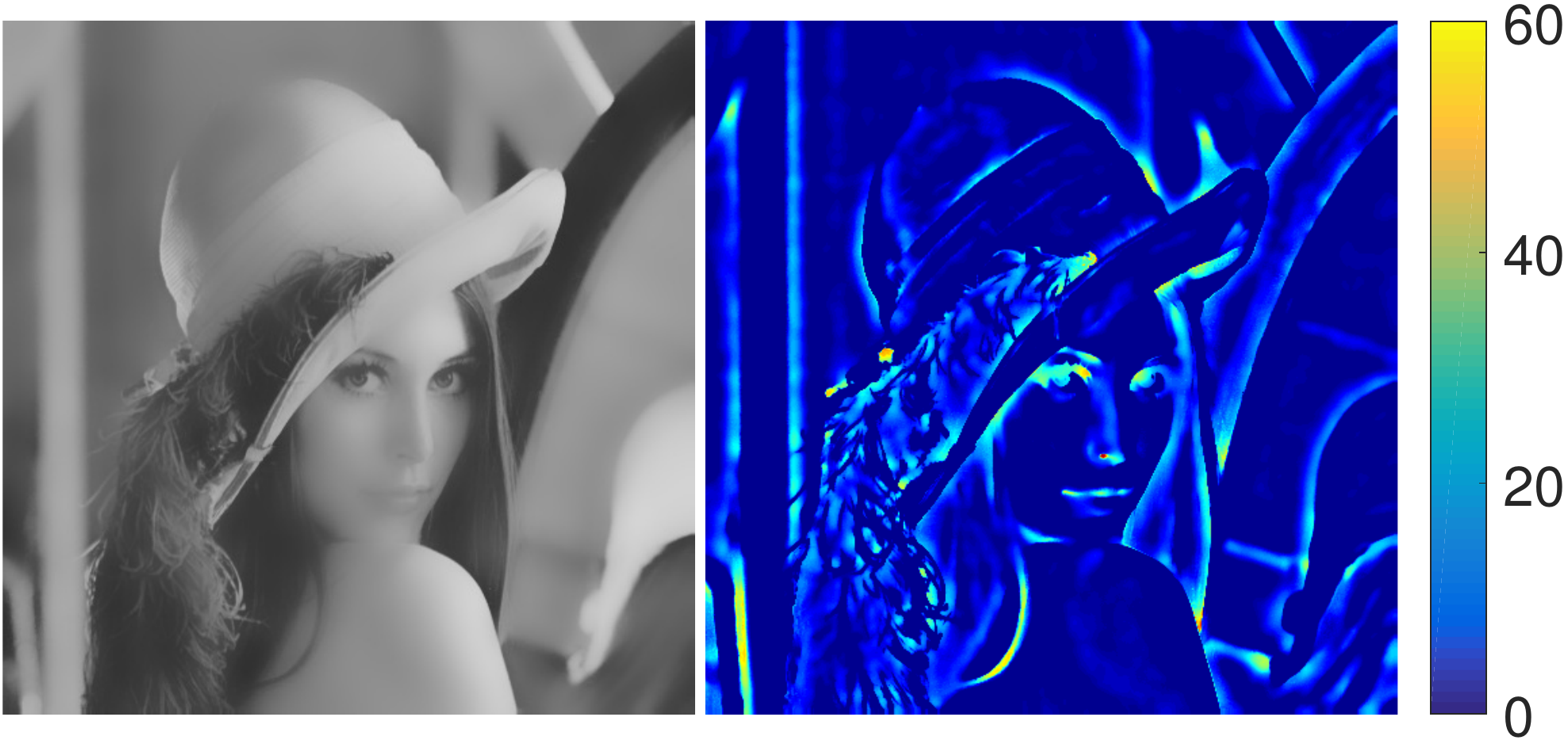}
\end{subfigure}
\rotatebox{90}{\footnotesize \qquad  Pan~\cite{Pan_MPE_2014}}

\begin{subfigure}[b]{0.23\linewidth}
     \includegraphics[width=\textwidth ]{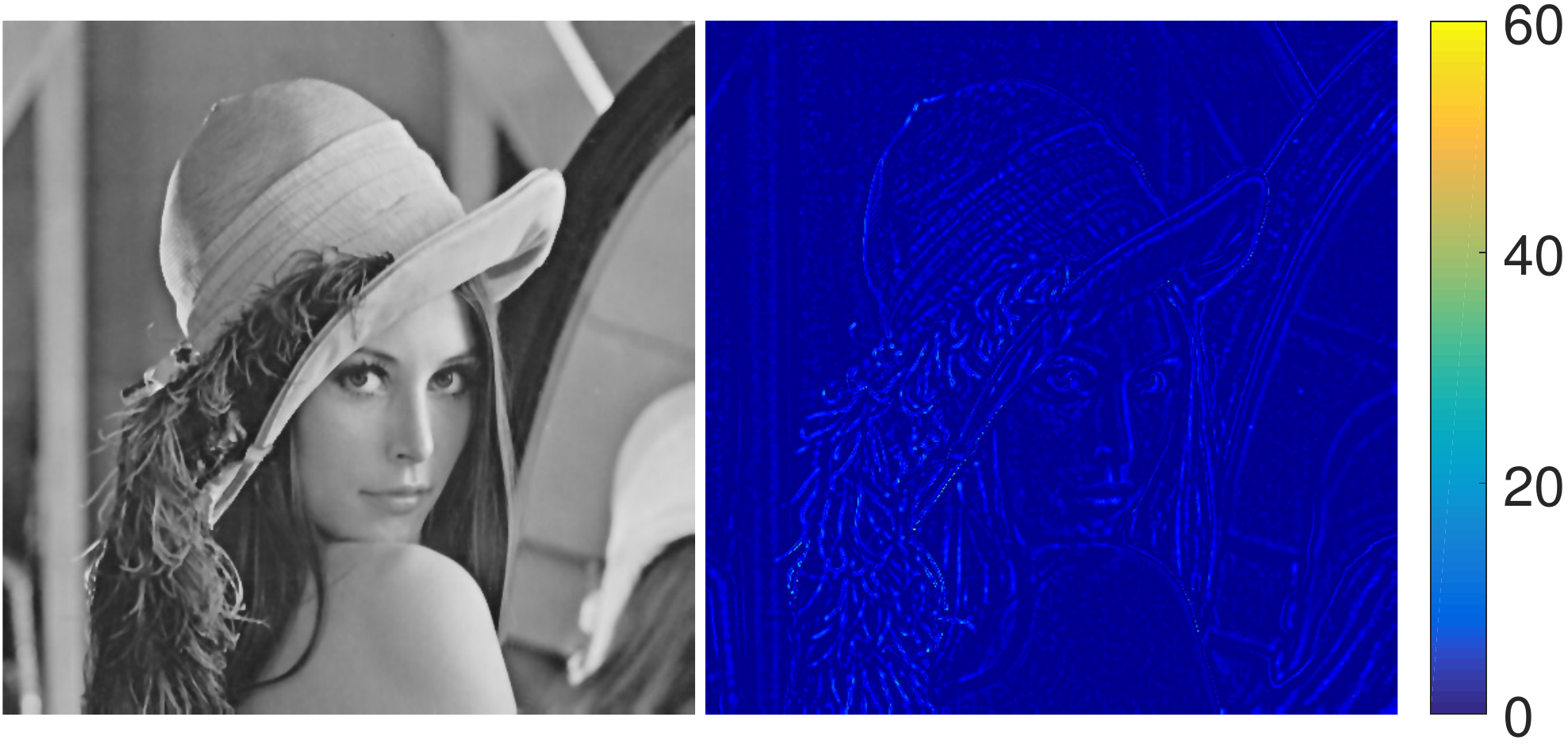}
\end{subfigure}
\begin{subfigure}[b]{0.23\linewidth}
     \includegraphics[width=\textwidth ]{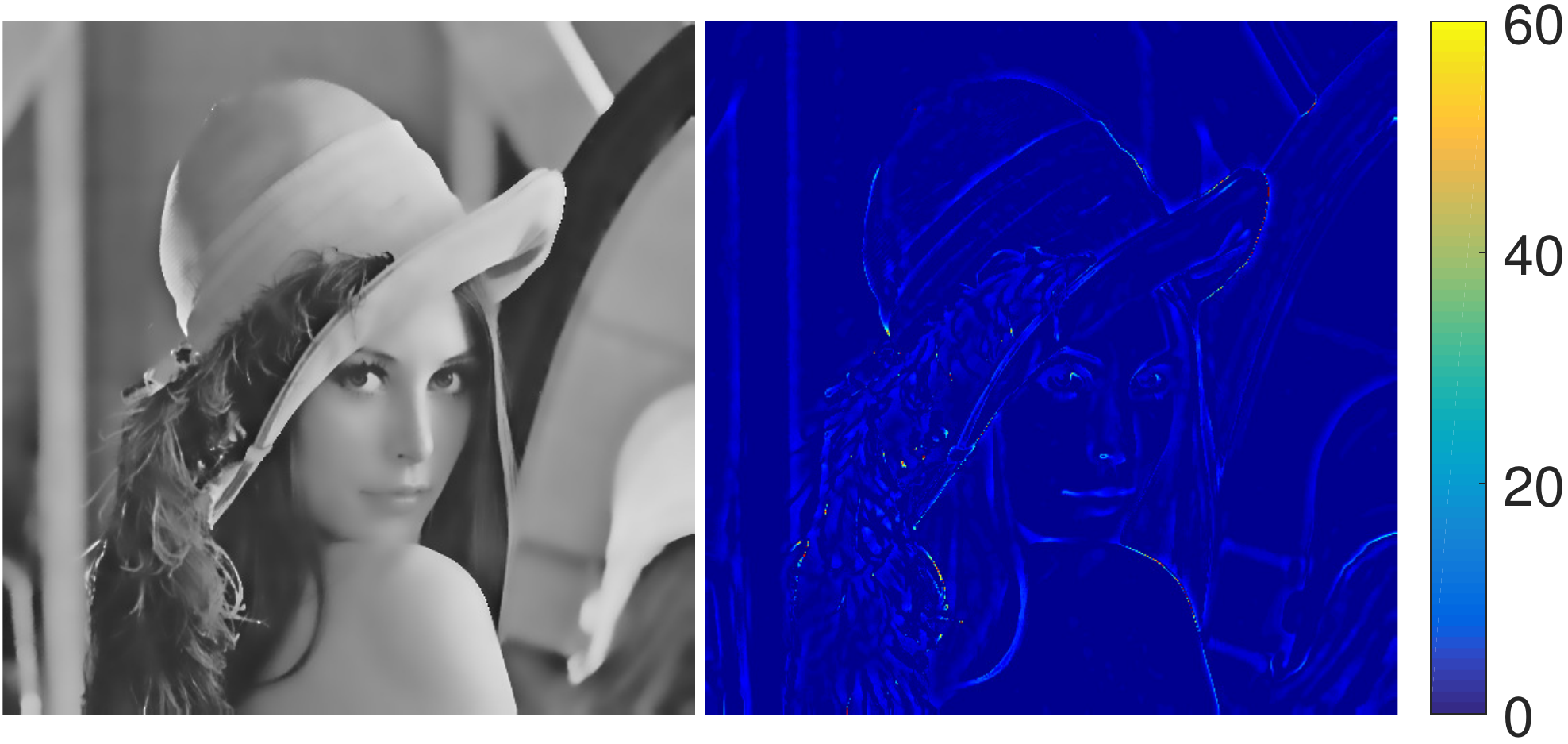}
\end{subfigure}
\begin{subfigure}[b]{0.23\linewidth}
     \includegraphics[width=\textwidth ]{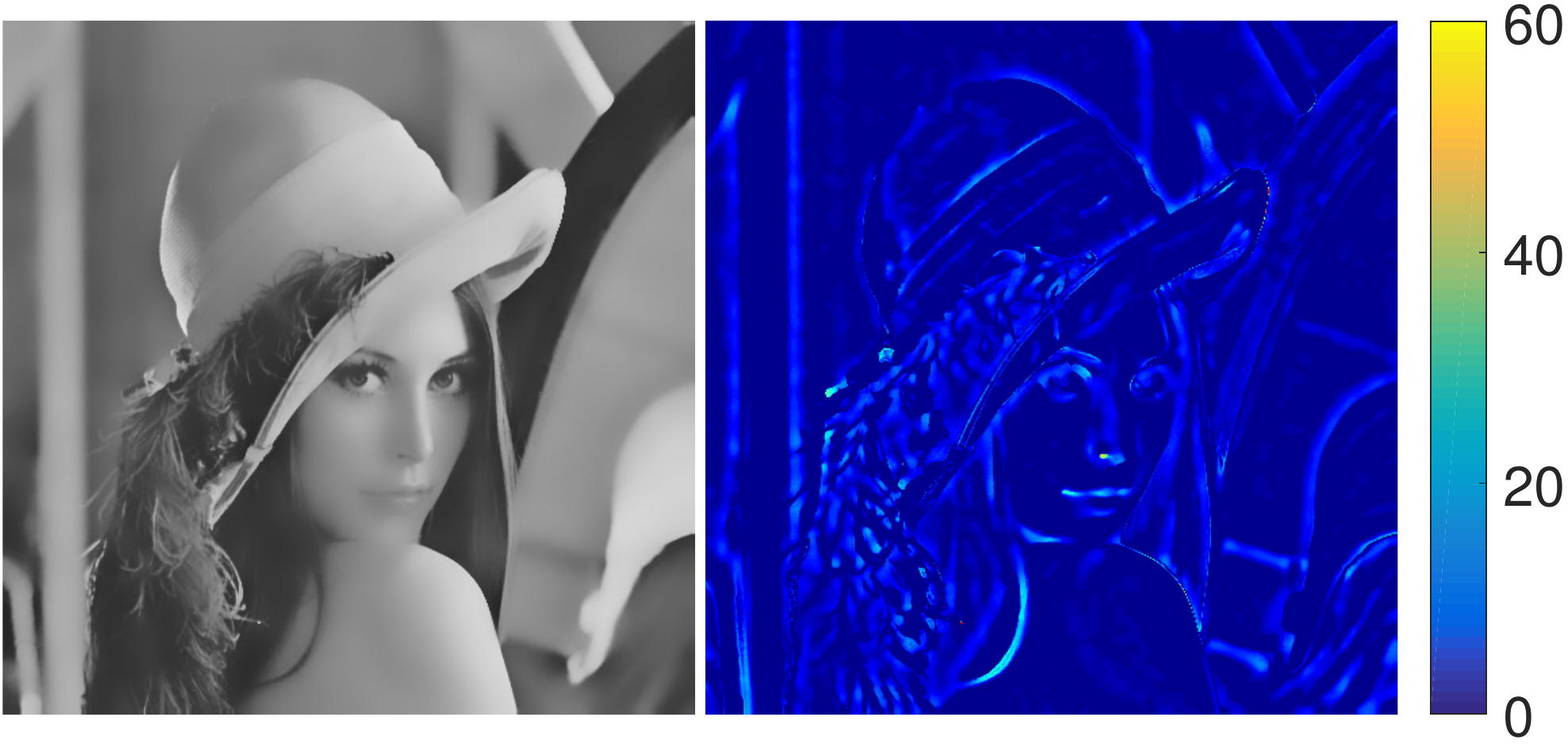}
\end{subfigure}
\begin{subfigure}[b]{0.23\linewidth}
     \includegraphics[width=\textwidth ]{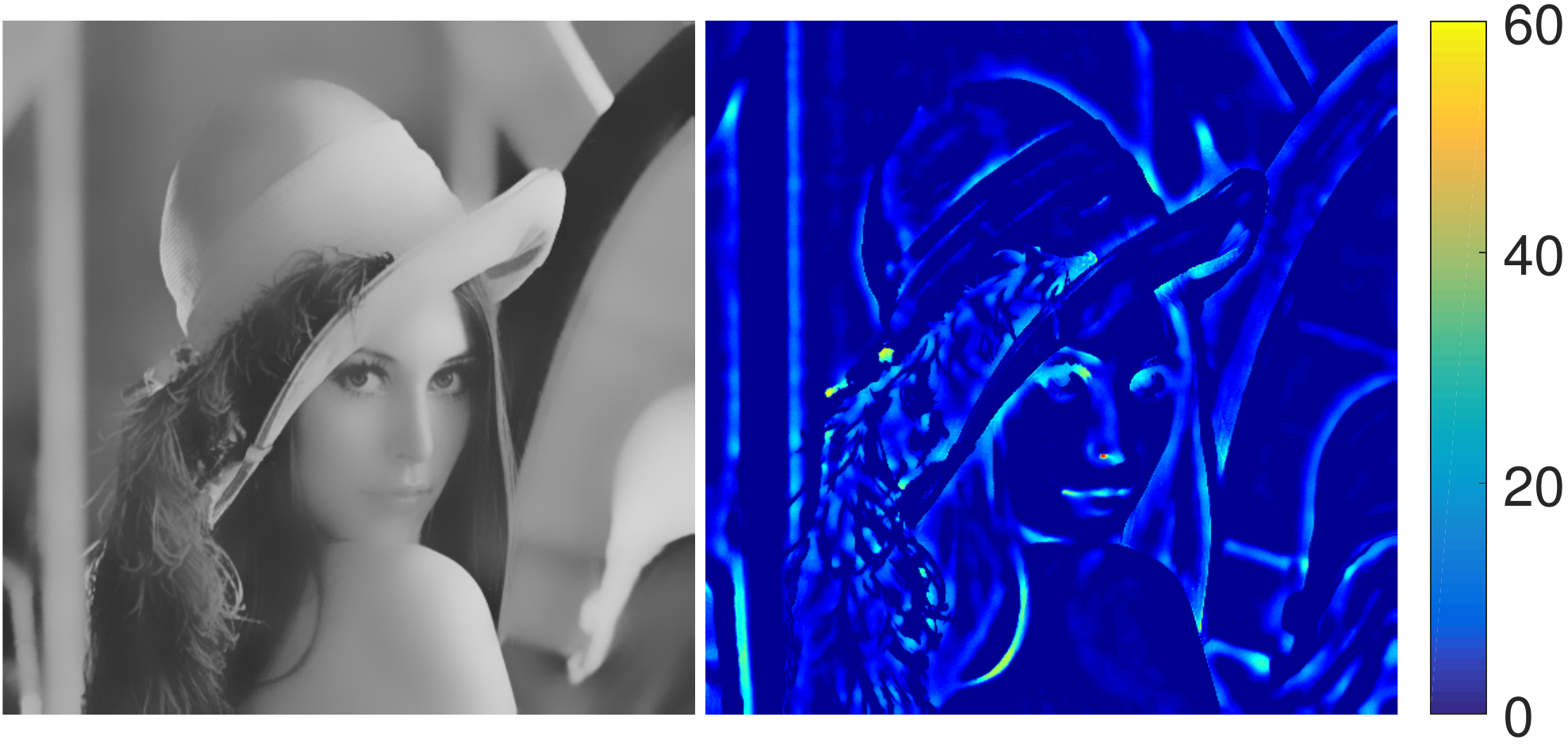}
\end{subfigure}
\rotatebox{90}{\footnotesize Chaudhury~\cite{Chaudhury_TIP_2013}}

\begin{subfigure}[b]{0.23\linewidth}
     \includegraphics[width=\textwidth ]{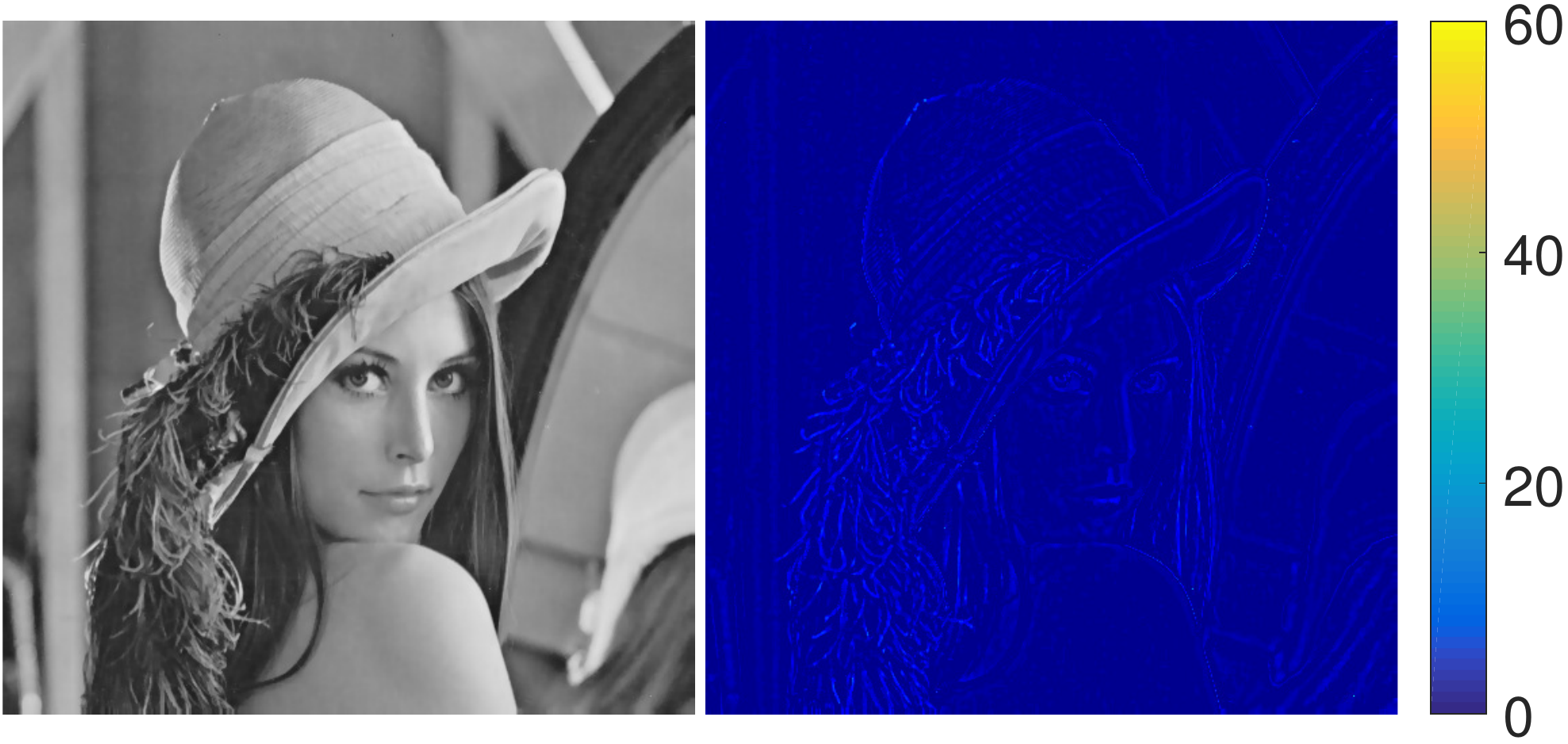}
     \caption{$\sigma_s = 1, \sigma_r = 10$}	
\end{subfigure}
\begin{subfigure}[b]{0.23\linewidth}
     \includegraphics[width=\textwidth ]{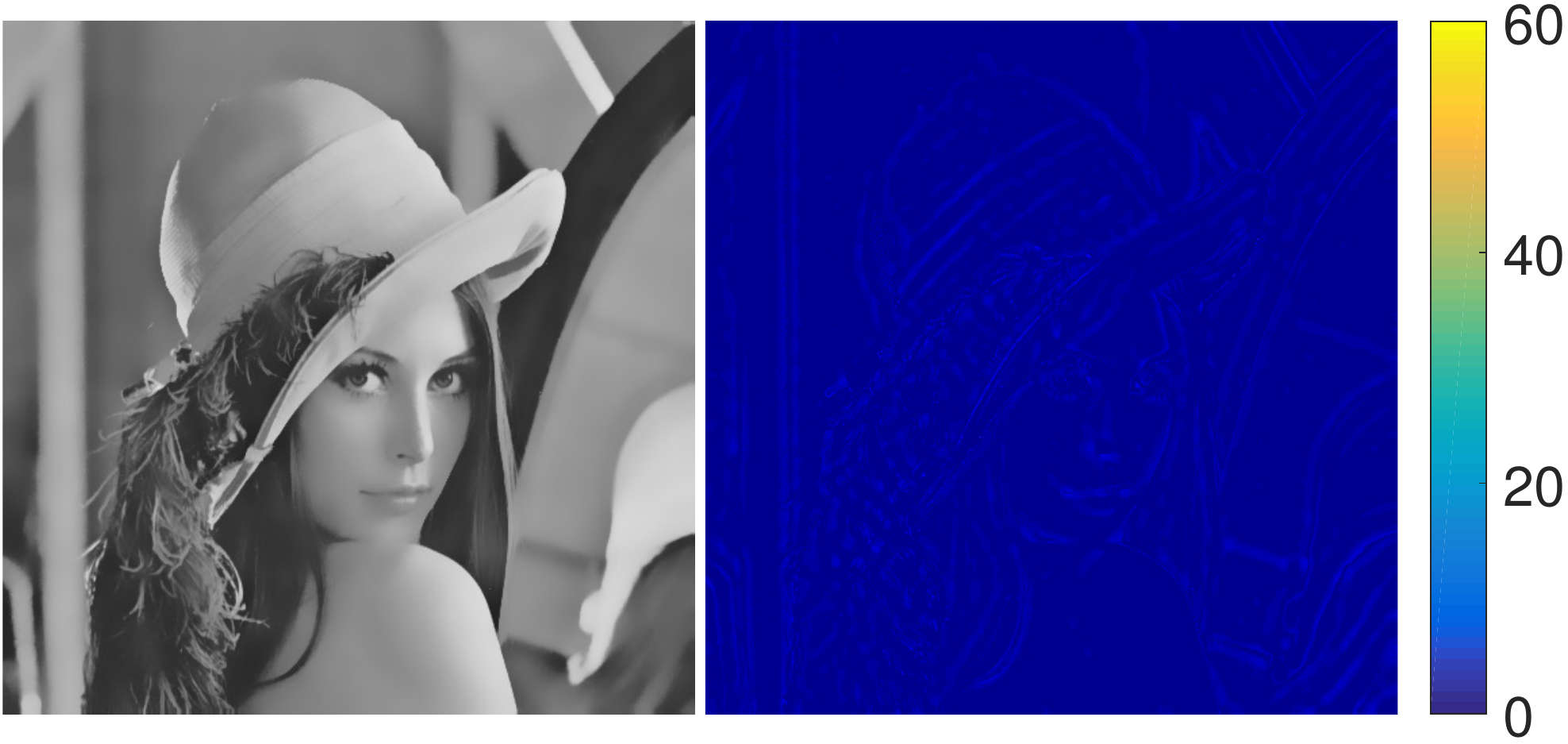}
     \caption{$\sigma_s = 4, \sigma_r = 23$}	
\end{subfigure}
\begin{subfigure}[b]{0.23\linewidth}
     \includegraphics[width=\textwidth ]{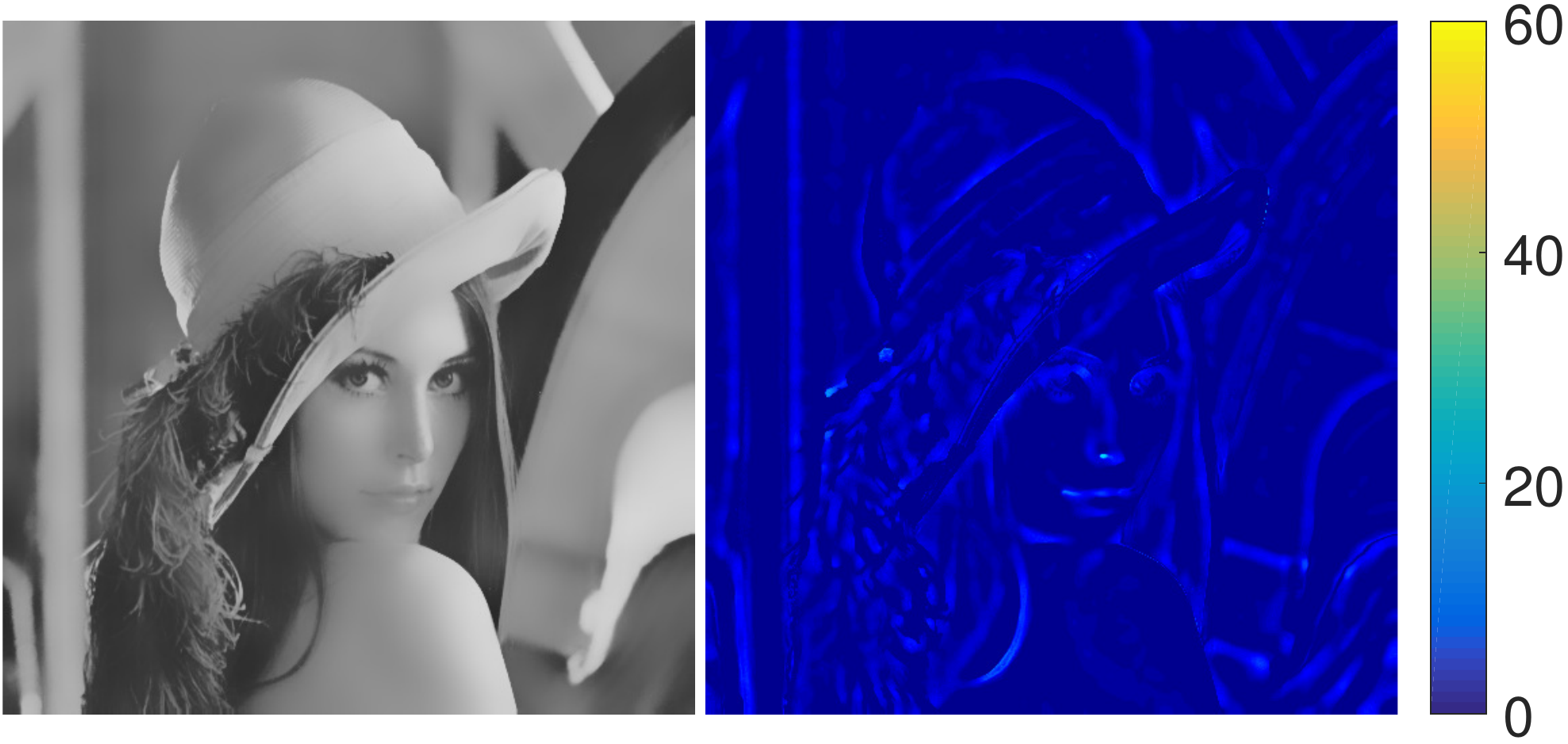}
     \caption{$\sigma_s = 7, \sigma_r = 37$}	
\end{subfigure}
\begin{subfigure}[b]{0.23\linewidth}
     \includegraphics[width=\textwidth ]{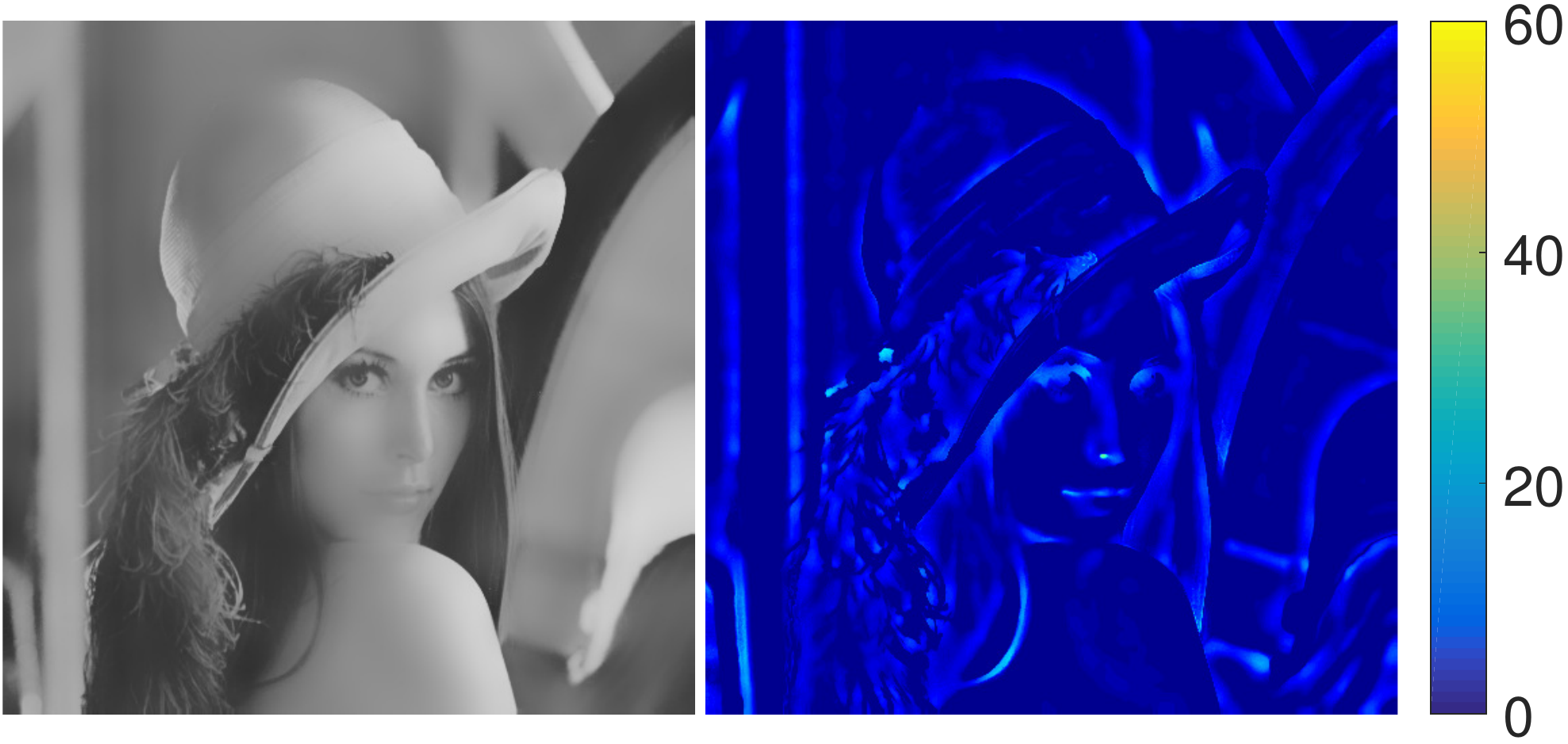}
     \caption{$\sigma_s = 10, \sigma_r = 50$}	
\end{subfigure}
\rotatebox{90}{\footnotesize \qquad \quad \quad Ours}
\caption{Lena: a demonstration of the visual quality of the approximation results. From the first row to the sixth row, the results are Porikli~\cite{Porikli_CVPR_2008}, Yang~\cite{Yang_CVPR_2009}, Gunturk~\cite{Gunturk_TIP_2011}, Pan~\cite{Pan_MPE_2014}, Chaudhury~\cite{Chaudhury_TIP_2013} and ours, each column shows the filtering results and approximation errors under different parameter settings,  where the gray images are the filtering results procured by these BF acceleration methods and the color-coded absolute error maps denote the filtering error compared with the brute-force implementation of BF. }
\label{fig:Lena}
\end{figure*}

\subsubsection{Shiftability property} Chaudhury~\etal~\cite{Chaudhury_TIP_2011} first employed the shiftability property of trigonometric functions to accelerate BF. Using Hermite polynomials, Dai~\etal~\cite{Dai_EL_2014} gave another application instance of the shiftability property. Unfortunately, the two methods can only deal with the Gaussian kernel. Even worse, they require a large number of terms to closely approximate a narrow Gaussian on the interval $[-255, 255]$ as illustrated in Fig~\ref{fig:shiftable_approxiamtion}. In~\cite{Chaudhury_TIP_2013}, Chaudhury adopted two measures to solve these problems: 1, shrinking the approximation interval; 2, dropping off the terms with small coefficients in the approximation series. But, his method inevitably increases the running time because shrinking operation happens at filtering. Dropping off small terms is because some $c_i$ in \eqref{eq:shiftable} are extremely small. Hence it is safe to get rid of them to decrease the computational cost without introducing significant errors. The advantage does not come at no cost as the nonnegative assumption of the approximation is broken as Fig~\ref{fig:shiftable_approxiamtion:Chaudhury2} shown.  

We inherit the idea of Chaudhury~\cite{Chaudhury_TIP_2013}, but take a different implementation. To decrease the size of approximation interval, we find that the larger values of $K_r(x)$ almost concentrate on a small interval $[-T_r, T_r]$. We introduce the truncated trigonometric function to approximate $K_r(x)$ on the precomputed interval $[-T_r, T_r]$. As for other values, we can simply set them as zeros. To verify the performance of our modifications, we plot Fig~\ref{fig:shiftable_approxiamtion:SSIM} to show the approximation error measured by the SSIM index with respect to the number $N$ of approximation terms and the parameter $\sigma_r$  of the Gaussian range kernel.
From the figure, we can observe that the approximation error of our approach is almost the smallest on a wide range. Compared with other shiftability property based methods~\cite{Dai_EL_2014}\cite{Chaudhury_TIP_2011}\cite{Chaudhury_TIP_2013}, the  number $N$ of approximation terms is nearly irrelevant to $\sigma_r$. In contrast, the number of the approximation terms of~\cite{Dai_EL_2014}\cite{Chaudhury_TIP_2011}\cite{Chaudhury_TIP_2013} are inversely proportional to $\sigma_r$. This is because they need many terms to approximate the long tail of the narrow Gaussian function as shown in Figs.~\ref{fig:shiftable_approxiamtion:Dai}-\ref{fig:shiftable_approxiamtion:Chaudhury2}. (The first row shows the approximation results with 5 terms. In the second row, Dai~\cite{Dai_EL_2014},  Chaudhury~\cite{Chaudhury_TIP_2011}, Chaudhury~\cite{Chaudhury_TIP_2013} and our method take 1700, 81, 42 and 5 terms respectively to obtain approximation curves with  the value 0.99 of the SSIM index.) So, the run times of these methods depend on $\sigma_r$. It is not a good behavior for acceleration methods. To reduce the number of approximation terms, we employ the best $N$-term approximation on truncated trigonometric functions to find the best approximation for arbitrary range kernels. Moreover, the best $N$-term approximation strategy also helps us solve the shortcoming of previous methods which can only approximate the Gaussian function. It is also worth noting that the dropping off small coefficients strategy adopted by Chaudhury~\cite{Chaudhury_TIP_2013} to reduce the approximation terms can be interpreted as a special case of best $N$-term approximation. Other than these, our acceleration strategy also has two advantages: First, our approximation is nonnegative as illustrated in Fig~\ref{fig:shiftable_approxiamtion:Ours}. In contrast, some values of the approximation in Figs~\ref{fig:shiftable_approxiamtion:Chaudhury1}~\ref{fig:shiftable_approxiamtion:Chaudhury2} are negative; Second, different from Figs~\ref{fig:shiftable_approxiamtion:Dai}~\ref{fig:shiftable_approxiamtion:Chaudhury1}, our approximation does not suffer from blowing up small coefficients. 




\subsection{BF comparison}

\begin{figure*}[t]
    \centering
    \begin{tabular}{ccc|ccc}
    \multicolumn{3}{c}{
    \begin{adjustbox}{valign=m}
        \begin{subfigure}[b]{0.47\linewidth}
    		\includegraphics[width=\textwidth ]{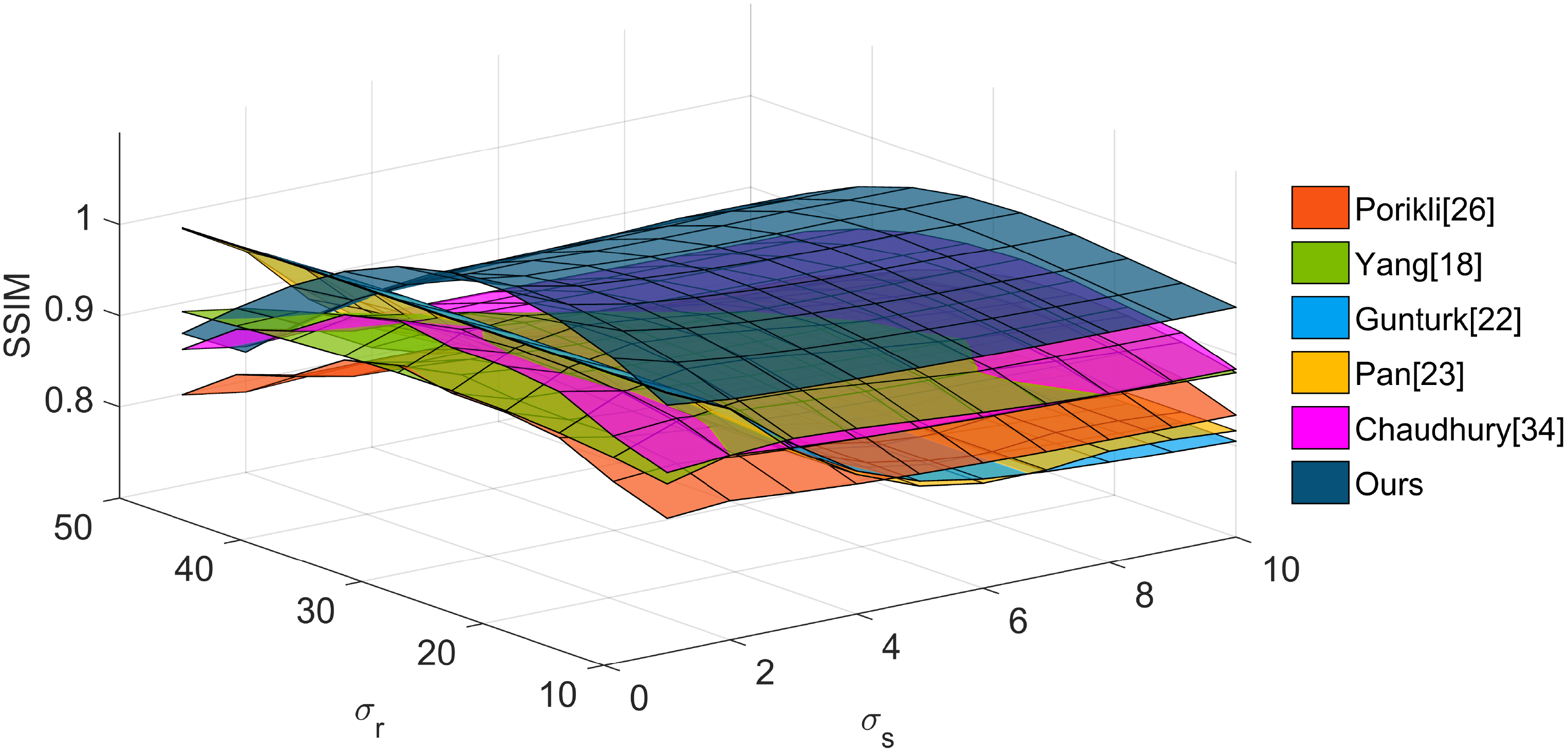}
    		\caption{SSIM}	
    	\end{subfigure}
    \end{adjustbox}
} & \multicolumn{3}{|c}{
    \begin{adjustbox}{valign=m}
        \begin{subfigure}[b]{0.47\linewidth}
    		\includegraphics[width=\textwidth ]{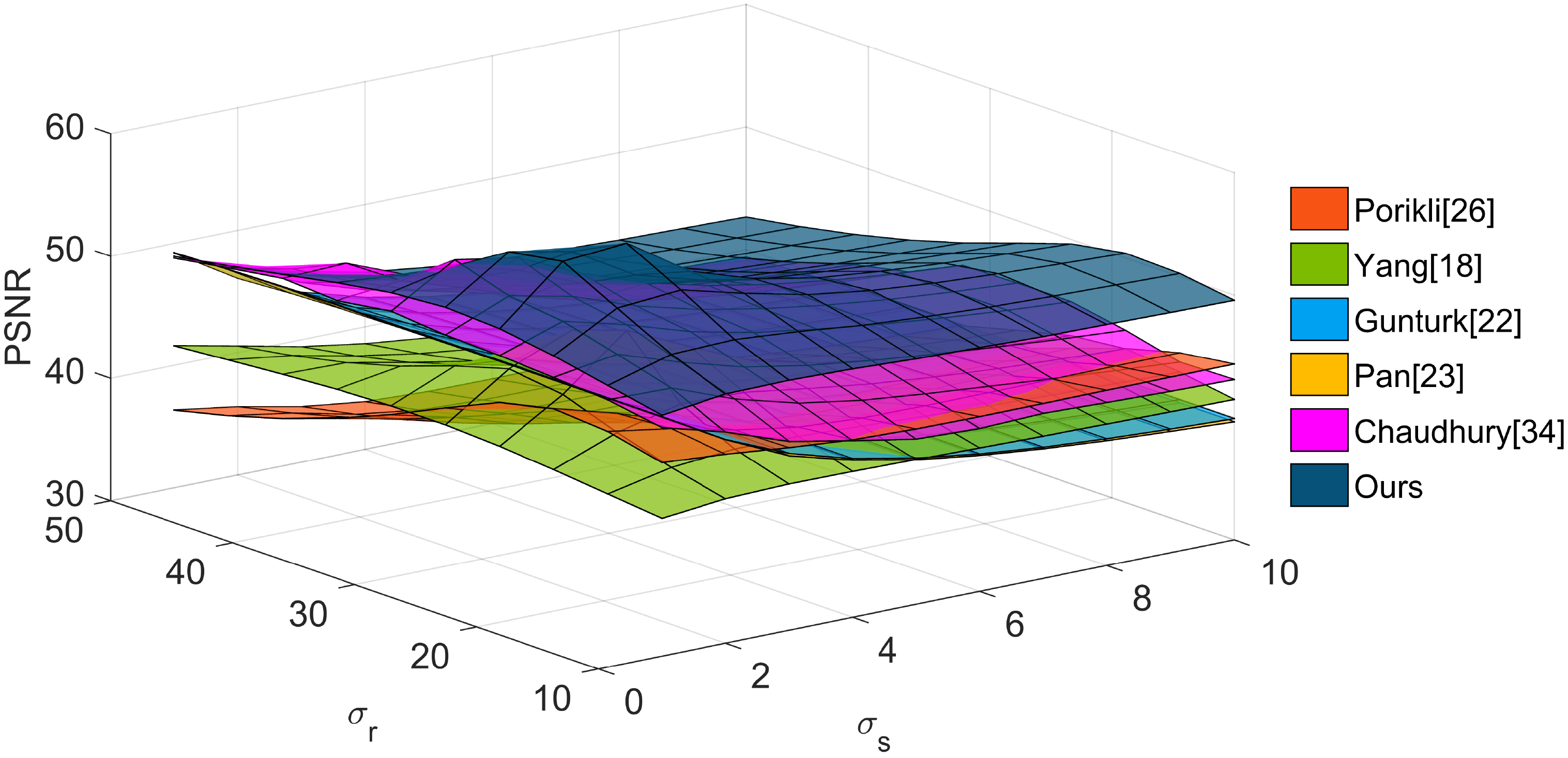}
    		\caption{PSNR}	
    	\end{subfigure}
    \end{adjustbox}
} \\
     \begin{adjustbox}{valign=m}
         \begin{subfigure}[b]{0.16\linewidth}
    		\includegraphics[width=\textwidth ]{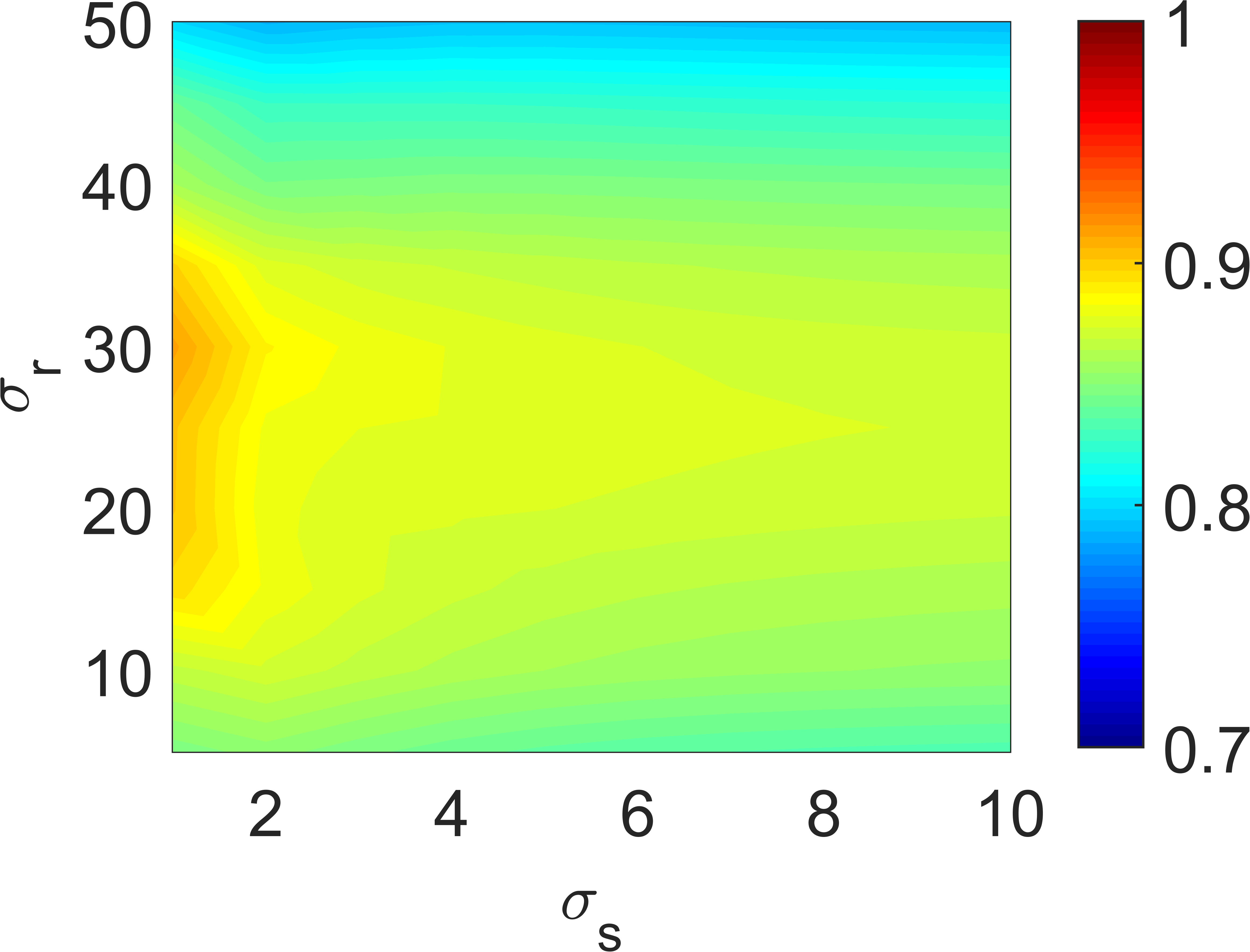}
            \captionsetup{skip=1pt}
    		\caption{ Porikli~\cite{Porikli_CVPR_2008}}	
    	\end{subfigure}
     \end{adjustbox}    &
     \begin{adjustbox}{valign=m}
         \begin{subfigure}[b]{0.16\linewidth}
    		\includegraphics[width=\textwidth ]{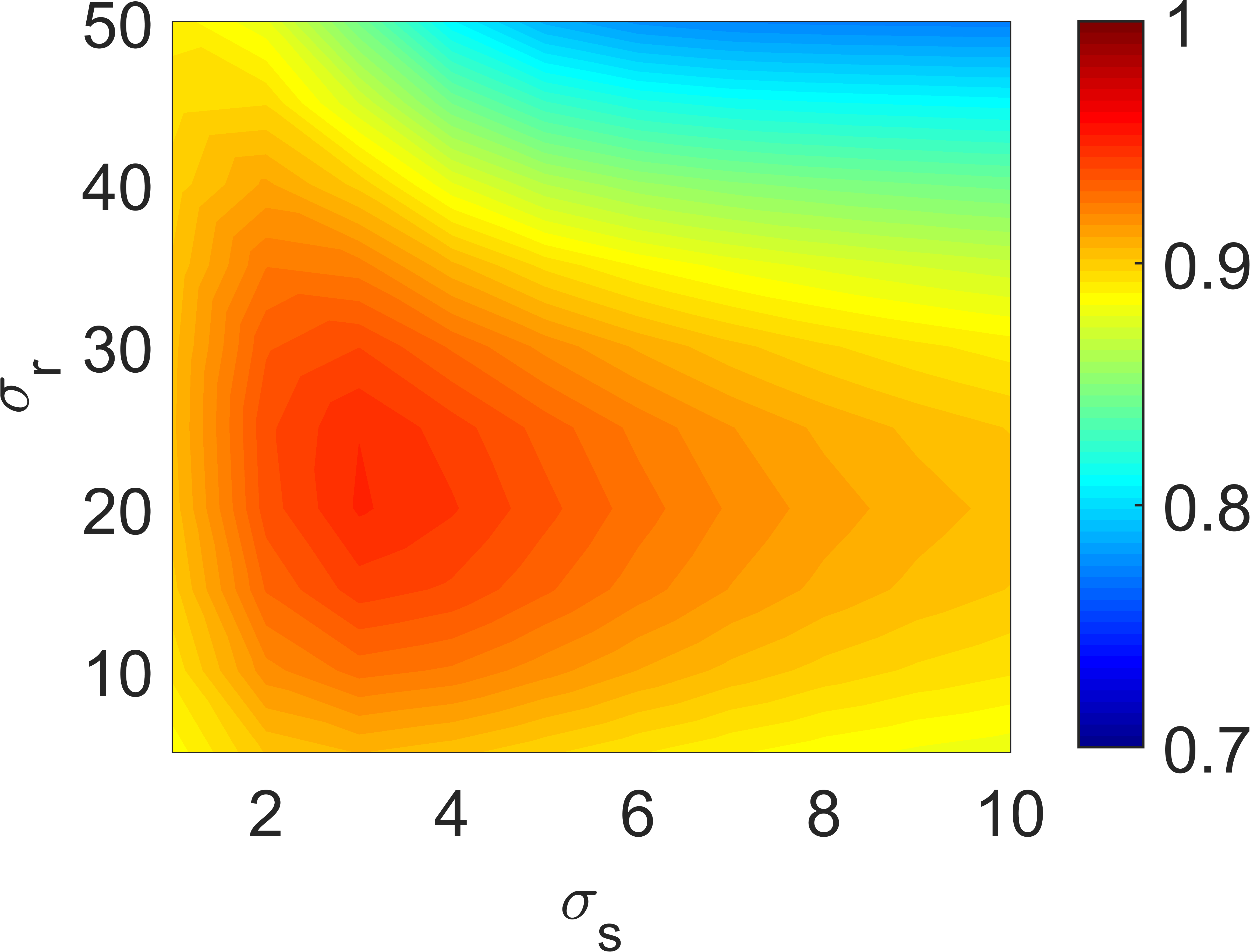}
            \captionsetup{skip=1pt}
    		\caption{Yang~\cite{Yang_CVPR_2009}}	
    	\end{subfigure}
     \end{adjustbox}    &
     \begin{adjustbox}{valign=m}
         \begin{subfigure}[b]{0.16\linewidth}
    		\includegraphics[width=\textwidth ]{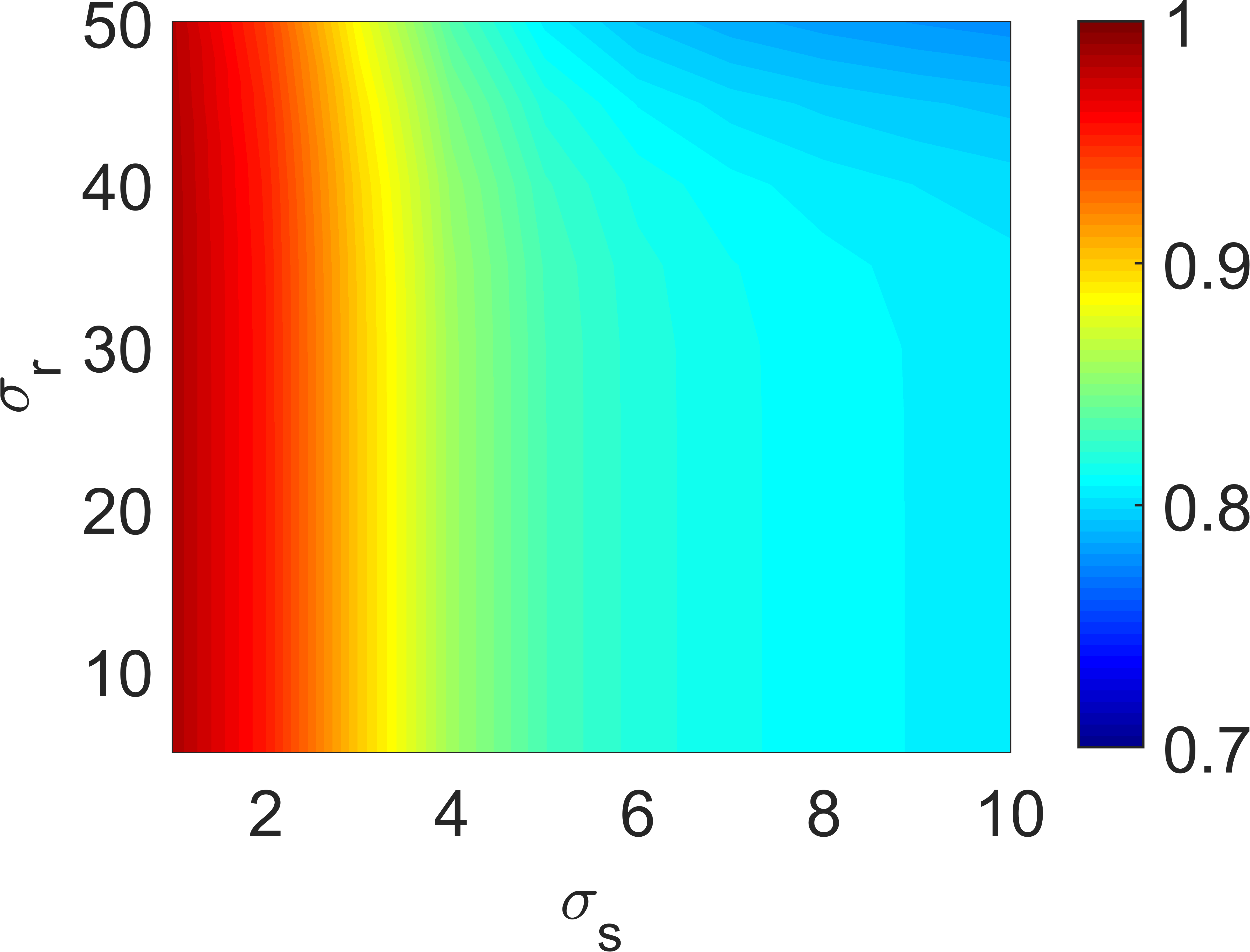}
            \captionsetup{skip=1pt}
    		\caption{Gunturk~\cite{Gunturk_TIP_2011}}	
    	 \end{subfigure}
     \end{adjustbox}    &
     \begin{adjustbox}{valign=m}
     \begin{subfigure}[b]{0.16\linewidth}
		\includegraphics[width=\textwidth ]{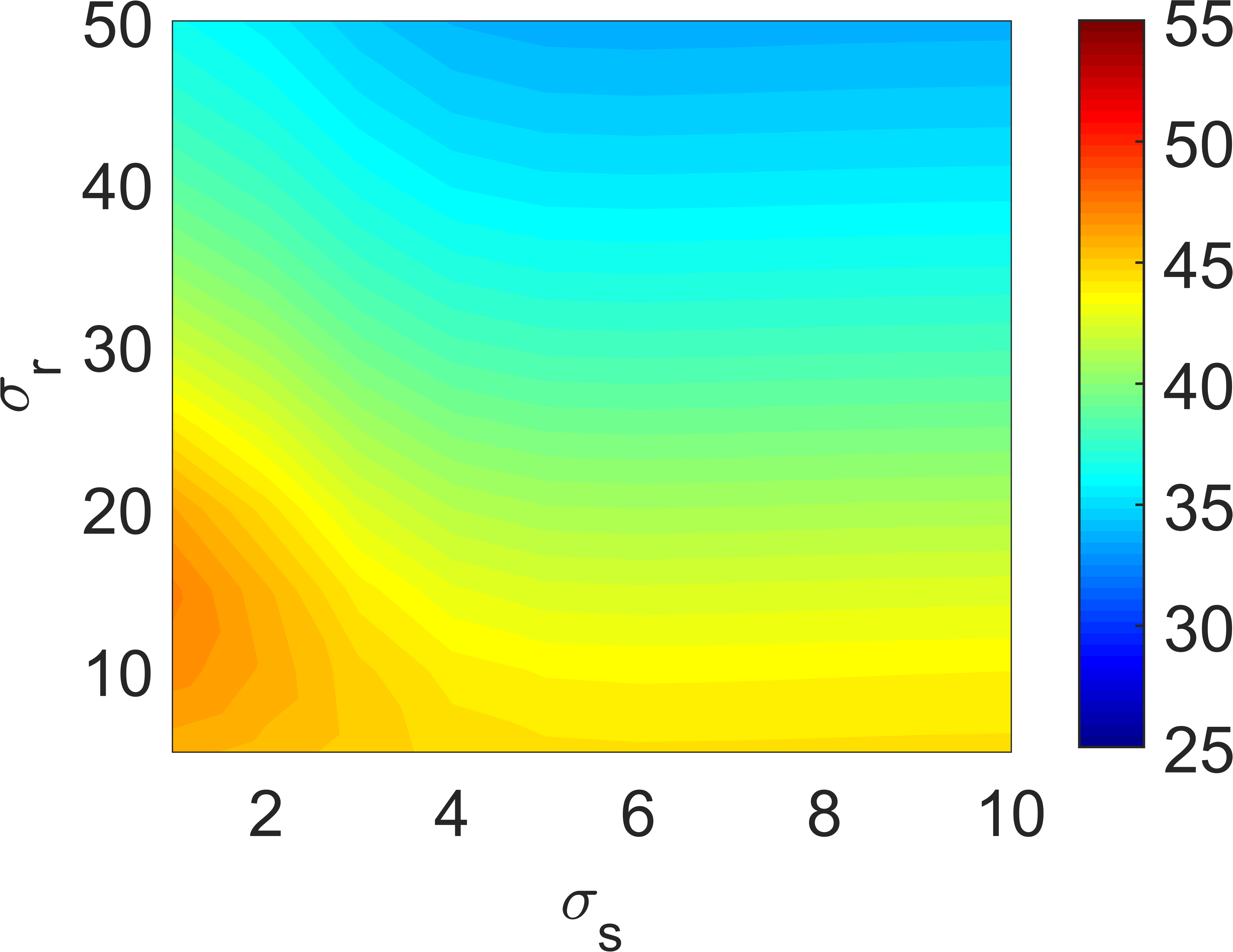}
        \captionsetup{skip=1pt}
		\caption{ Porikli~\cite{Porikli_CVPR_2008}}	
	\end{subfigure}
     \end{adjustbox}   &
     \begin{adjustbox}{valign=m}
        \begin{subfigure}[b]{0.16\linewidth}
		\includegraphics[width=\textwidth ]{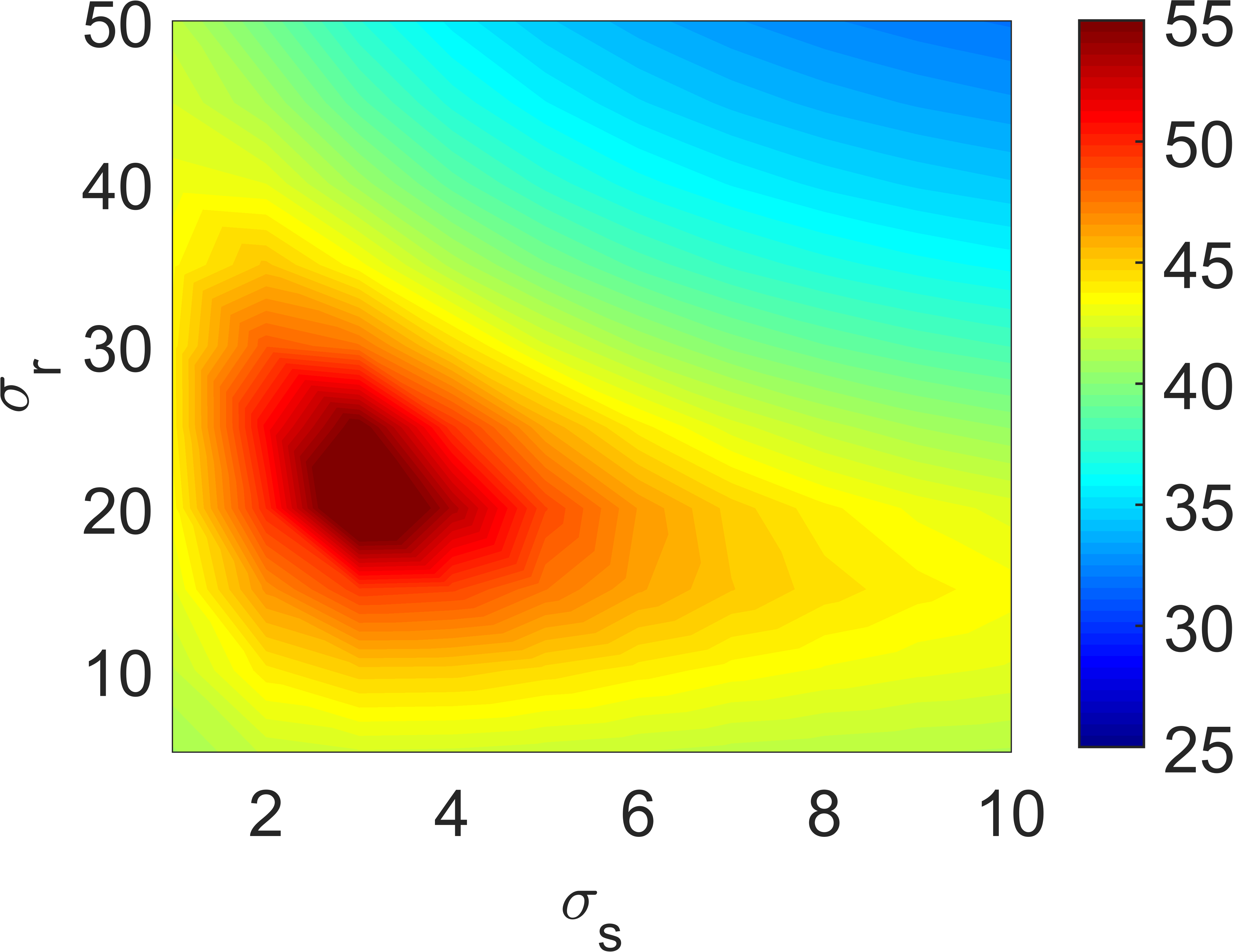}
        \captionsetup{skip=1pt}
		\caption{Yang~\cite{Yang_CVPR_2009}}	
	\end{subfigure}
     \end{adjustbox}    &
     \begin{adjustbox}{valign=m}
         \begin{subfigure}[b]{0.16\linewidth}
		\includegraphics[width=\textwidth ]{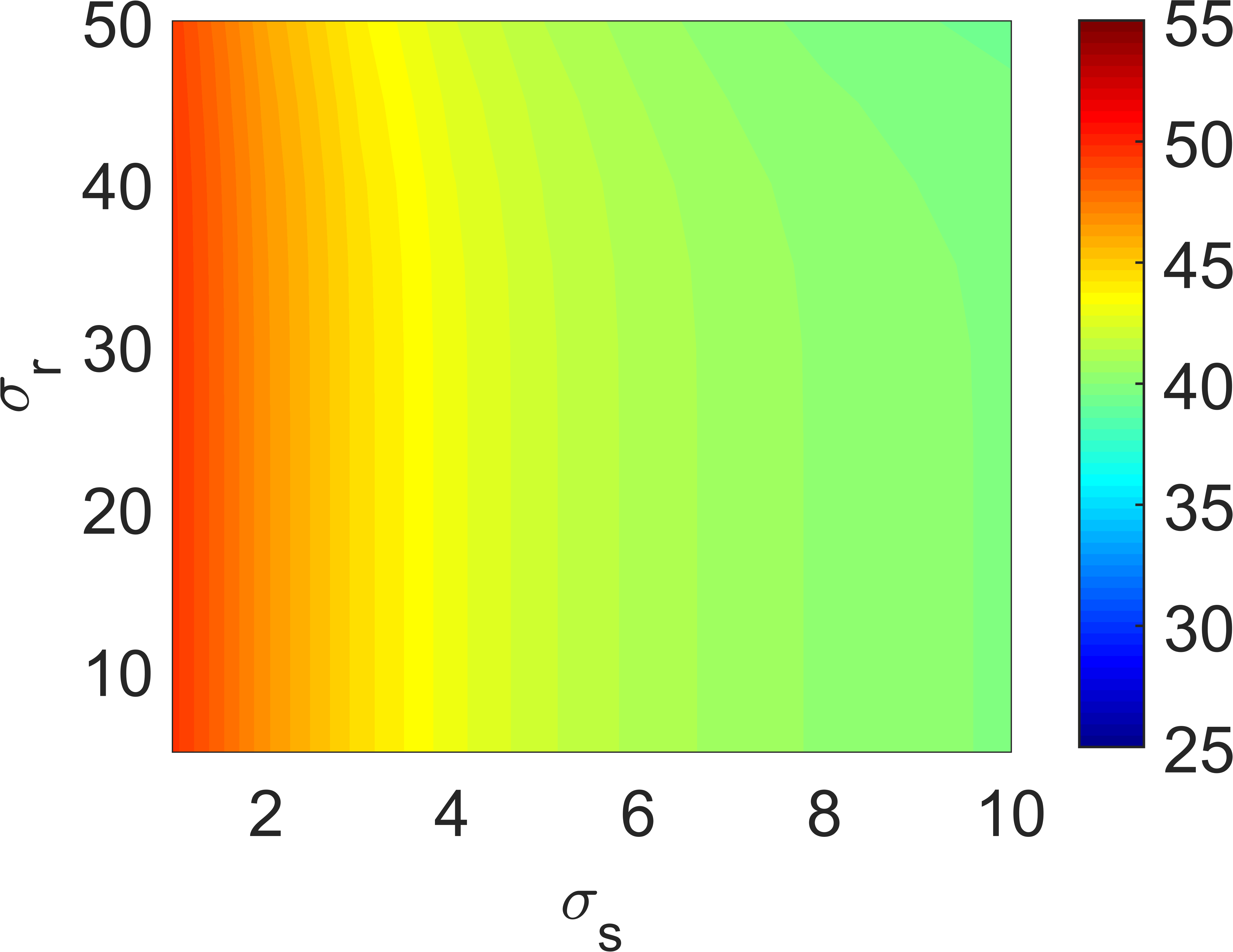}
        \captionsetup{skip=1pt}
		\caption{Gunturk~\cite{Gunturk_TIP_2011}}	
	\end{subfigure}
     \end{adjustbox}   \\
     \begin{adjustbox}{valign=m}
     \begin{subfigure}[b]{0.16\linewidth}
		\includegraphics[width=\textwidth ]{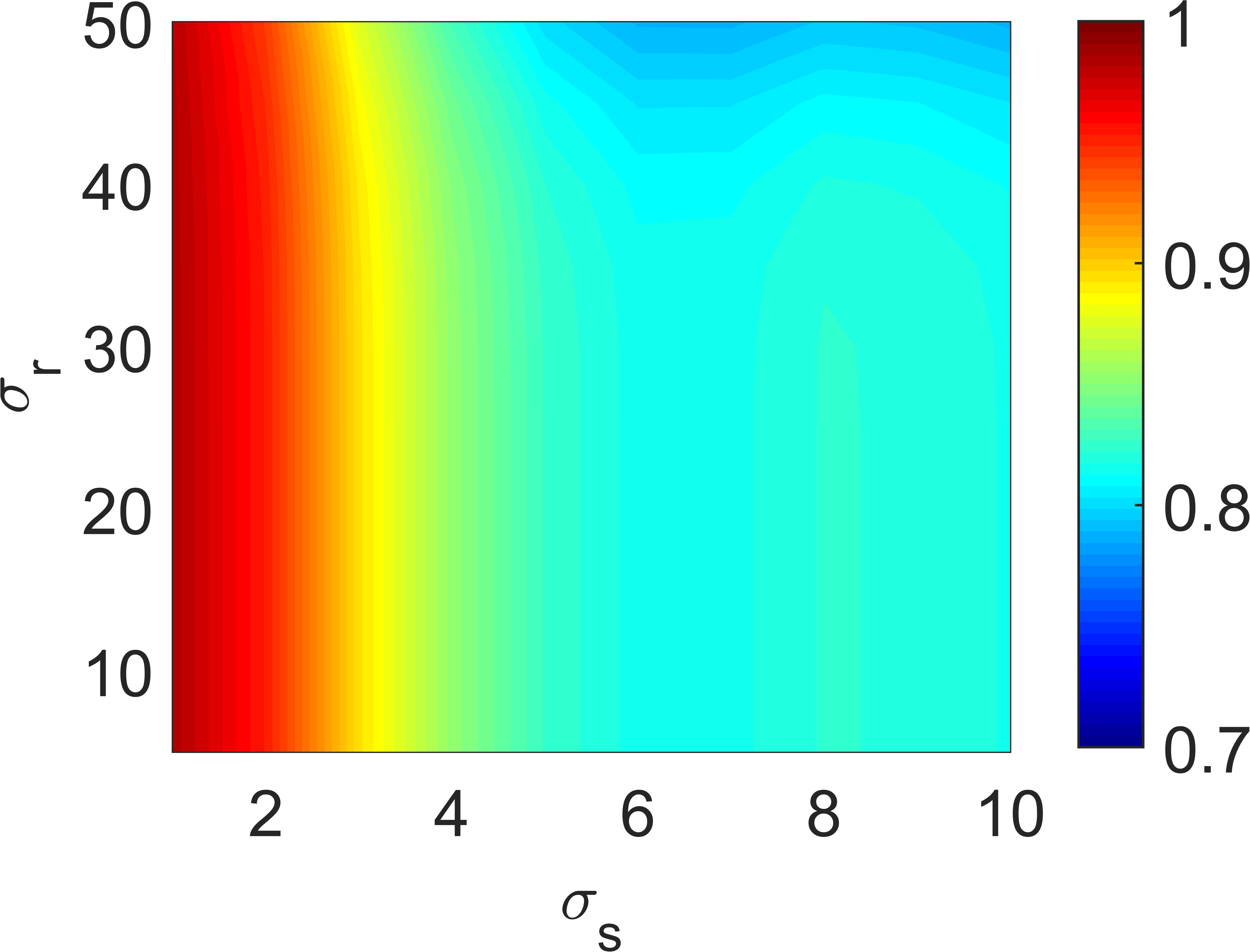}
        \captionsetup{skip=1pt}
		\caption{Pan~\cite{Pan_MPE_2014}}	
	\end{subfigure}
     \end{adjustbox}     &
     \begin{adjustbox}{valign=m}
     \begin{subfigure}[b]{0.16\linewidth}
		\includegraphics[width=\textwidth ]{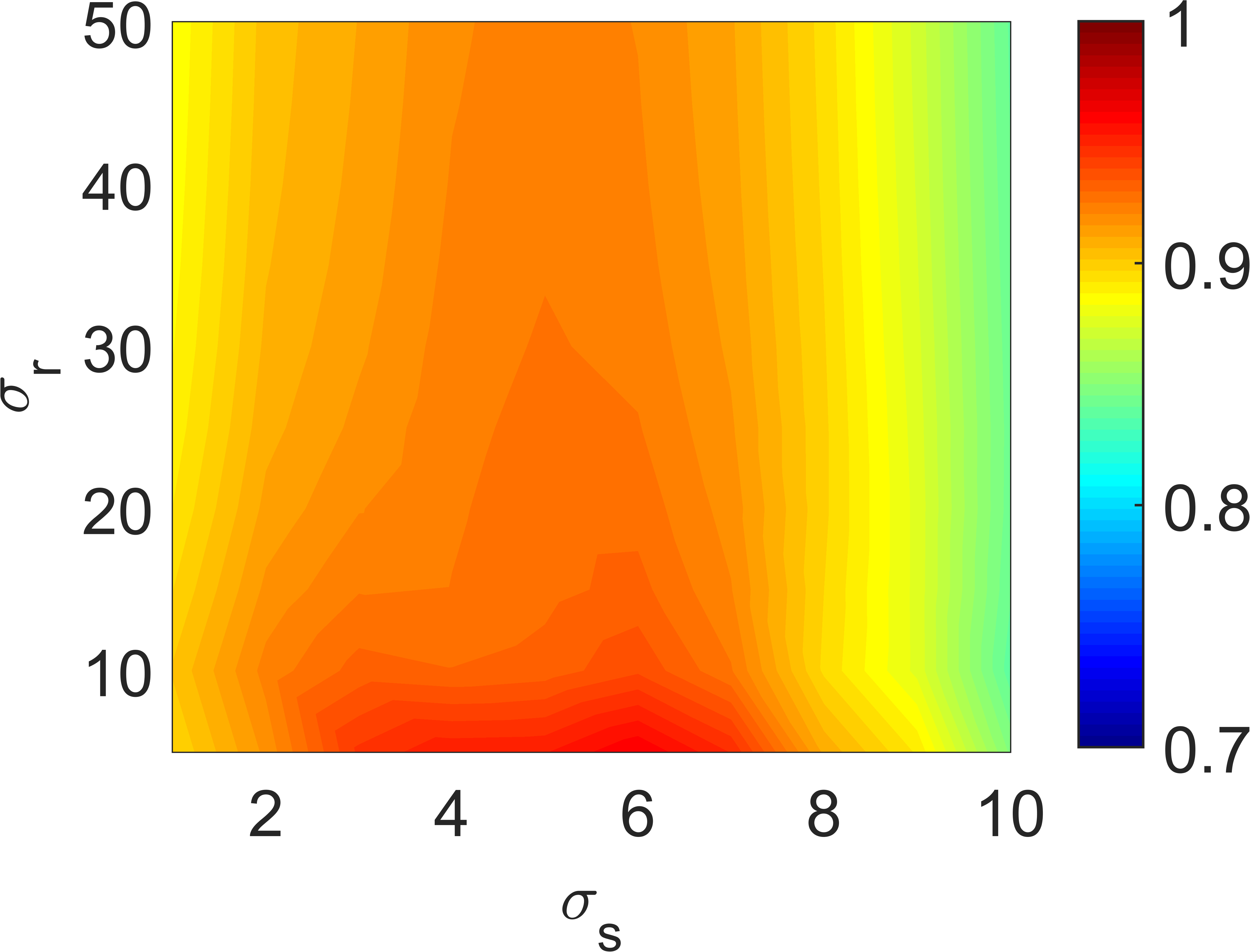}
        \captionsetup{skip=1pt}
		\caption{Chaudhury~\cite{Chaudhury_TIP_2013}}	
	\end{subfigure}
     \end{adjustbox}   &
     \begin{adjustbox}{valign=m}
     \begin{subfigure}[b]{0.16\linewidth}
		\includegraphics[width=\textwidth ]{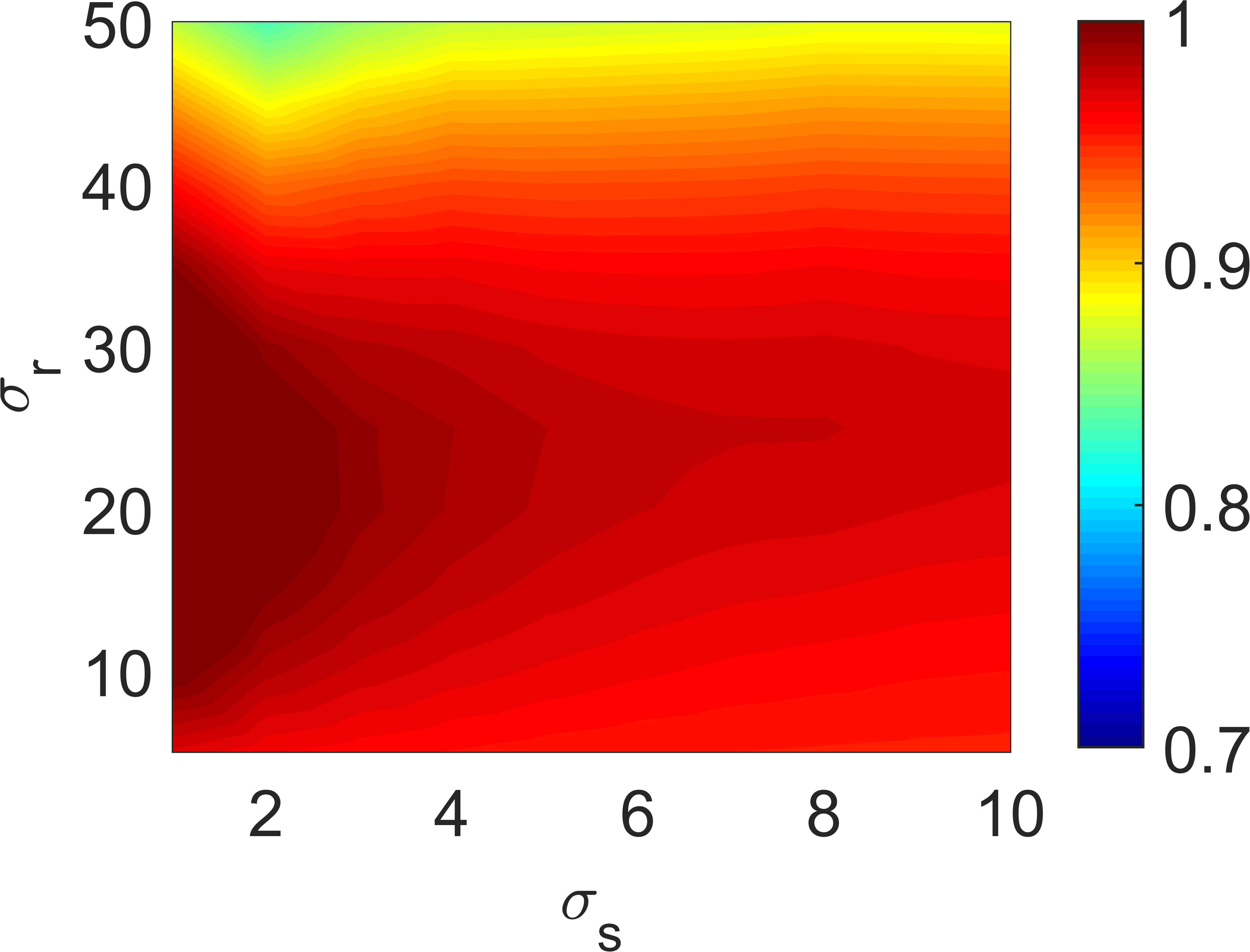}
        \captionsetup{skip=1pt}
		\caption{Ours}	
	\end{subfigure}
     \end{adjustbox}    &
     \begin{adjustbox}{valign=m}
     \begin{subfigure}[b]{0.16\linewidth}
		\includegraphics[width=\textwidth ]{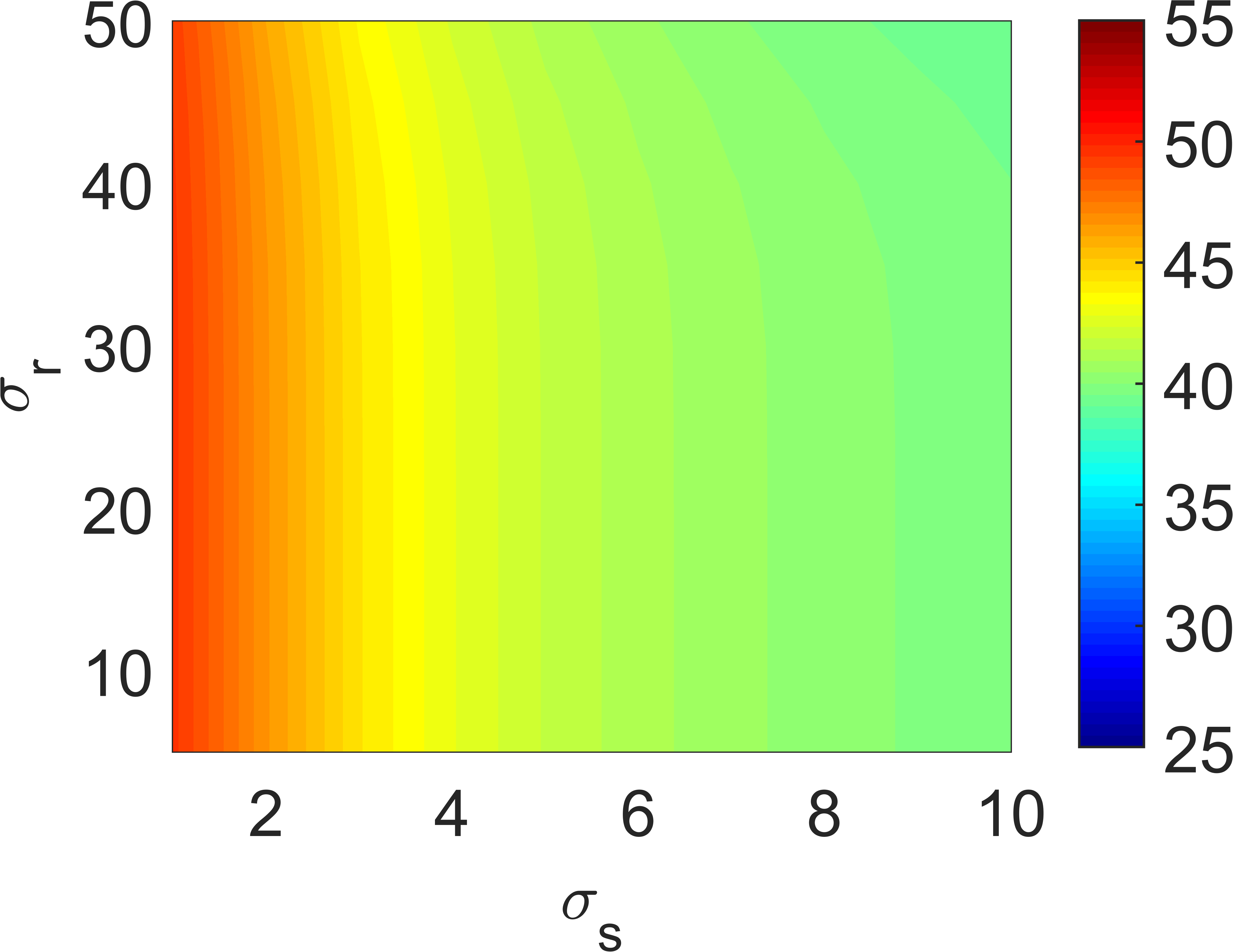}
        \captionsetup{skip=1pt}
		\caption{Pan~\cite{Pan_MPE_2014}}	
	\end{subfigure}
     \end{adjustbox}     &
     \begin{adjustbox}{valign=m}
        \begin{subfigure}[b]{0.16\linewidth}
		\includegraphics[width=\textwidth ]{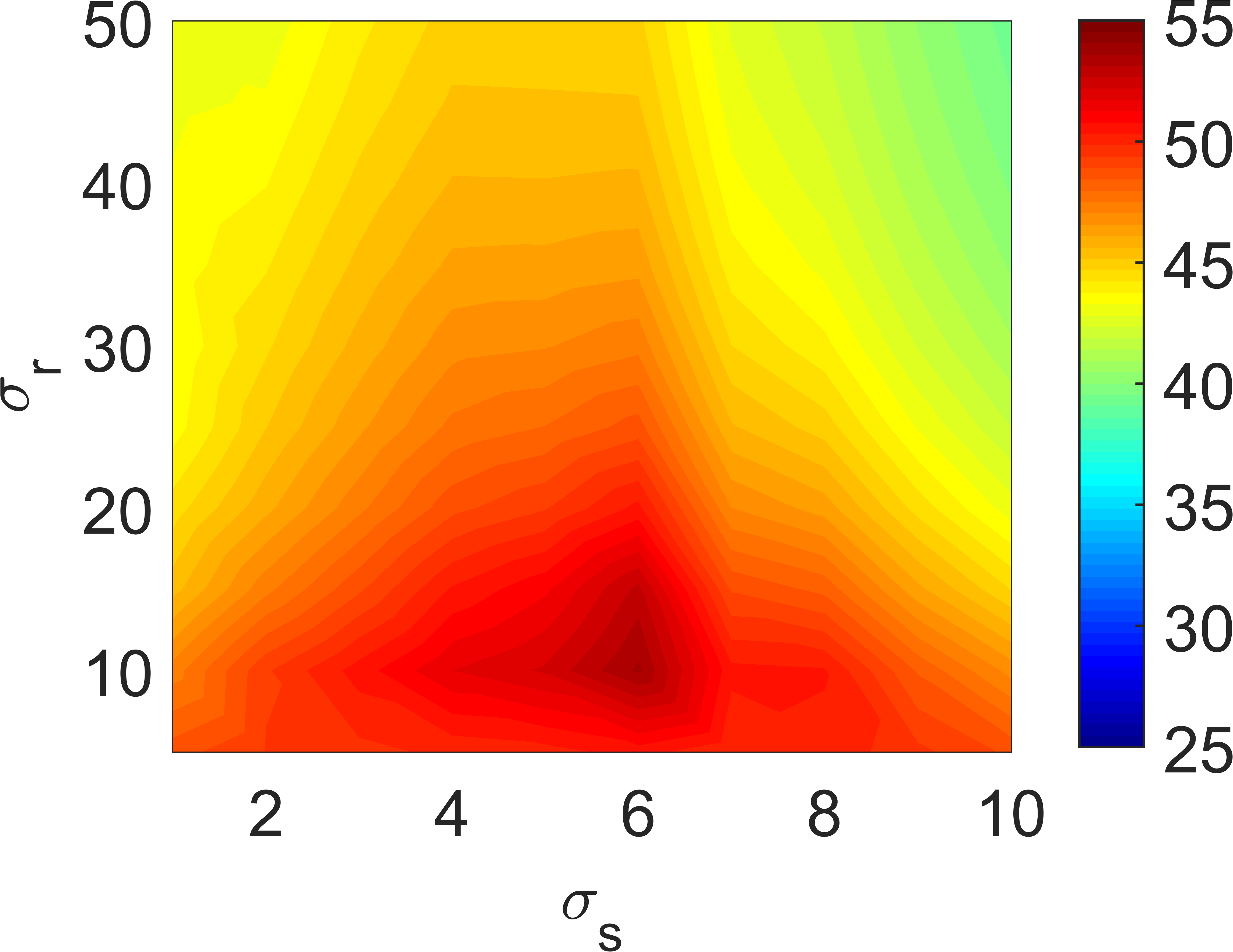}
        \captionsetup{skip=1pt}
		\caption{Chaudhury~\cite{Chaudhury_TIP_2013}}	
	\end{subfigure}
     \end{adjustbox}    &
     \begin{adjustbox}{valign=m}
       \begin{subfigure}[b]{0.16\linewidth}
		\includegraphics[width=\textwidth ]{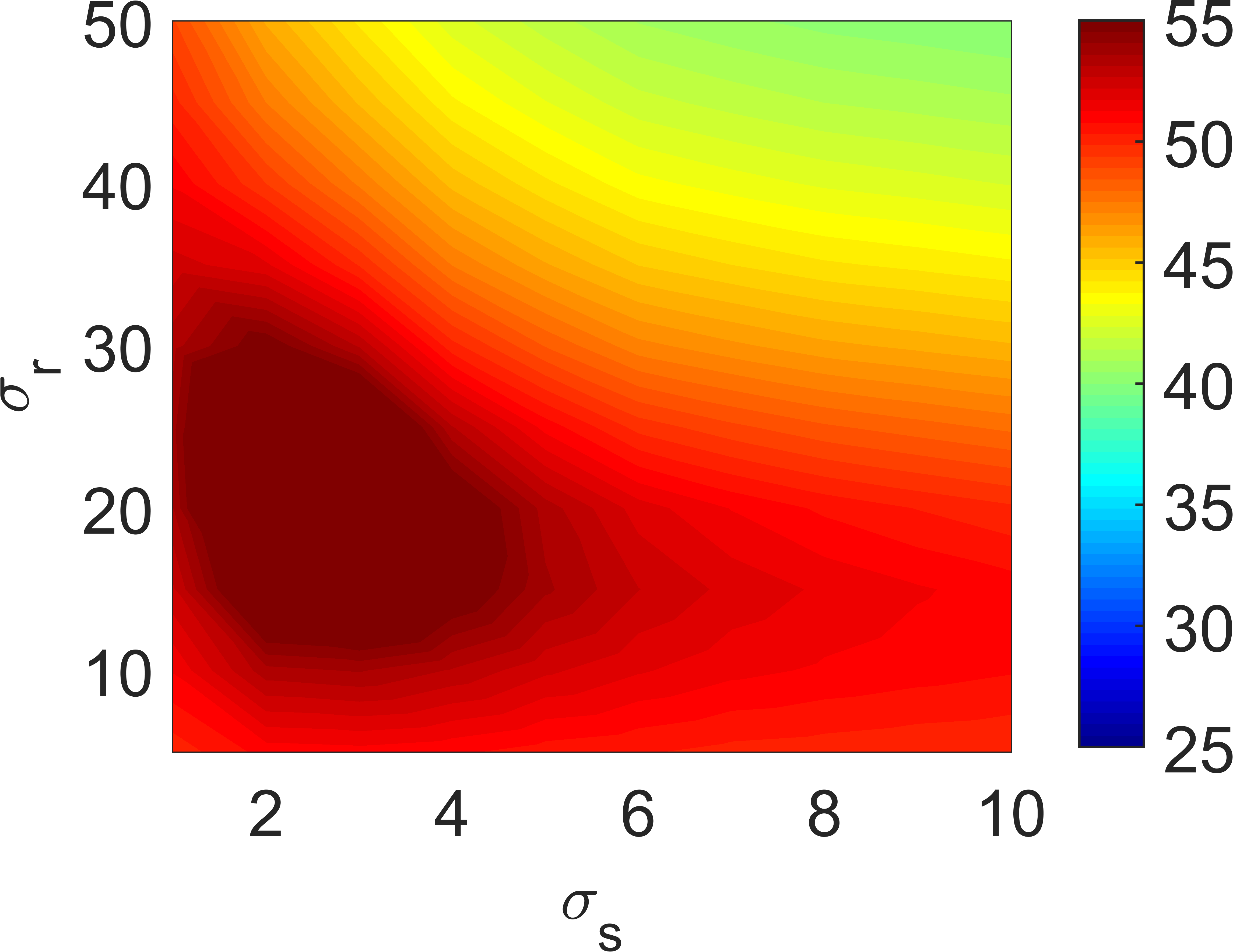}
        \captionsetup{skip=1pt}
		\caption{Ours}	
	\end{subfigure}
     \end{adjustbox}
    \end{tabular}
\caption{Qualitative evaluation for the filtering results of six methods, where SSIM and PSNR are chosen as the quantitative index to measure the approximation error. (a) (b) illustrate the SSIM and PSNR approximation error surfaces.  (c)-(n) plot the the SSIM and PSNR indices of each method, where the left part shows SSIM, and the right part shows PSNR.}	
\label{fig:BF_SSIM_sigma_sigma}
\end{figure*}

In this section, we compare our acceleration method with five state-of-the-art methods (\ie Porikli~\cite{Porikli_CVPR_2008},   Yang~\cite{Yang_CVPR_2009},  Gunturk~\cite{Gunturk_TIP_2011},  Pan~\cite{Pan_MPE_2014} and Chaudhury~\cite{Chaudhury_TIP_2013}) in terms of qualitative and quantitative aspects. The techniques adopted by the six methods are summarized in Table~\ref{tab:summary}. For a fair comparison, we implement all acceleration methods in C++ without SIMD instructions on a laptop with a 2.0 GHz CPU. Here only experiments on approximating the bilateral filter with Gaussian spatial and range kernels are provided to illustrate the validity and the effectiveness of the proposed method because the shiftability property based acceleration technique~\cite{Chaudhury_TIP_2013} can only speed up the Gaussian range kernel. Moreover, we set the number of bins equal to 32 in our evaluation for the PBFIC based method~\cite{Yang_CVPR_2009} and the box filtering based methods~\cite{Gunturk_TIP_2011,Pan_MPE_2014,Porikli_CVPR_2008}.  

\subsubsection{Accuracy}

We start our experiments from three famous images (\ie Lena, Barbara and Boat). The statistical data in the following paragraphs is an average of the three images. First, we perform the visual comparison of the fast bilateral filtering schemes. The filtering results of Lena are illustrated in Fig~\ref{fig:Lena}. The color-coded images represent the absolute error between the filtering image and ground truth. The variances of the spatial kernel and the range kernel are $\sigma_s \in \{1, 4, 7, 10\}, \sigma_r \in \{ 10, 23, 37, 50\}$.  The number of boxes used is $3$ as we find that the total run time of three box filters is nearly same to the kernel separation method used by Yang~\cite{Yang_CVPR_2009}, and the number of approximation terms for range kernel is $5$.

Absolute error images in Fig~\ref{fig:Lena} show that our method produces smaller error than the other methods. Hence, better accuracy is achieved in approximating BF.  Besides the visual comparison, we also offer a convincing and quantitative illustration for the advantages of our method over previous methods. The SSIM index and the peak signal-to-noise ratio (PSNR) are exploited to evaluate the approximation errors of different methods. The statistical data with respect to different  bandwidth parameters are plotted in Fig~\ref{fig:BF_SSIM_sigma_sigma}. We can easily verify that our method achieves the lowest approximation error for rather wide parameter variation interval.

\subsubsection{Efficiency}

\begin{table}[t]
\caption{Run time comparison. The size of tested image is chosen as $1024 \times 1024$.}
\label{tab:run_time}
\centering
\begin{tabularx}{\linewidth}{@{}|c|Y|Y|Y|Y|Y|Y|Y|@{}}
\hline
\multirow{2}{*}{} & \multicolumn{7}{c|}{$\sigma_r$, where $\sigma_s= 6$} \\ \cline{2-8}
 & 15  & 20 & 25 & 30 & 35 & 40 & 45  \\ \hline
Porikli~\cite{Porikli_CVPR_2008} & 1.69s  & 1.69s & 1.65s & 1.64s & 1.66s & 1.63s & 1.65s  \\ \hline
Yang~\cite{Yang_CVPR_2009} & 1.59s & 1.62s & 1.60s & 1.61s & 1.59s & 1.63s & 1.62s \\ \hline
Gunturk~\cite{Gunturk_TIP_2011} & 1.63s & 1.61s & 1.63s & 1.60s & 1.62s & 1.60s & 1.64s  \\ \hline
Pan~\cite{Pan_MPE_2014} & 1.61s & 1.63s & 1.60s & 1.62s & 1.61s & 1.63s & 1.60s \\ \hline
Chaudhury~\cite{Chaudhury_TIP_2013} & 1.81s & 1.67s & 1.59s & 1.52s & 1.48s  & 1.41s & 1.35s \\ \hline
Ours & $\bm{1.39}$s & $\bm{1.37}$s & $\bm{1.39}$s & $\bm{1.38}$s & $\bm{1.34}$s & $\bm{1.36}$s & $\bm{1.37}$s \\ \hline \hline \hline
\multirow{2}{*}{} & \multicolumn{7}{c|}{$\sigma_s$, where $\sigma_r=30$} \\ \cline{2-8}
 & 3  & 4 & 5 & 6 & 7 & 8 & 9  \\ \hline
Porikli~\cite{Porikli_CVPR_2008} & 1.63s  & 1.64s & 1.63s & 1.64s & 1.65s & 1.64s & 1.65s \\ \hline
Yang~\cite{Yang_CVPR_2009} & 1.61s & 1.63s & 1.60s & 1.61s & 1.62s & 1.63s & 1.62s\\ \hline
Gunturk~\cite{Gunturk_TIP_2011} & 1.60s & 1.59s & 1.61s & 1.60s & 1.61s & 1.63s &  1.62s \\ \hline
Pan~\cite{Pan_MPE_2014} & 1.61s & 1.61s & 1.60s & 1.62s  & 1.63s & 1.62s & 1.64s \\ \hline
Chaudhury~\cite{Chaudhury_TIP_2013} & 1.51s & 1.50s & 1.52s & 1.52s  & 1.51s & 1.52s & 1.53s \\ \hline
Ours & $\bm{1.38}$s & $\bm{1.37}$s & $\bm{1.37}$s & $\bm{1.38}$s & $\bm{1.39}$s & $\bm{1.37}$s & $\bm{1.39}$s \\ \hline
\end{tabularx}
\end{table}

A simple comparison of the run time of BF acceleration schemes is not reasonable because all acceleration methods make a tradeoff between speed and accuracy, and the speed is usually inversely proportional to the accuracy. To conduct a fair comparison, we keep the PSNR indices of filtering results unchanged and measure the speed of each method. Strictly, the PSNR indices cannot be kept the same since they are affected by various reasons. Here, we tweak the parameters of each acceleration method and make the PSNR indices of the filtering results around $32dB$ because these results are visually distinguished from the ground truth. Table~\ref{tab:run_time} plots run times. We can observe that our approach consumes the smallest run time among different $\sigma_s$ and $\sigma_r$. More importantly, unlike the shiftability property based methods, our run time does not depend on the parameter $\sigma_r$. This is because many terms are used to approximate the long tails of the Gaussian function by other methods. In contrast, we only approximate the  value on the truncated  interval.  Note that for large $\sigma_r$, the shiftability properties based methods (\ie Chaudhury~\cite{Chaudhury_TIP_2013} and ours) achieve the best performance. Unlike Chaudhury, our method needs to perform 3-D box filtering caused by the dimension promotion technique. Although both our method and the dimension promotion based methods~\cite{Gunturk_TIP_2011,Pan_MPE_2014} exploit the dimension promotion technique, our method is faster than these methods. This is because previous methods need the multiplication operation along $z$ axis and our approach only requires the addition operation.  This also proves that the filtering accuracy can be significantly improved without sacrificing efficiency.


\subsubsection{Acceleration by pre-computation}

Our method can be accelerated further by pre-computation. In this section, we will investigate how to integrate pre-computation with our algorithm with respect to two cases.

\textbf{Case 1:} \emph{The input image $I$ is given, but the parameters setting $(\sigma_r, \sigma_s)$ of BF is not determined}. This is a frequently encountered situation in practice. For example, when  an image already has been loaded into an image manipulation software such as Adobe Photoshop but the filtering parameters are still not determined by users, we are in this situation. If the image manipulation software can pre-compute something during the time waiting for the filtering parameters, the latency time for final results will be undoubtedly decreased.  Badly, authors of existing acceleration methods~\cite{Yang_CVPR_2009,Gunturk_TIP_2011,Pan_MPE_2014,Porikli_CVPR_2008,Chaudhury_TIP_2013} do not propose any method to complete the task because all their acceleration schemes depend on the exact values of $\sigma_r, \sigma_s$. That is to say, existing  methods cannot perform any calculation without knowing $(\sigma_r, \sigma_s)$. Unlike them, our method is able to perform pre-computation with little modification. Strictly speaking, our algorithm also cannot determine the exact value of $T_r$ before knowing $\sigma_r$,  but we can make a tiny modification for the original algorithm to pre-compute the value of  $F_c(\bm{y}, z)$ and $F_s(\bm{y}, z)$ with respect to some predefined values $\{T_{r_i}\}$. The specific reason is that  $F_c(\bm{y}, z)$ and $F_s(\bm{y}, z)$ in \eqref{eq:numerator_2D_box_N_terms} only involves an unknown parameter $T_r$ indicating the truncated region of the range kernel $K_r(x)$. Once the exact value of $T_r$ is known, we can always take the minimum $T_{r_i}$ such that $T_{r_i} \geq T_r$  from the predefined set $\{T_{r_i}\}$ to replace $T_r$.  In plain English,  this is equivalent to extend the truncated region from $[-T_r, T_r]$ to $[-T_{r_i}, T_{r_i}]$ and employ $\cos(\frac{\pi k}{T_{r_i}}x)$ to approximate the values of $K_r(x)$ on the region $[-T_{r_i}, T_{r_i}]$.  Since $F_c(\bm{y}, z)$ and $F_s(\bm{y}, z)$ have been figured out during the time loading images, our algorithm only need to perform box filtering which is a fast computation. Hence, without the computational burden for $F_c(\bm{y}, z)$ and $F_s(\bm{y}, z)$,  the actually filtering time can be reduced from about $1.37s$ to about $0.69s$ according to our experiment.  
At last, to describe the pre-computation algorithm more formally, we outline the major computation steps of our  pre-computation algorithm in the following:
\begin{itemize}
	\item Pre-computation while loading an image into the manipulation software
	\begin{enumerate}
		\item Input image $I$ and predefined values $\{T_{r_i}\}$.
		\item Calculate a set of $F_c(\bm{y}, z)$ and $F_s(\bm{y}, z)$ according to $I$ and the values in $\{T_{r_i}\}$.
	\end{enumerate}
	\item Filtering procedure after specifying the parameter $(\sigma_r, \sigma_s)$ 
	\begin{enumerate}
		\item Find the minimum  $T_{r_i}$ such that $T_{r_i} \geq 3\sigma_r$  from the predefined set $\{T_{r_i}\}$.
		\item Compute the coefficients $a_k $ of the best $N$-term approximation for the range kernel on the range $[-T_{r_i}, T_{r_i}]$.
		\item Put the pre-computed $F_c(\bm{y}, z)$ and $F_s(\bm{y}, z)$ corresponding to $T_{r_i}$ as well as $a_k $ into our fast computation formulas listed in Section~\ref{sec:3DFBF}  to yield final filtering results.
	\end{enumerate}
\end{itemize}


\begin{figure}[t]
\centering
\begin{subfigure}[b]{0.48\textwidth}
\includegraphics[width=\textwidth ]{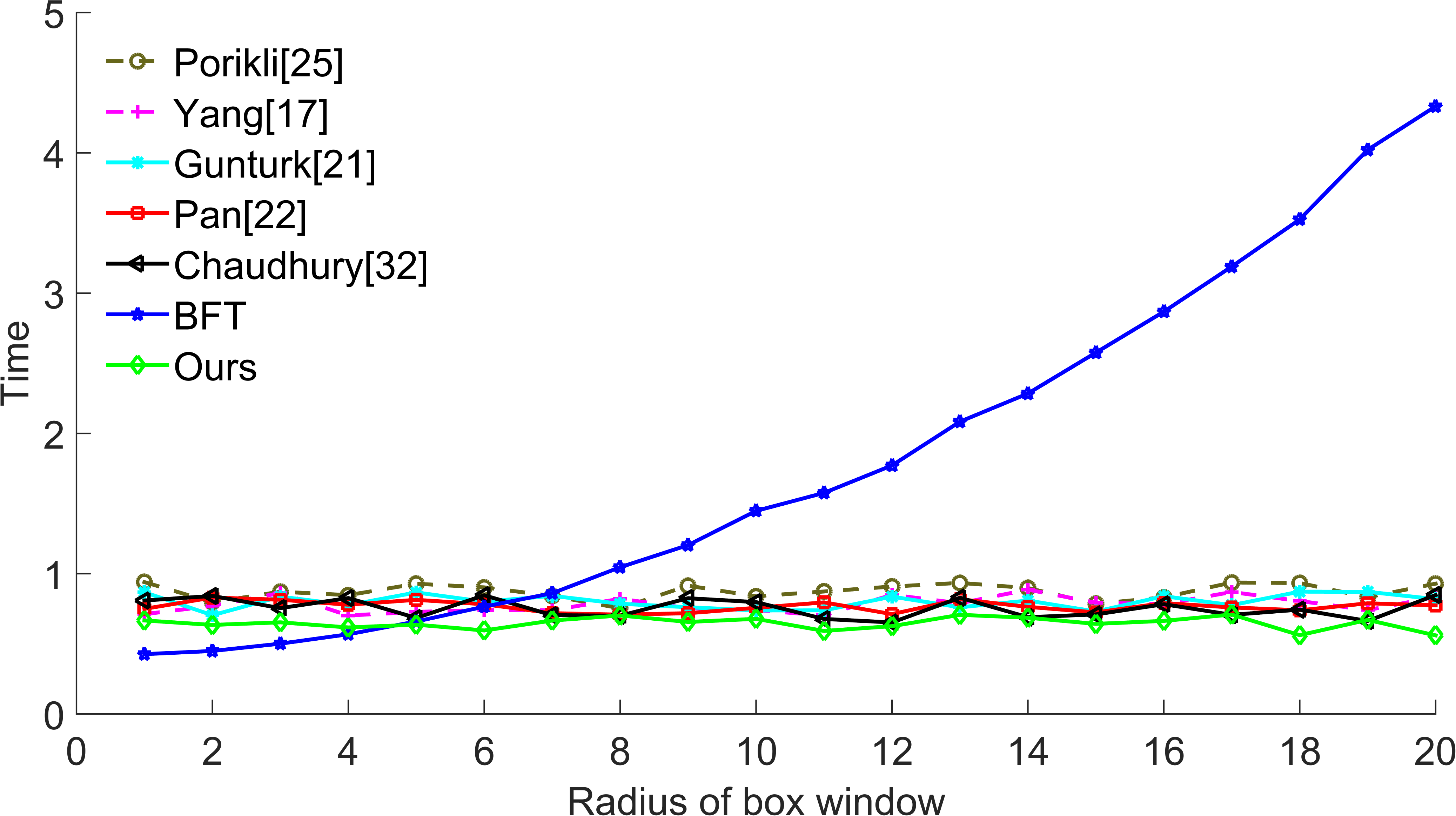}
\end{subfigure}
\caption{Run time illustration, where the abscissa axis denotes the radius of box window and the ordinate axis presents the run time of each method. BFT achieves the smallest run time when the radius is small and increases its run time with the radius of box window. In contrast, run times of other acceleration algorithms are constants and therefore are not dependent on window's radius.}
\label{fig:time}
\end{figure}

\textbf{Case 2:} \emph{Both input image $I$ and parameters setting $(\sigma_r, \sigma_s)$ are given}. This situation implies that we need to process input image immediately. Taking a close look at \eqref{eq:BF_promotion} \eqref{eq:BF_PBFIC} \eqref{eq:shiftable}, we can find that all acceleration methods involve exponential function $\exp(x)$ or trigonometric functions $\sin(x), \cos(x)$. As we known, both exponential function and trigonometric function are time-consuming operations. Since the value of input image is discrete, we can pre-compute a look-up table for the value of exponential/trigonometric functions on these discrete points to speed up the computation further. It is also wroth noting that the major reason that slows down the brute-force implementation of BF is also the exponential function used in \eqref{eq:BF}. Taking the look-up table technique, we can accelerate the brute-force implementation too. However, we have to point out that the look-up table technique cannot reduce the computational complexity of the brute-force implementation and thus its run time still depends on the size of box window $\mathcal{N}_x$ of BF. Fig~\ref{fig:time} illustrates the run time of seven methods with respect to different window radii. We can plot this figure because Table~\ref{tab:run_time} implies that the speed of different acceleration methods is rather robust for different parameter settings, thus the run time at arbitrary parameter settings such as $(\sigma_r=40 , \sigma_s=5)$ can stand for the performance of one kind of acceleration method. Here the seven methods include Porikli~\cite{Porikli_CVPR_2008},  Yang~\cite{Yang_CVPR_2009}, Gunturk~\cite{Gunturk_TIP_2011}, Pan~\cite{Pan_MPE_2014}, Chaudhury~\cite{Chaudhury_TIP_2013}, ours as well as the brute-force implementation accelerated by the look-up table (BFT). From Fig~\ref{fig:time}, we can verify that the run time of BFT increases with the radius of the box window $\mathcal{N}_x$. In addition, when the size of window is small, the fastest method is BFT. The reason is that the overall computational cost of BFT is dominated by the size of the window $\mathcal{N}_x$, so smaller window size, lower computational cost and the run time will increase with the window size. Unlike BFT, the run time of all acceleration methods do not change with the window size because their computational complexity is $O(|I|)$. Note that disregarding the small window case, our acceleration algorithm overwhelms other acceleration methods.

\begin{table}[t]
\caption{Support regions and the sizes of the box filters used to represent 2-D Haar functions.}
\label{tab:support_regions}
\centering
\begin{tabularx}{\linewidth}{@{}YYY@{}}
\hline
 & Support Region  & Support Region's Size \\ \hline
$\ddot{B}(\bm{x})$ & $A \times A$ & $2T_s \times 2T_s$ \\
$\ddot{B}^1_{0, j_2,k_2}(\bm{x})$ & $A \times A^{j_2,k_2}_1$ & $2T_s \times 2^{-j_2+1}T_s$ \\
$\ddot{B}^2_{0, j_2,k_2}(\bm{x})$ & $A \times A^{j_2,k_2}_2$ & $2T_s \times 2^{-j_2+1}T_s$ \\
$\ddot{B}^1_{j_1,k_1,0}(\bm{x})$ & $A^{j_1,k_1}_1 \times A$ & $ 2^{-j_1+1}T_s \times 2T_s $ \\
$\ddot{B}^2_{j_1,k_1,0}(\bm{x})$ & $A^{j_1,k_1}_2 \times A$ & $2^{-j_1+1}T_s \times 2T_s$ \\
$\ddot{B}^1_{j_1,k_1,j_2,k_2}(\bm{x})$ & $A^{j_1,k_1}_1 \times A^{j_2,k_2}_1$ & $ 2^{-j_1+1}T_s \times 2^{-j_2+1}T_s $ \\
$\ddot{B}^2_{j_1,k_1,j_2,k_2}(\bm{x})$ & $A^{j_1,k_1}_1 \times A^{j_2,k_2}_2$ & $ 2^{-j_1+1}T_s \times 2^{-j_2+1}T_s $\\
$\ddot{B}^3_{j_1,k_1,j_2,k_2}(\bm{x})$ & $A^{j_1,k_1}_2 \times A^{j_2,k_2}_2$ & $ 2^{-j_1+1}T_s \times 2^{-j_2+1}T_s $\\
$\ddot{B}^4_{j_1,k_1,j_2,k_2}(\bm{x})$ & $A^{j_1,k_1}_2 \times A^{j_2,k_2}_1$ & $ 2^{-j_1+1}T_s \times 2^{-j_2+1}T_s $ \\ \hline
\end{tabularx}
\end{table}


\section{Conclusion}
In this paper, we propose a unified framework to accelerate BF with arbitrary spatial and range kernels. Unlike previous approaches, our method jointly employs five techniques: truncated spatial and range kernels, the best $N$-term approximation of these kernels as well as three existing acceleration techniques to speed up BF.  It thus can inherit the advantage of  previous acceleration algorithms while avoiding the problems of previous approaches.
Moreover, we first transform 2-D BF into a set of 3-D box filters and disclose that BF can be fast computed by the 3-D SAT.
Taking advantage of the expressive orthogonal functions used in the best $N$-term approximation scheme, our approach employs fewer terms which means faster computing speed, while obtaining more gratifying filtering results among
a great wide of parameter settings. More importantly, the strength of our method has been verified by
several carefully designed experiments. All experiments indicate that our method outperforms other methods.  Especially, the filtering accuracy is significantly improved without sacrificing efficiency.

\appendices
\section{Box filter representation}
\label{sec:box_filter}

\begin{figure}[t]
    \begin{subfigure}[b]{0.47\linewidth}
        \includegraphics[width=\textwidth ]{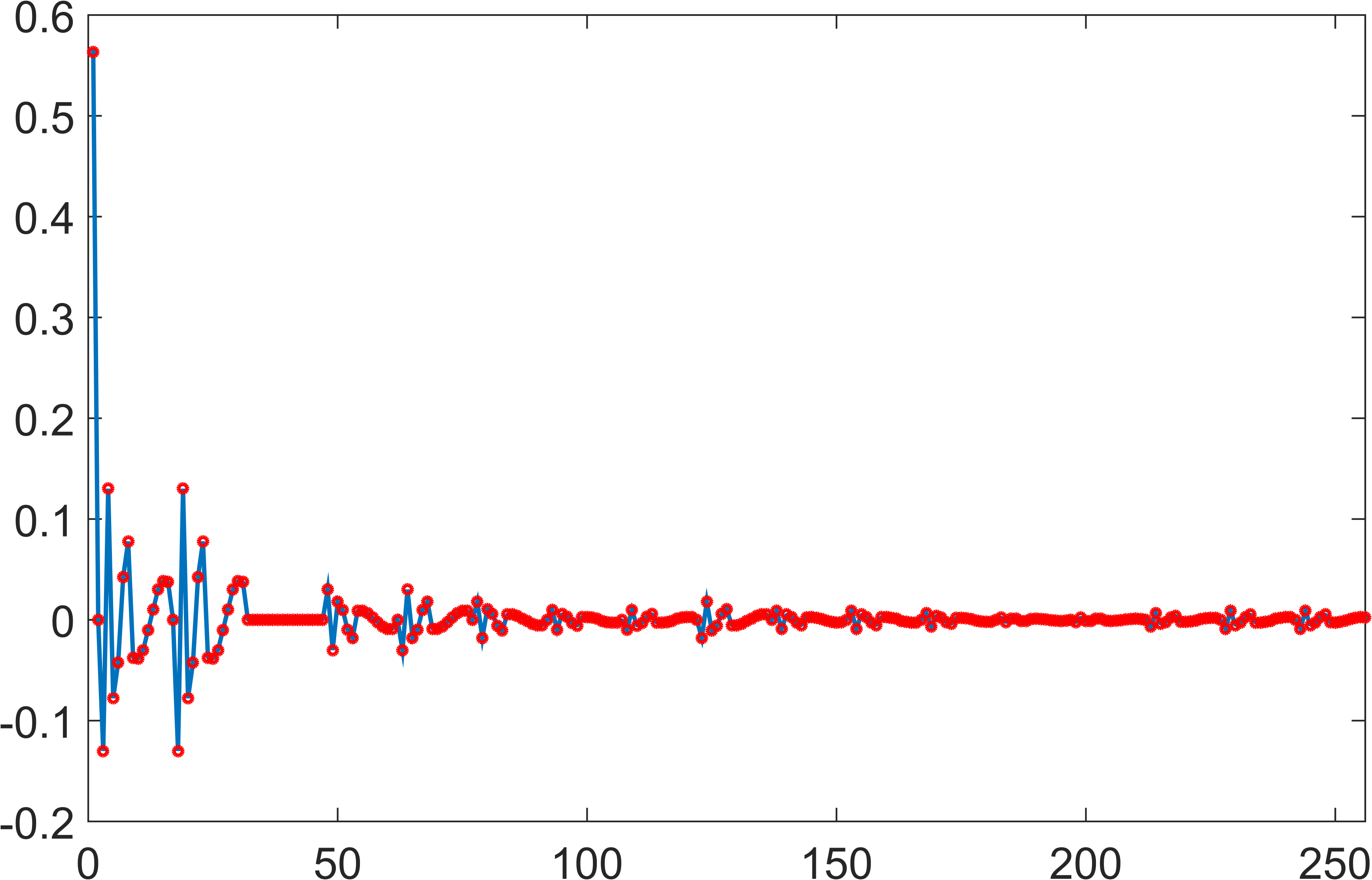}
        \caption{Haar decomposition}
        \label{fig:decomposition:Haar}
    \end{subfigure}
    \begin{subfigure}[b]{0.47\linewidth}
        \includegraphics[width=\textwidth ]{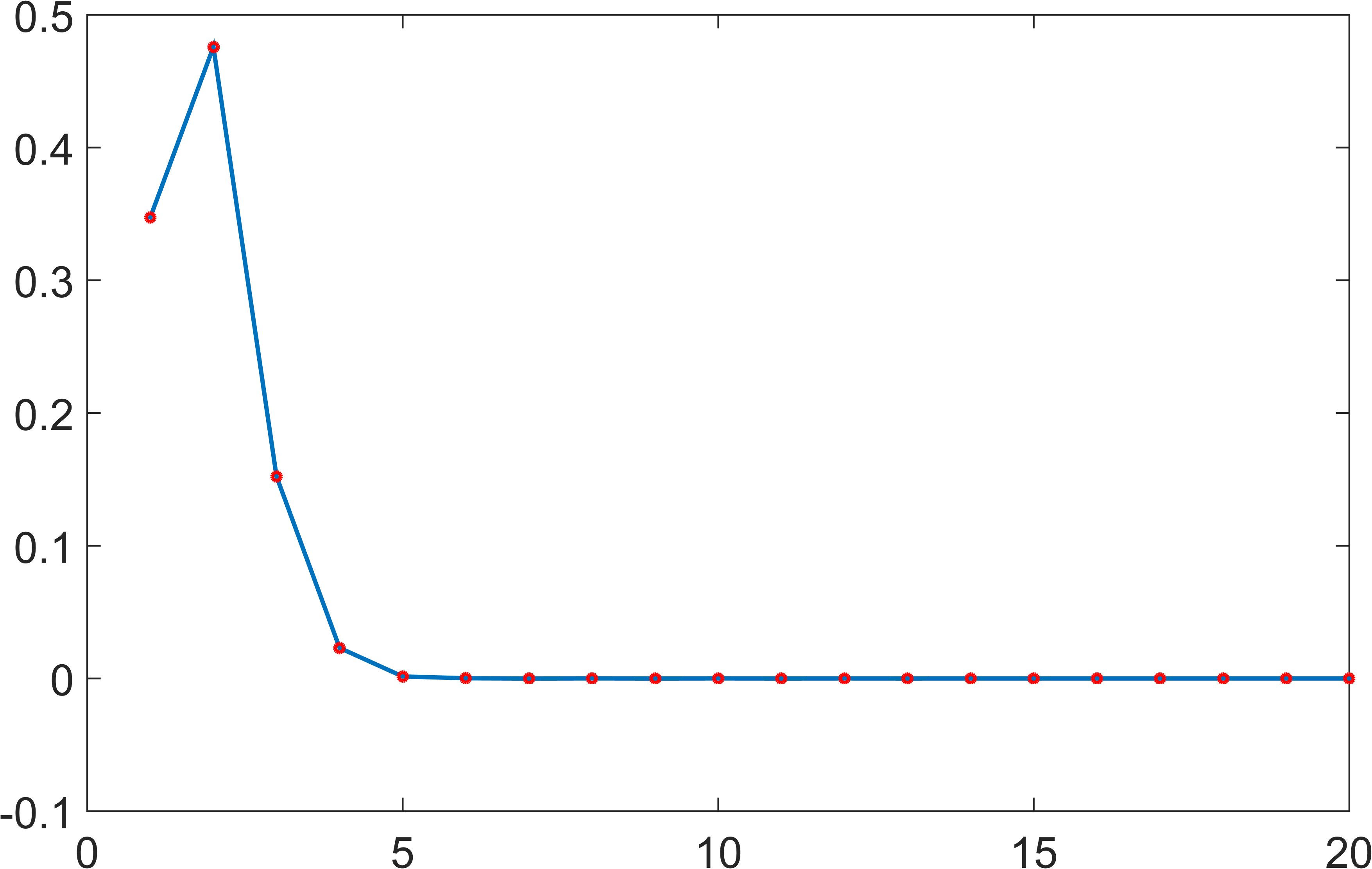}
        \caption{Cosine decomposition}
        \label{fig:decomposition:Cosine}
    \end{subfigure}
    \caption{Visual illustration for the amplitude of the decomposition coefficients of the spatial kernel $G_5(x)$ and range kernel $G_{40}(x)$ on the Haar and trigonometric basis.} 
    \label{fig:decomposition}
\end{figure}

All functions in the four sets $\{\phi(x)\phi(y)\}$, $\{\phi(x)\psi_{j_2,k_2}(y)\}$, $\{\psi_{j_1,k_1}(x) \phi(y)\}$ $\{\psi_{j_1,k_1}(x)\psi_{j_2,k_2}(y) \}$ can be represented by a linear combination of box filters. First, $\phi(x)$ achieves $1$ in the support region $A = [-T_s, T_s]$. Similarly, let $Z_k = \sign(k)(2|k|-1)T_s$, $A^{j,k}_1 = [{2^{-j}}{(Z_k-T_s)}, {2^{-j}}{Z_k}]$ and $A^{j,k}_2 = [{2^{-j}}{Z_k}, {2^{-j}}{(Z_k+T_s)}]$, the support region of $\psi_{j,k}(x)$ can be presented as $A^{j,k}_1 \bigcup A^{j,k}_2$ because $\psi_{j,k}(x) = 1$ for $x \in A^{j,k}_1$ and $\psi_{j,k}(x) = -1$ for $x \in A^{j,k}_2$.
Second, $\phi(x)\phi(y)$ represents a 2-D box function $\ddot{B}(\bm{x})$ with the support region $A \times A$ as $\phi(x)$ is constant in the region $A$. Unlike $\phi(x)\phi(y)$, $\phi(x)\psi_{j_2,k_2}(y)$ and $\psi_{j_1,k_1}(x) \phi(y)$ are binary functions equal to 1 in the regions $A \times A^{j_2,k_2}_1$, $A^{j_1,k_1}_1 \times A$ and obtain $-1$ in the regions $A \times A^{j_2,k_2}_2$, $A^{j_1,k_1}_2 \times A$, respectively. Hence putting $\ddot{B}^1_{0, j_2,k_2}(\bm{x})$, $\ddot{B}^2_{0, j_2,k_2}(\bm{x})$, $\ddot{B}^1_{j_1,k_1,0}(\bm{x})$ and $\ddot{B}^2_{j_1,k_1,0}(\bm{x})$ be four 2-D box functions defined on the support regions $A \times A^{j_2,k_2}_1$, $A \times A^{j_2,k_2}_2$, $A^{j_1,k_1}_1 \times A $ and $A^{j_1,k_1}_2 \times A $ respectively, we thus have $\phi(x)\psi_{j_2,k_2}(y) = \ddot{B}^1_{0, j_2,k_2}(\bm{x}) - \ddot{B}^2_{0, j_2,k_2}(\bm{x})$,  $\psi_{j_1,k_1}(x) \phi(y) = \ddot{B}^1_{j_1,k_1,0}(\bm{x}) - \ddot{B}^2_{j_1,k_1,0}(\bm{x})$. Similar to $\phi(x)\psi_{j_2,k_2}(y)$ and $\psi_{j_1,k_1}(x) \phi(y)$, $\psi_{j_1,k_1}(x) \psi_{j_2,k_2}(y)$ is also a binary function. Unlike them, $\psi_{j_1,k_1}(x) \psi_{j_2,k_2}(y)$ equals to $1$ on the regions $A^{j_1,k_1}_1 \times A^{j_2,k_2}_1$, $A^{j_1,k_1}_2 \times A^{j_2,k_2}_2$ and obtains -1 on the region $A^{j_1,k_1}_1 \times A^{j_2,k_2}_2$, $A^{j_1,k_1}_2 \times A^{j_2,k_2}_1$. Let $\ddot{B}^1_{j_1,k_1,j_2,k_2}(\bm{x})$, $\ddot{B}^2_{j_1,k_1,j_2,k_2}(\bm{x})$, $\ddot{B}^3_{j_1,k_1,j_2,k_2}(\bm{x})$ and $\ddot{B}^4_{j_1,k_1,j_2,k_2}(\bm{x})$ be four 2-D box functions with the support regions $A^{j_1,k_1}_1 \times A^{j_2,k_2}_1$, $A^{j_1,k_1}_1 \times A^{j_2,k_2}_2$, $A^{j_1,k_1}_2 \times A^{j_2,k_2}_2$ and $A^{j_1,k_1}_2 \times A^{j_2,k_2}_1$, respectively, we can reformulate $\psi_{j_1,k_1}(x)\psi_{j_2,k_2}(y)$ as $\ddot{B}^1_{j_1,k_1,j_2,k_2}(\bm{x}) - \ddot{B}^2_{j_1,k_1,j_2,k_2}(\bm{x}) + \ddot{B}^3_{j_1,k_1,j_2,k_2}(\bm{x}) - \ddot{B}^4_{j_1,k_1,j_2,k_2}(\bm{x}) $. Finally, we list the support regions and the size of the box filters used to represent 2-D Haar functions in Table~\ref{tab:support_regions} for reference.

Given all this, we can reexpress $ K_s(\| \bm{x} \|) $ in \eqref{eq:Haar_N_terms} as
\begin{IEEEeqnarray}{C}
\begin{split}
 & K_s(\| \bm{x} \|)
\approx  \sum_{c_0 \in \Lambda^s_1} c_0  B(\bm{x}) \\
+ & \sum_{c_{0, j_2,k_2} \in \Lambda^s_2}  c_{0, j_2,k_2} \sum_{i=1}^2 (-1)^{i+1} B^i_{0, j_2,k_2}(\bm{x})  \\
+ & \sum_{c_{ j_1,k_1,0} \in \Lambda^s_3}  c_{j_1,k_1,0} \sum_{i=1}^2 (-1)^{i+1} B^i_{j_1,k_1,0}(\bm{x}) \\
+ & \sum_{c_{j_1,k_1,j_2,k_2} \in \Lambda^s_4} c_{j_1,k_1,j_2,k_2} \sum_{i=1}^4 (-1)^{i+1} B^i_{j_1,k_1,j_2,k_2}(\bm{x})
\end{split}
\label{eq:2D_box_N_terms}
\end{IEEEeqnarray}
which is consisted of several box filters. Further, let $\ddot{B}_{\bm{j}}(\bm{x})$ present the 2-D box functions used in \eqref{eq:2D_box_N_terms},  $c_{\bm{j}}$ be the corresponding coefficients of $\ddot{B}_{\bm{j}}(\bm{x})$ and $\Lambda^s$ stands for the collection of $c_{\bm{j}}$, $N^{\bm{j}}_{\bm{x}}$ be the support region of $\ddot{B}_{\bm{j}}(\bm{x})$, we have \eqref{eq:Box_N_terms}.


\section{How to compute the best $N$-term approximation?}
\label{sec:attenuation}

\begin{table}[t]
\centering
\caption{The corresponding relationship between $N$ and $M$ for spatial kernel $G_5(x)$ and range kernel $G_{40}(x)$.} 
\label{tab:NM}
\begin{tabularx}{\linewidth}{@{}c|Y|Y|Y|Y|Y|Y|Y|Y|Y@{}}
\hline
\multirow{2}{*}{Spatial kernel} & $N$ & 1 & 2 & 3 & 4 & 5 & 6 & 7 & 8 \\ \cline{2-10}
                  & $M$ &  1 &  19 & 19  &  19 &  19 &  19 &  19 & 19  \\ \hline \hline
\multirow{2}{*}{Range kernel} & $N$ & 1 & 2 & 3 & 4 & 5 & 6 & 7 & 8 \\ \cline{2-10}
                  & $M$ & 2  &  2 & 3  & 4  &  5 & 6  & 7  &  8 \\ \hline
\end{tabularx}
\end{table}

Theoretically, finding the best $N$-term approximation of a given function requires us to minimize the objective function~\eqref{eq:N_term}. However, in practice the minimizing is not necessary because the best $N$-term approximation for arbitrary functions can be obtained by selecting the first $N$ largest coefficients and the amplitude of coefficients attenuates very fast. The two points imply that the first $N$ largest coefficients must reside in the first $M$ coefficients, where $M$ is a constant which is equal or greater than $N$. More clearly speaking, to find the best $N$-term approximation, we only need to calculate the first $M$ coefficients and then pick up the first $N$ largest coefficients from the first $M$ coefficients.

As an example, Fig~\ref{fig:decomposition} illustrates the coefficients on the Haar basis and trigonometric basis for the Gaussian function respectively, where the abscissa axis denotes the linear order of each basis functions, the ordinate axis represents the amplitude of coefficient of each basis functions. Note that since each Haar function $\psi_{i,j}$ is identified by two indices $i$, $j$, we order each Haar function according to following rules: for two Haar functions $\psi_{i_1,j_1}$, $\psi_{i_2,j_2}$, if $i_1 < i_2$ or $i_1 = i_2$ and $j_2 < j_2$,  $\psi_{i_1,j_1}$ is putted before $\psi_{i_2,j_2}$; otherwise, $\psi_{i_1,j_1}$ is located after $\psi_{i_2,j_2}$. As for the trigonometric basis, all coefficients of sine functions are zeros, we thus sort the cosine functions $\cos(k \omega x)$ according the index $k$. From Figs~\ref{fig:decomposition:Haar}~\ref{fig:decomposition:Cosine}, we can observe that both the coefficients of Haar basis functions and the coefficients of trigonometric basis functions decay dramatically. The corresponding relationship between $N$ and $M$ for spatial kernel $G_5(x)$ and range kernel $G_{40}(x)$ is listed in Table~\ref{tab:NM} which tells us the minimal number $M$ of the first $M$ coefficients containing the first $N$ largest coefficients. From this table we can reasonably conclude that finding the best $N$-term approximation only needs to figure out the first $20N$ coefficients and then to select the first $N$ largest coefficients and their corresponding coefficients.


\section*{Acknowledgment}
This work was supported by National Natural Science Foundation of China with Nos. 61331018, 91338202, 61572405 and 61571046, and by China 863 program with No. 2015AA016402.

\ifCLASSOPTIONcaptionsoff
  \newpage
\fi



\bibliographystyle{IEEEtran}
\bibliography{./bib/paper}
%
%
%

%
\begin{IEEEbiography}[{\includegraphics[width=1in,height=1.25in,clip,keepaspectratio]{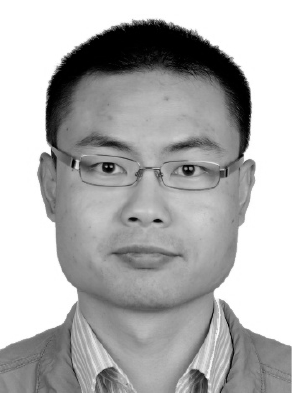}}]{Longquan Dai}
received his B.S. degree in Electronic Engineering from Henan University of Technology, China, in 2006. He received his M.S. degree in Electronic Engineering from Shantou University, China, in 2010. Currently, he is working toward the PhD degree in Computer Science  at institute of automation, Chinese academy of sciences, China. His research interests research interests lie in computer graphics, computer vision and optimization-based techniques for image analysis and synthesis.
\end{IEEEbiography}

\vspace{-1cm}
\begin{IEEEbiography}[{\includegraphics[width=1in,height=1.25in,clip,keepaspectratio]{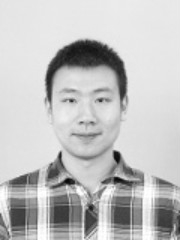}}]{Mengke Yuan}
received his B.S. and M.S. degree in Mathematics from Zhengzhou University in 2012 and 2015 respectively.
He is currently pursuing the PhD degree in Computer Science at Institute of Automation, Chinese
Academy of Sciences.His research interests lie in
computer vision,optimization-based techniques for image analysis and synthesis and Machine Learning.
\end{IEEEbiography}
\vspace{-1cm}
\begin{IEEEbiography}[{\includegraphics[width=1in,height=1.25in,clip,keepaspectratio]{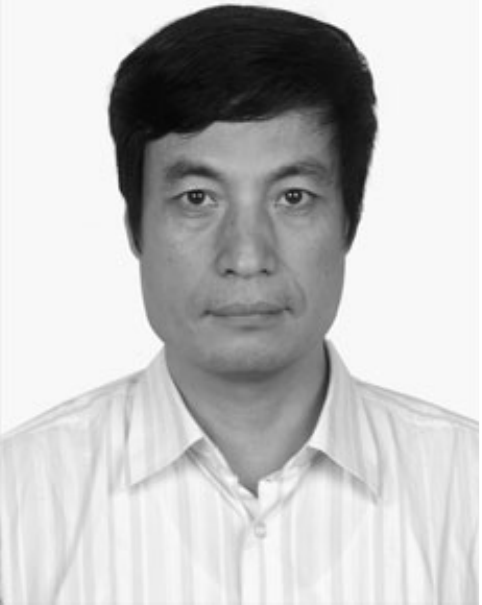}}]{Xiaopeng Zhang}
(M’11) received the B.S. and M.S. degrees in mathematics from Northwest
University, Xi’an, China, in 1984 and 1987,
respectively, and the Ph.D. degree in computer
science from the Institute of Software, Chinese
Academy of Sciences, Beijing, China, in 1999. He is
currently a Professor with the National Laboratory
of Pattern Recognition, Institute of Automation,
Chinese Academy of Sciences. His main research
interests are computer graphics and computer vision.
\end{IEEEbiography}






\end{document}